%% file: main.tex
\documentclass{article}

\PassOptionsToPackage{numbers}{natbib}
\usepackage[main,final]{neurips_2026}

\usepackage[utf8]{inputenc} 
\usepackage[T1]{fontenc}    
\usepackage{hyperref}       
\usepackage{url}            
\usepackage{booktabs}       
\usepackage{amsfonts}       
\usepackage{nicefrac}       
\usepackage{microtype}      
\usepackage{xcolor}         
\usepackage{microtype}
\usepackage{graphicx}
\usepackage{subcaption}
\usepackage{booktabs} 
\usepackage{tcolorbox}
\usepackage{ragged2e}
\usepackage{xcolor}
\usepackage{multirow}
\usepackage{wrapfig}
\usepackage{placeins}
\usepackage{amsmath}
\usepackage{amssymb}
\usepackage{mathtools}
\usepackage{amsthm, enumitem}
\usepackage{hyperref}
\usepackage{pgffor}
\usepackage{array}
\input{mathcommands}
\newcommand{\ours}{{\texttt{TabClustPFN}}}
\newcommand{\pin}{{\textsc{PIN}}}
\newcommand{\cin}{{\textsc{CIN}}}

\title{TabClustPFN: A Prior-data Fitted Network for Tabular Data Clustering}

\author{%
  \makebox[0.45\textwidth][c]{Tianqi Zhao} \\
  Renmin University of China
  \And
  \makebox[0.45\textwidth][c]{Guanyang Wang} \\
  Rutgers University
  \AND
  \makebox[0.45\textwidth][c]{Yan Shuo Tan} \\
  National University of Singapore
  \And
  \makebox[0.45\textwidth][c]{Qiong Zhang\thanks{Correspondence: \texttt{qiong.zhang@mbzuai.ac.ae}.}} \\
  Mohamed bin Zayed University \\
  of Artificial Intelligence
}

\begin{document}

\maketitle

\begin{abstract}
Prior-data Fitted Networks (PFNs) have reframed supervised tabular learning as single-pass in-context inference without per-dataset optimization. 
Extending this paradigm to clustering is appealing yet fundamentally more challenging, due to absent supervision, unknown cluster cardinality, and label switching inherent to partition outputs.
Existing PFN-based clustering methods address these challenges only partially, either requiring known cardinality as input or relying on unstable label-ordering conventions and overly restrictive synthetic priors. 
We introduce \ours{}, a clustering PFN that resolves all challenges jointly through co-designed prior, objective, and architecture. Our hybrid pretraining prior captures heterogeneous real-tabular geometry; our decoupled partition inference network--cardinality inference network architecture jointly infers cluster assignments and cardinality in a single pass; and our SoftARI training objective is permutation-invariant by construction, eliminating the need for any label-ordering convention. 
On a 44 curated real-world tabular benchmark, \ours{} achieves state-of-the-art clustering performance against classical, deep, and amortized baselines, with runtime comparable to efficient baselines.
Code is available \href{https://github.com/Tianqi-Zhao/TabClustPFN}{here}.
\end{abstract}

\input{sections/intro}

\input{sections/tabpfn}

\input{sections/method}
\input{sections/exp}
\input{sections/discussion}
\input{sections/conclusion}

\begin{ack}
Tianqi Zhao and Qiong Zhang are supported by the National Key R\&D Program of China Grant 2024YFA1015800. Yan Shuo Tan is supported by NUS Start-up Grant A-8000448-00-00 and MOE AcRF Tier 1 Grants A-8002498-00-00 and A-8004458-00-00.

This work was conducted while Qiong Zhang was at the Institute of Statistics and Big Data, Renmin University of China. Qiong Zhang is now affiliated with Mohamed bin Zayed University of Artificial Intelligence.
\end{ack}

\bibliography{biblio}
\bibliographystyle{abbrv}


\clearpage
\appendix
\input{sections/appendix/related}
\input{sections/appendix/training_details}

\input{sections/appendix/exp_details}
\input{sections/appendix/additional_results}

\input{sections/appendix/ablation}


\FloatBarrier
\newpage
\input{sections/checklist.tex}

\end{document}

%% file: mathcommands.tex
\usepackage{amsmath,amsfonts,bm, bbm}

\newcommand{\eg}{{\em e.g.,~}}

\def\mX{{\bm{X}}}

\def\mZ{{\bm{Z}}}

\DeclareMathAlphabet{\mathsfit}{\encodingdefault}{\sfdefault}{m}{sl}
\SetMathAlphabet{\mathsfit}{bold}{\encodingdefault}{\sfdefault}{bx}{n}

\def\gD{{\mathcal{D}}}

\def\gN{{\mathcal{N}}}
\def\gO{{\mathcal{O}}}

\def\sP{{\mathbb{P}}}

\DeclareMathOperator*{\argmax}{arg\,max}

\newcommand{\bea}{\begin{eqnarray*}}
\newcommand{\eea}{\end{eqnarray*}}
\newcommand{\ba}{\begin{eqnarray*}}
\newcommand{\ea}{\end{eqnarray*}}
\newcommand{\be}{\begin{equation}}
\newcommand{\ee}{\end{equation}}
\newcommand{\bi}{\begin{itemize}}
\newcommand{\ei}{\end{itemize}}

\makeatletter
\newcommand*\rel@kern[1]{\kern#1\dimexpr\macc@kerna}
\newcommand*\widebar[1]{%
  \begingroup
  \def\mathaccent##1##2{%
    \rel@kern{0.8}%
    \overline{\rel@kern{-0.8}\macc@nucleus\rel@kern{0.2}}%
    \rel@kern{-0.2}%
  }%
  \macc@depth\@ne
  \let\math@bgroup\@empty \let\math@egroup\macc@set@skewchar
  \mathsurround\z@ \frozen@everymath{\mathgroup\macc@group\relax}%
  \macc@set@skewchar\relax
  \let\mathaccentV\macc@nested@a
  \macc@nested@a\relax111{#1}%
  \endgroup
}
\makeatother

%% file: sections/intro.tex
\section{Introduction}
\label{sec:intro}
\begin{wrapfigure}[15]{r}{0.45\textwidth}
\vspace{-0.4cm}
\centering
\includegraphics[width=0.85\linewidth]{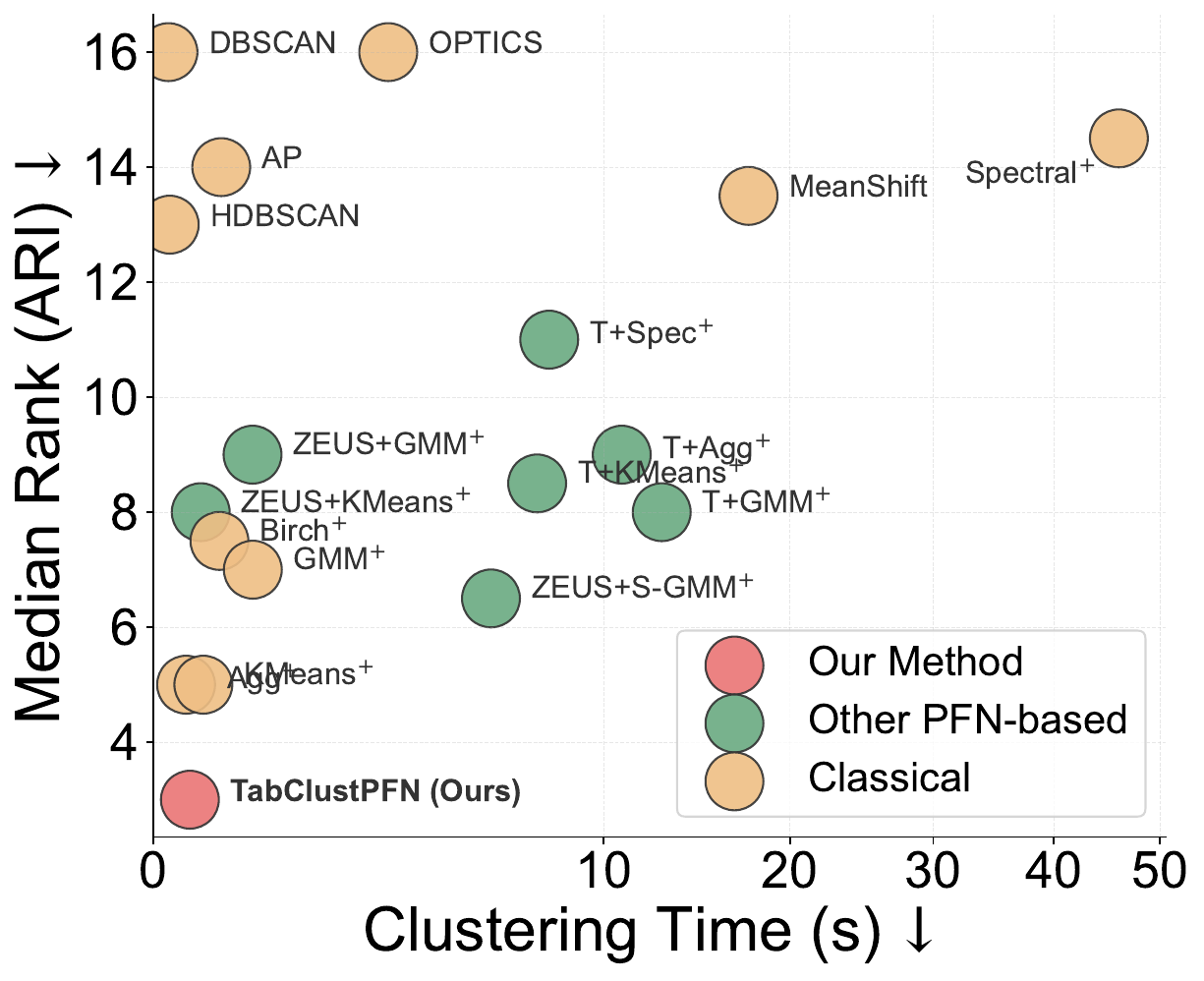}
\caption{\textbf{Time \& median ARI rank (lower is better).} \ours{} achieves the best ARI with runtime comparable to simple methods and faster than other PFN-based methods.}
\label{fig:performance}
\vspace{-15pt}
\end{wrapfigure}
Prior-data Fitted Networks (PFNs)~\citep{hollmann2025accurate, qu2025tabicl,
grinsztajn2025tabpfn, ma2025tabdpt, zhang2025mitra, qu2026tabiclv2} have established a new paradigm for supervised tabular learning.
By pretraining a transformer to approximate a Bayesian posterior predictive distribution over a broad synthetic prior, PFN-based models such as TabPFN v2.5~\citep{grinsztajn2025tabpfn} achieve strong generalization via in-context-learning in a single forward pass, without any per-dataset optimization.
This approach has proven remarkably effective, delivering state-of-the-art performance~\citep{erickson2025tabarena} across diverse tabular classification and regression tasks while remaining orders of magnitude faster than dataset-specific alternatives.

\emph{Applying this paradigm to clustering would be highly valuable}.
While clustering is central to tabular exploratory analysis, the lack of labels forces costly trial-and-error and unreliable heuristics to select the cardinality, an algorithm with a suitable inductive bias, and hyperparameters.
The PFN paradigm offers a way to amortize these choices into pretraining, using a synthetic prior to encode clustering biases and in-context inference to predict both $\mZ$ and $K$ in a single pass.
However, employing the PFN paradigm to clustering is fundamentally harder than its supervised counterpart.

Three obstacles arise specifically from the unsupervised nature of clustering:

\textbf{(1) No supervision signal}: 
unlike supervised PFNs, where labels define class boundaries, clustering contexts contain only features. The model must hence infer the notion of clusters from a synthetic prior, making its design critical to capturing meaningful cluster structure and real-data geometry.

\textbf{(2) Unknown cardinality}: 
Context labels define the output space in supervised PFNs, whereas clustering must infer $K$ from data, a challenge even for classical methods~\citep{lloyd1982least, ester1996density}.

\textbf{(3) Label switching}: any permutation of the cluster labels yields an equally valid partition.
Standard cross-entropy objectives assume a fixed label correspondence and are therefore
inapplicable.

Recent PFN-based methods \emph{only partially} address these challenges.
ZEUS~\citep{marszaek2025zeus} learns expressive tabular embeddings but delegates clustering to $k$-means with known $K$, hence leaving challenge~(2) unresolved.
Cluster-PFN~\citep{bhaskaran2025transformer} jointly infers assignments and cardinality
but resolves label switching by hard ordering clusters according to their distance to the origin, a convention that is undefined when two centers are equidistant (\eg $-1$ and $+1$ in 1D) and unstable in high dimensions~\citep{aggarwal2001surprising}.
Its pretraining prior is further restricted to GMMs with $D \leq 5$, which do not capture the nonlinear geometry or feature heterogeneity of real tabular data.
See a detailed comparison in App.~\ref{app:cluster-pfn}.

\textbf{No existing method resolves all 3 challenges simultaneously.}
We introduce \ours{}, a PFN that jointly resolves these challenges via a tightly co-designed prior, objective, and architecture.
This joint design is crucial: realistic tabular priors induce diverse partitions that demand permutation-invariant loss, while robust inference of $K$ under such partition-level objectives requires decoupling cluster-assignment inference from cardinality inference.
Our contributions are:
\begin{itemize}[leftmargin=*]

\item \textbf{A hybrid pretraining prior [challenge~(1)].}
We combine controlled-overlap GMMs with nonlinear invertible transformations and
mixed numerical-categorical feature generation~\citep{marszaek2025zeus}, covering
the heterogeneous geometries and feature types found in real tabular data.
Ablations confirm each component contributes to generalization (Tab.~\ref{tab:prior_ablation_ari}).

\item \textbf{A decoupled novel PIN-CIN architecture [challenge~(2)].}
Our model has 2 components: a Partition Inference Network (PIN) maintains $K$ explicit cluster prototypes, iteratively co-refined with data representations via cross-attention; and a Cardinality Inference Network (CIN) estimates $p(K|\mX)$ from permutation-invariant Gram matrix summaries of PIN outputs across all candidate $K$ simultaneously, unlike prior methods that require $K$ as input or use sequential single $K$ extraction.

\item \textbf{SoftARI: a permutation-invariant training objective [challenge~(3)].}
We train the \pin{} with a differentiable relaxation of the Adjusted Rand Index (ARI), which is permutation-invariant by construction and requires no label ordering convention.
SoftARI matches or outperforms matching-based cross-entropy alternatives while halving per-iteration pretraining cost (Tab.~\ref{tab:pin_loss_ablation}).

\item \textbf{State-of-the-art results on a broad tabular clustering benchmark.}
We introduce a curated benchmark of $44$ real-world tabular datasets and evaluate
\ours{} against classical, deep, and amortized clustering baselines (App.~\ref{app:related_work}).
\ours{} achieves SOTA performance (Fig.~\ref{fig:performance}), confirmed
by paired $t$-tests, Wilcoxon tests, and critical-difference diagrams (App.\ref{app:significance}).
\end{itemize}

%% file: sections/tabpfn.tex
\section{An overview of TabPFN}
\label{sec:pfn_overview}
TabPFN~\citep{hollmann2025accurate} is a canonical \emph{Prior-data Fitted Network} (PFN) that serves as a tabular foundation model for  supervised learning, casting prediction as Bayesian inference.
Given a dataset $\gD=\{(X_i,Y_i)\}_{i=1}^n$ drawn IID from an unknown distribution $\pi^*$ and a prior $p$ over a space of data-generating distributions $\Pi$, TabPFN approximates the posterior predictive distribution (PPD): $p(y| x,\gD) = \int_\Pi \pi(y|x)\,p(\pi|\gD)\,d\pi.$
To achieve this without intractable integrals, TabPFN uses PFN: a transformer is pretrained to approximate this PPD directly, enabling in-context learning (ICL).
At test time, the model parameters are fixed; it processes the dataset $\gD$ as a sequence of tokens in the context window and predicts the target $y$ for a query $x$ in a single forward pass.
Another core innovation of TabPFN lies in its pretraining data. 
Instead of using real-world datasets, the model is pretrained on a large collection of synthetic data sampled from a prior defined by structural causal models~\citep{pearl2009causality}. 
This prior is designed to produce distributions that mimic the complexity of real tabular data, including nonlinear feature interactions, heteroskedastic noise, and mixed feature types. 
Pretraining proceeds by repeatedly sampling synthetic datasets from this prior ($\approx$ 130M datasets). 
Each dataset is divided into a context set (pseudo-training) and a query set (pseudo-test), and the transformer is trained to maximize the likelihood of query labels given the context.

In summary, the PFN framework comprises three components: (i) a \textbf{\emph{task-level prior}} that defines the distribution over data-generating processes; (ii) a \textbf{\emph{transformer architecture}} that performs amortized inference via ICL; and
(iii) a \textbf{\emph{pretraining objective}} that aligns predictions with the true PPD.
This structure provides a principled blueprint for extension. 


%% file: sections/method.tex
\section{TabClustPFN}
\label{sec:methods}
\begin{wrapfigure}{r}{0.45\textwidth}
\vspace{-42pt}
\centering
\includegraphics[width=0.85\linewidth]{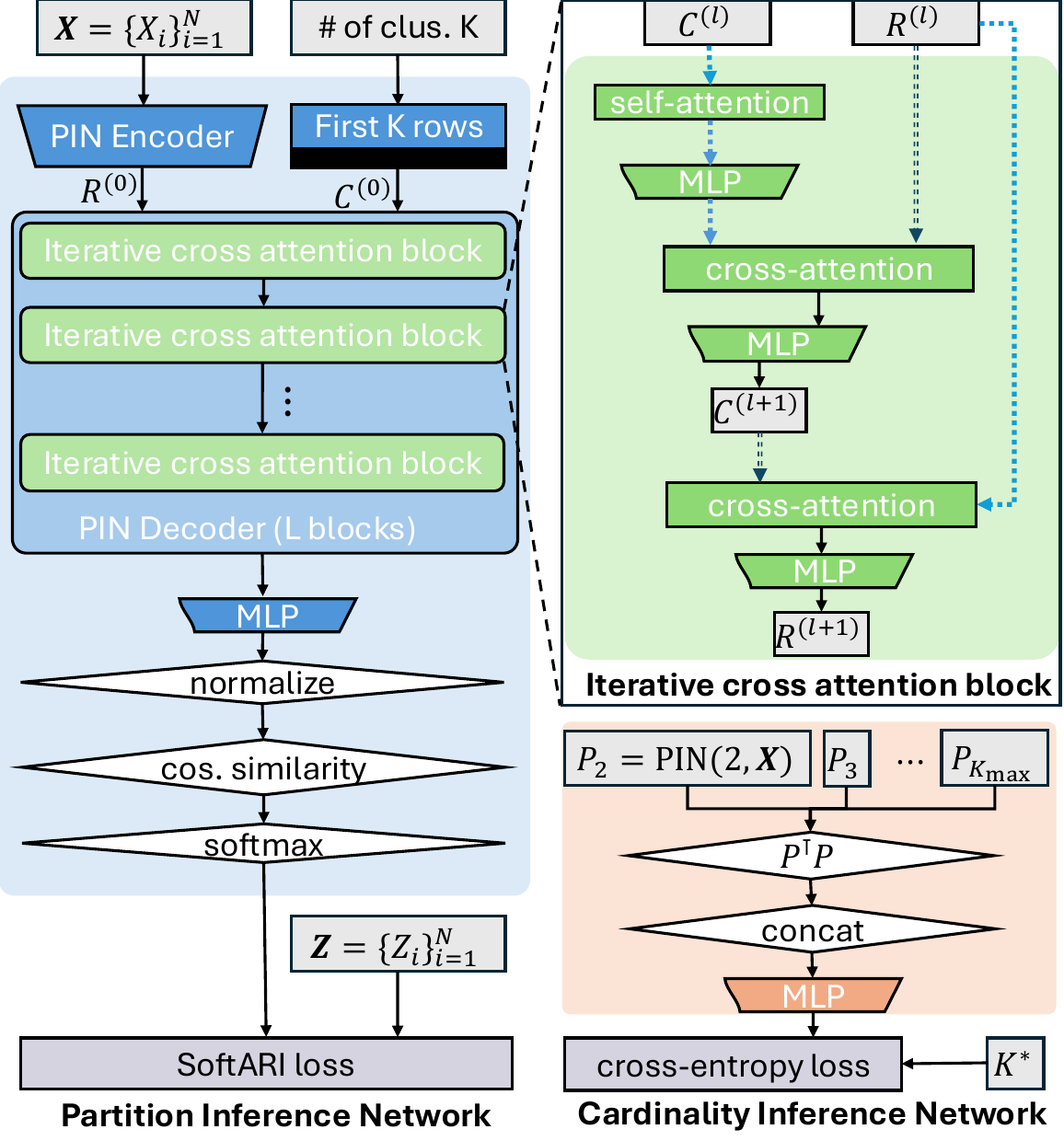}
\caption{\textbf{TabClustPFN architecture \& loss.}
Given dataset $\mX$ and candidate cluster number $K$, the PIN uses a transformer encoder followed by $L$ layers of iterative cross-attention between data representations and learnable cluster prototypes to produce soft assignments $P$.
Dotted blue lines denote query paths, and dashed arrows indicate keys and values.
Cosine similarity between normalized representations and prototypes yields cluster posteriors.
For $K\in\{2,\ldots,K_{\max}\}$, Gram matrices $P_K^\top P_K$ are summarized and fed to an MLP-based CIN to estimate $p(K|\mX)$.
PIN is trained with permutation-invariant SoftARI loss, while CIN uses cross-entropy on true $K^*$ under decoupled optimization.
}
\label{fig:architecture}
\vspace{-10pt}
\end{wrapfigure}
We consider a latent-variable model where each observation $X_i$ is generated independently from one of $K$ clusters, 
\begin{equation}
\label{eq:generate_model}
X_i|Z_i=k \sim f_k,  
\end{equation}
with latent assignments $Z_i \in [K] = \{1,\ldots,K\}$ and unknown cluster distributions $f_k$ for $K\geq 2$.
Given a dataset $\mX = \{X_i\}_{i=1}^{N}$, we perform Bayesian inference over both assignments $\mZ = \{Z_i\}_{i=1}^{N}$ and the cluster number $K$, approximating the joint posterior 
\[p(K, \mZ |\mX) = p(\mZ | K, \mX)\, p(K|\mX),\] 
which factorizes into a posterior over partitions conditioned on the cluster number $K$ and a marginal posterior over $K$.
\ours{} mirrors this factorization with two modules.
The \emph{Partition Inference Network} (PIN) approximates $p(\mZ | K, \mX)$ and performs amortized inference over cluster assignments for a given $K$.
The \emph{Cardinality Inference Network} (CIN) approximates $p(K|\mX)$ and predicts cluster number directly from the dataset.
Both components are implemented as prior-fitted networks that operate via in-context learning, enabling dataset adaptation without retraining.
Fig.~\ref{fig:architecture} overviews the architecture and loss functions of \ours{}, while Sec.~\ref{sec:ablation} provides ablations of the design choices. 
We describe each of the three components of our PFN-based method respectively below. 
All details related to pretraining are given in  App.~\ref{app:training_details}.

\subsection{Prior data}
\label{sec:prior_data}
The effectiveness of a PFN depends critically on the diversity and fidelity of its pretraining prior. 
Accordingly, we design a synthetic data-generating process (DGP) that follows~\eqref{eq:generate_model}, where each dataset is generated from latent cluster assignments $Z_i$ and cluster-specific distributions $f_{Z_i}$ (see App.~\ref{app:prior_data}). 
To ensure rich and challenging training, we define two complementary classes of cluster distributions $f_k$.
In the first class, we consider \textit{GMM-based clusters}. 
Each $f_k$ is a Gaussian, forming a standard GMM. 
To vary the difficulty of clustering across datasets, we explicitly control the degree of cluster overlap. 
For each dataset, we sample a target maximum pairwise overlap, measured via Bayes misclassification probability, so that some datasets contain well-separated clusters while others contain highly overlapping, more challenging clusters. 
Further diversity is achieved by randomizing mixture proportions and covariance structures, including spherical versus ellipsoidal shapes and homoscedastic versus heteroscedastic settings. 
In the second class, we consider \textit{iResNet-transformed clusters}. 
We generate $K$ latent Gaussian clusters and then apply a randomly sampled invertible iResNet~\citep{pmlr-v97-behrmann19a} transformation following ZEUS~\citep{marszaek2025zeus}. 
This bijective, nonlinear mapping produces complex, non-Gaussian cluster shapes while preserving cluster identity and the number of components. 
Categorical variables are incorporated following the same ZEUS procedure. 
We call this the \emph{ZEUS prior} in short.
Together, these two mechanisms ensure that pretraining exposes the PFN to both standard parametric clusters and more complex, highly structured cluster geometries.

\subsection{Network architecture}
\label{sec:network}
\textbf{Transformer based \pin}. 
We implement \pin{} as a Transformer based PFN mapping a dataset $\mX$ and cluster cardinality $K$ to a soft partition $\pin_{\theta}: (\mX, K)\mapsto P \in [0,1]^{N \times K}$, $P_{ik} = \hat{p}(Z_i=k|K,\mX)$. 
\pin{} follows an encoder–decoder design: the encoder maps $\mX$ to representations in $\mathbb{R}^d$, while the decoder refines them using learnable cluster prototypes.
For each $K$, the decoder selects the first $K$ prototypes from a shared $K_{\max}\times d$ matrix and applies $L$ layers of prototype self-attention followed by bidirectional cross-attention between prototypes and data, enabling iterative co-refinement.
\emph{Crucially, both the class prototypes and the input samples are permutation-invariant with respect to their row ordering.} 
Motivated by Set Transformer~\citep{lee2019set}, all attention blocks in the \pin{} decoder are therefore implemented without positional encodings.
Finally, normalized data representations and prototypes are compared via cosine similarity, and a softmax over clusters yields assignment probabilities. 
This realizes a fully amortized, permutation-invariant analogue of centroid-based clustering.
We adopt the TabICL encoder~\citep{qu2025tabicl} (column-wise embedding followed by row-wise interaction) for reproducibility, though the framework is agnostic to the specific tabular encoder.


\textbf{MLP based \cin{}}.
Rather than operating on the raw dataset, \cin{} leverages the partition structure inferred by \pin{} across candidate values of $K$.
For each $K \in \{2, \ldots, K_{\max}\}$, we compute the corresponding soft assignment matrix $P_K = \pin_{\theta}(\mX, K)$ and form the Gram matrix $G_K = P_K^\top P_K$, which summarizes pairwise cluster co-assignment statistics.
The lower-triangular entries of $G_K$ are vectorized and concatenated across all candidate $K$, forming the input to a two-layer perceptron $\cin{}_{\phi}$.
To ensure invariance to permutations of cluster labels, we impose a canonical ordering: the diagonal elements and off-diagonal elements are sorted separately in descending order. 
This separation is necessary because the two groups encode fundamentally different quantities — diagonal entries correspond to cluster self-mass (soft cluster sizes), whereas off-diagonal entries represent inter-cluster co-assignment strength. 
Sorting them jointly would mix these semantically distinct statistics and distort their structural roles, while separate sorting preserves their meaning and removes dependence on arbitrary cluster index assignments.
The network outputs a $(K_{\max}-1)$-dimensional vector giving an estimated posterior over cluster cardinalities.
We exclude $K=1$, focusing on datasets that contain at least two latent clusters.



\subsection{Loss functions}
\label{sec:loss}
\textbf{Loss for \pin{}}. 
Clustering is invariant to permutations of cluster labels, so the training objective for \pin{} must not depend on their indexing. 
Standard supervised losses violate this requirement: identical partitions differing only by label permutations incur large loss, producing spurious signals and destabilizing training. 
A natural choice is the Adjusted Rand Index (ARI)~\citep{hubert1985comparing}, a widely used measure of agreement based on pairwise co-assignment. 
ARI is permutation-invariant and aligns with standard evaluation. 
However, it is defined over hard assignments via an $\argmax$ and is therefore non-differentiable, making it incompatible with gradient-based optimization.

We use \emph{SoftARI}~\citep{flynt2019sari}, a differentiable relaxation that replaces hard assignments with the model’s predicted cluster probabilities and computes soft pairwise co-assignment scores.
Let $Z_{ik} = \mathbbm{1}(Z_i = k)$ be the one-hot encoding of the ground-truth cluster assignment, which is available for synthetic pretraining datasets.
SoftARI computes soft pairwise co-assignment scores using $P_{ik}$ and $Z_{ik}$, yielding a fully differentiable and permutation-invariant objective:
\[
\mathrm{SoftARI}(\pin_{\theta}(K,\mX),\mZ)
=\frac{
\sum_{k,\ell} \binom{n_{k\ell}}{2}
-\{\sum_k \binom{n_{k\cdot}}{2}
\sum_\ell \binom{n_{\cdot\ell}}{2}\}/
\binom{n}{2}
}{
\left\{
\sum_k \binom{n_{k\cdot}}{2}
+
\sum_\ell \binom{n_{\cdot\ell}}{2}
\right\}/2
-
\{\sum_k \binom{n_{k\cdot}}{2}
\sum_\ell \binom{n_{\cdot\ell}}{2}\}/
\binom{n}{2}
}    
\]
where $n_{k\ell} = \sum_{i=1}^n P_{ik} Z_{i\ell}$, $n_{k\cdot} = \sum_{\ell} n_{k\ell}$, and $n_{\cdot\ell} = \sum_{k} n_{k\ell}$.
The loss for \pin{} is defined as
\begin{equation}
\label{eq:pin_loss}
\mathcal{L}_{\pin{}}(\theta) = - \mathrm{SoftARI}(\pin_{\theta}(K^*,\mX),\mZ),    
\end{equation}
where $K^*$ is the true cluster number, available during pretraining on synthetic datasets (Sec.~\ref{sec:prior_data}).

\textbf{Loss for \cin{}}. 
Since the true cluster number $K^*$ is available from the DGP, we can supervise \cin{} using the standard cross-entropy loss over the candidate set $\{2, \ldots, K_{\max}\}$.
Let $\cin{}_{\phi}(\mX) = \{\hat{p}(K | \mX)\}_{K=2}^{K_{\max}}$ denote the predicted posterior.
The loss is 
\begin{equation}
\label{eq:cin_loss}
\mathcal{L}_{\cin{}}(\phi) = - \log \hat{p}(K^* | \mX).    
\end{equation}

\subsection{Model pretraining}
Following the TabPFN paradigm, we pretrain the model on synthetic datasets sampled from the DGP described in Sec.~\ref{sec:prior_data}. 
We initialize the encoder with pretrained TabICL weights~\citep{qu2025tabicl} and randomly initialize the remaining parameters.  
At each optimization step, we draw a batch of 512 synthetic tasks (datasets). 
For each dataset, we sample the number of cluster $K\in[2,10]$, the number of samples $N\in[500,1000]$ and features $D\in[2,64]$, sample a distribution over $\mathbb{R}^D$ from the prior, then generate the data and ground-truth cluster labels from this distribution.
Each dataset is generated on-the-fly and never reused.
We adopt a \emph{decoupled optimization scheme} when optimizing the \pin{} loss in~\eqref{eq:pin_loss} and the \cin{} loss in~\eqref{eq:cin_loss}. Specifically, CIN operates on detached PIN outputs, and its gradients are not backpropagated through PIN. 
This decoupling mitigates interference between partition inference and cardinality estimation, leading to improved performance (Tab.~\ref{tab:decoupled_loss_ablation}).
All parameters are optimized with AdamW using a cosine learning-rate schedule with $2,000$ warm-up steps and a peak learning rate of $10^{-4}$. 
Training runs for $10,000$ optimization steps on four RTX $5090$ GPUs, yielding a favorable performance–compute trade-off at a cost of $\approx 92$ GPU hours.

%% file: sections/exp.tex
\section{Empirical results}
\label{sec:experiment}
We evaluate the performance of our pretrained TabClustPFN on both unseen synthetic datasets and a curated real-world benchmark.
We first describe the baseline methods and the evaluation metric.

\begin{wraptable}{r}{0.45\textwidth}
\centering
\vspace{-12pt}
\caption{\textbf{Performance on synthetic datasets}. Median ARI/NMI ranks and $k$-MAE ($\downarrow$). Brackets show IQR. Bold and underline indicate best and second-best per column.
}
\label{tab:autok_gen_rank_comparison}
\resizebox{\linewidth}{!}{
\begin{tabular}{lccc}
\toprule
Method & ARI Rank [IQR] $\downarrow$ & NMI Rank [IQR] $\downarrow$ & $k$-MAE [IQR] $\downarrow$ \\
\midrule
TabClustPFN        & \textbf{2 [4]} & \textbf{3 [5]} & \textbf{0 [1]} \\
\midrule
AP                 & 14 [1]         & 13 [3]         & 44 [12]        \\
Agg$^+$            & 9 [4]          & 10 [4]         & \underline{1 [4]}          \\
Birch$^+$          & 9 [3]          & 10 [3]         & \underline{1 [4]}          \\
DBSCAN             & 16 [2]         & 17 [1]         & 5 [5]          \\
GMM$^+$            & 6 [4]          & 6 [5]          & \underline{1 [3]}          \\
HDBSCAN            & 15 [4]         & 15 [1]         & 3 [4]          \\
KMeans$^+$      & 5 [5]          & 6 [4]          & \underline{1 [3]}          \\
MeanShift          & 15 [2]         & 16 [2]         & 4 [5]          \\
OPTICS             & 16 [2]         & 16 [1]         & 4 [6]          \\
Spectral$^+$       & 14 [5]         & 14 [5]         & 2 [5]          \\
\midrule
T+Agg$^+$          & 5 [7]          & \underline{5 [7]} & \underline{1 [3]}          \\
T+GMM$^+$          & 5 [4]          & \underline{5 [5]} & \underline{1 [3]}          \\
T+KMeans$^+$       & \underline{4 [5]} & \underline{5 [5]} & \underline{1 [4]}          \\
T+Spec$^+$         & 5 [5]          & \underline{5 [4]} & \underline{1 [3]}          \\
\midrule
Zeus+GMM$^+$       & 10 [6]         & 9 [7]          & 6 [5]          \\
Zeus+KMeans$^+$    & 9 [7]          & 9 [8]          & 6 [5]          \\
Zeus+S-GMM$^+$     & 8 [6]          & 8 [7]          & 6 [5]          \\
\bottomrule
\end{tabular}}
\vspace{-15pt}
\end{wraptable}
\textbf{Baselines}.
We evaluate \ours{} against a wide range of classical and modern clustering algorithms, covering scenarios with known and unknown cluster numbers. 
This includes all methods available in \texttt{scikit-learn} at the time of writing, as well as two PFN-based approaches: ZEUS+X~\citep{marszaek2025zeus} and T+X~\citep{li2025clustering}, where X denotes a classical clustering method used in an intermediate step. 
A full list of method names and their acronyms used in the results is provided in App.~\ref{app:baselines}. 
Methods marked with $^{*}$ indicate oracle knowledge of the cluster number, while those with $^{+}$ select it by maximizing the silhouette score. 
Methods without a suffix infer the number of clusters automatically.
Since silhouette-based selection may not align with every method's clustering assumptions, we report additional results using method-specific criteria and hyperparameter search for DBSCAN, GMM, and Spectral (App.~\ref{app:median_ari_nmi_group_by_difficulty}).
\emph{We exclude Cluster-PFN~\citep{bhaskaran2025transformer} due to its different setting and instead compare it within its intended regime in App.~\ref{app:cluster-pfn}}.
Cluster-PFN is not even comparable to GMM when $D\geq 3$.
This results in $13$ baselines with known $K$ and $18$ with selected $K$.

\textbf{Performance metric}.
We assess clustering quality using standard metrics: ARI and Normalized Mutual Information (NMI).  
Because raw ARI and NMI values can vary substantially across datasets, we report rank-based metrics in the main text.  
Each method is ranked independently per dataset, and we report the median and IQR of the ranks. 
We use median (IQR) rather than mean (std) for robustness to outliers.
When the cluster number is unknown, we additionally evaluate cluster cardinality estimation using the median absolute error ($k$-MAE) between the predicted and true cluster number.  
Other metrics, including median ARI and NMI values, are provided in App.~\ref{app:median_ari_nmi_group_by_difficulty}.

\subsection{Generalization on synthetic datasets}
We first evaluate TabClustPFN on $49$ synthetic datasets, comprising $25$ GMM-generated and $24$ ZEUS-generated datasets.  
All datasets are held out from pretraining.  
Detailed statistics are provided in Tab.~\ref{tab:synthetic_stats_gmm} and Tab.~\ref{tab:synthetic_stats_zeus} in App.~\ref{sec:dataset_stats}.  
Results are reported in Tab.~\ref{tab:autok_gen_rank_comparison}, which shows the ranks of $18$ methods that select $K$. Since the number of datasets is odd, the reported medians of rank-based metrics take integer values. 
Ranks for methods with known $K$ are provided in App.~\ref{app:synthetic_rank_true_k}.  
Across all metrics and methods, TabClustPFN significantly outperforms other methods, achieving the SOTA performance.

\textbf{A closer look at the model}.
To understand the representations learned by our model, we visualize them on representative synthetic datasets in 
\begin{figure}[htbp]
\centering
\begin{minipage}[t]{0.47\textwidth}
\centering
\includegraphics[width=0.24\linewidth]{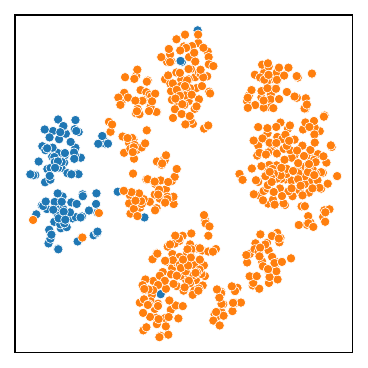}
\includegraphics[width=0.24\linewidth]{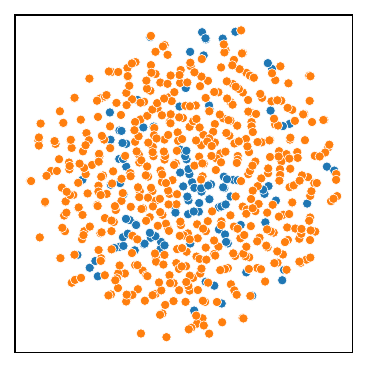}
\includegraphics[width=0.24\linewidth]{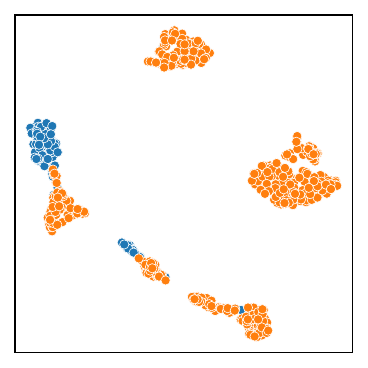}
\includegraphics[width=0.24\linewidth]{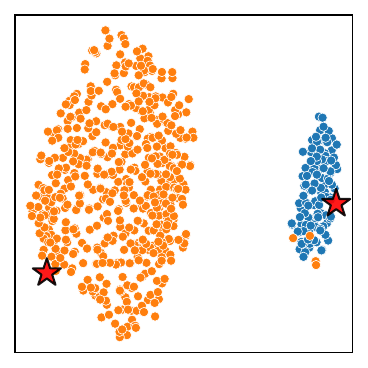}\\
\includegraphics[width=0.24\linewidth]{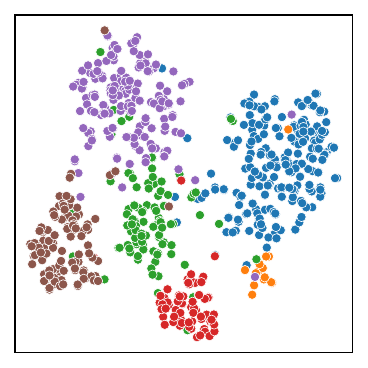}
\includegraphics[width=0.24\linewidth]{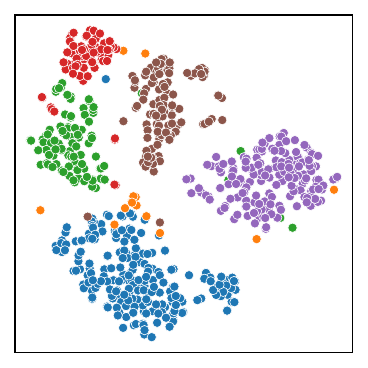}
\includegraphics[width=0.24\linewidth]{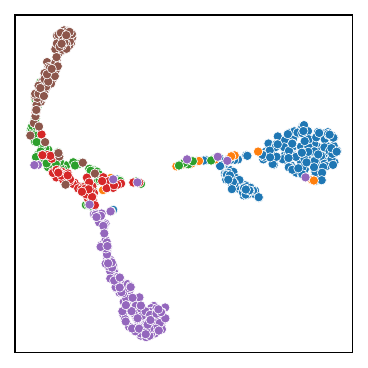}
\includegraphics[width=0.24\linewidth]{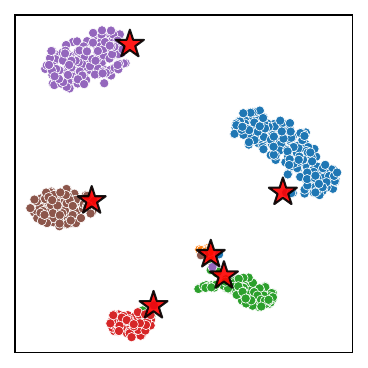}\\
\includegraphics[width=0.24\linewidth]{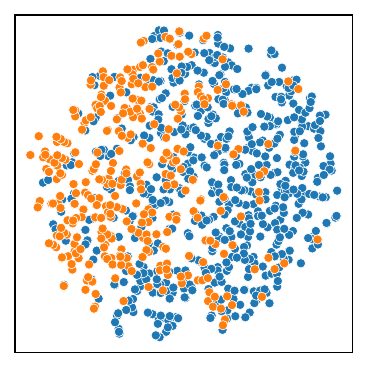}
\includegraphics[width=0.24\linewidth]{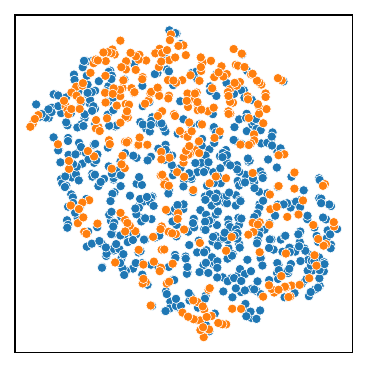}
\includegraphics[width=0.24\linewidth]{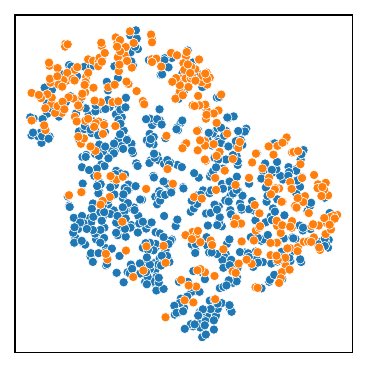}
\includegraphics[width=0.24\linewidth]{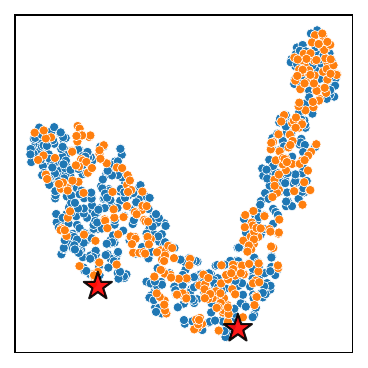}
\captionof{figure}{\textbf{Representation $t$-SNE}. Columns: raw features, TabICL encoder, ZEUS, TabClustPFN. Rows: increasing difficulty.}
\label{fig:representation_visualization}
\end{minipage}
\hfill
\begin{minipage}[t]{0.47\textwidth}
\centering
\includegraphics[width=0.24\linewidth]{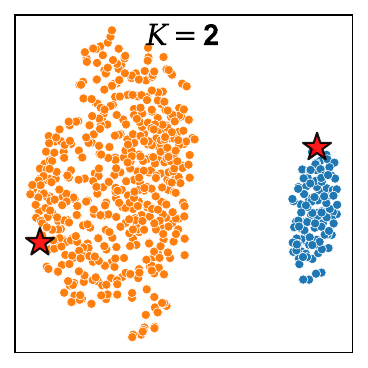}
\includegraphics[width=0.24\linewidth]{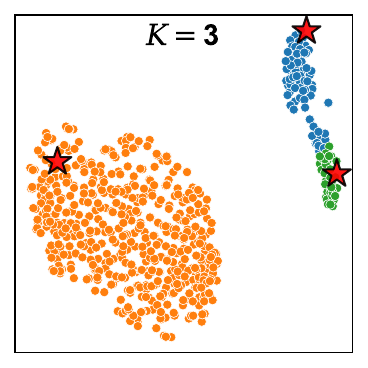}
\includegraphics[width=0.24\linewidth]{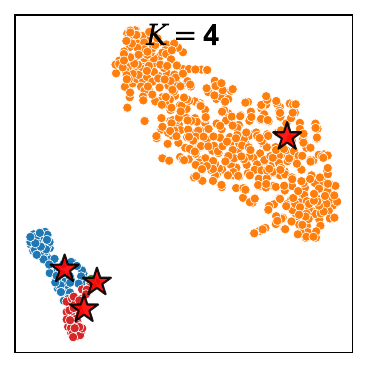}
\includegraphics[width=0.24\linewidth]{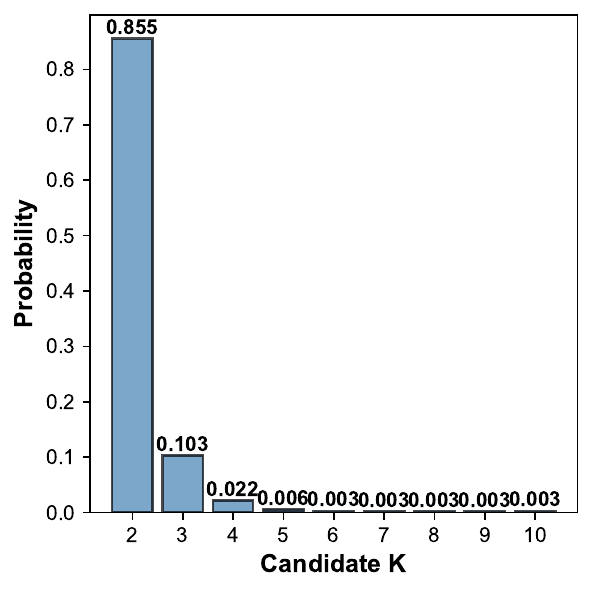}\\
\includegraphics[width=0.24\linewidth]{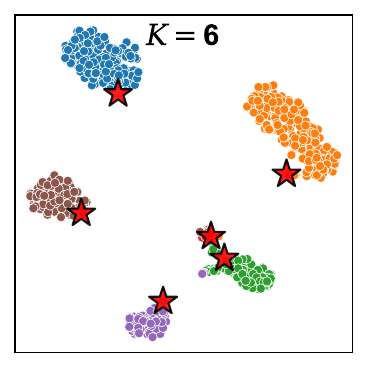}
\includegraphics[width=0.24\linewidth]{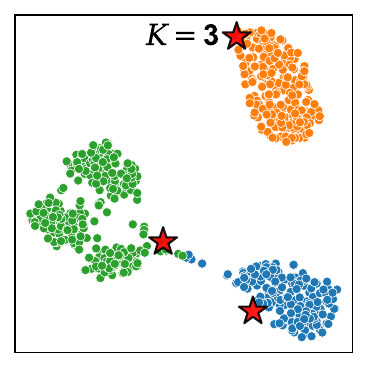}
\includegraphics[width=0.24\linewidth]{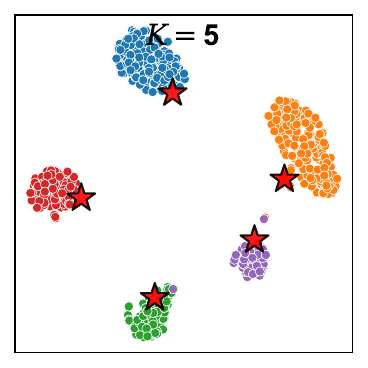}
\includegraphics[width=0.24\linewidth]{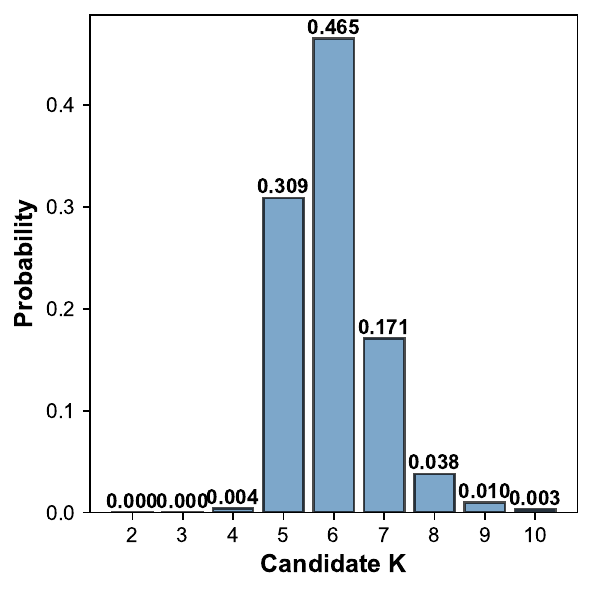}\\
\includegraphics[width=0.24\linewidth]{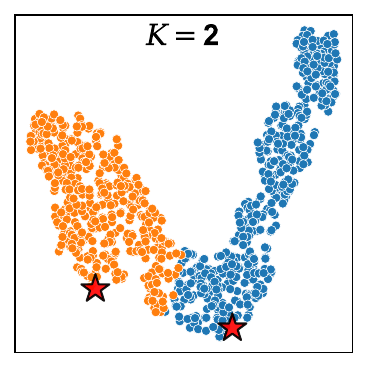}
\includegraphics[width=0.24\linewidth]{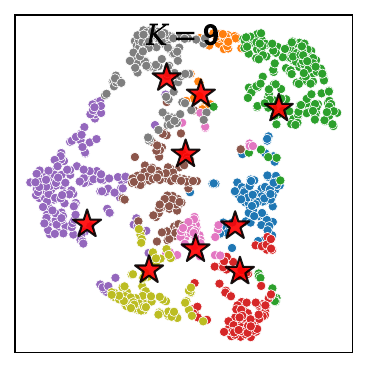}
\includegraphics[width=0.24\linewidth]{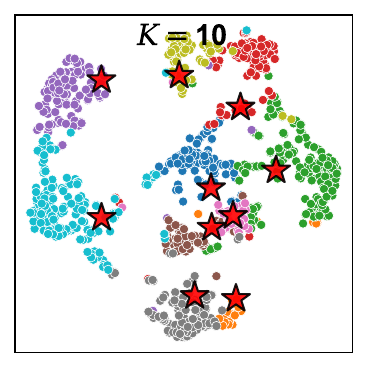}
\includegraphics[width=0.24\linewidth]{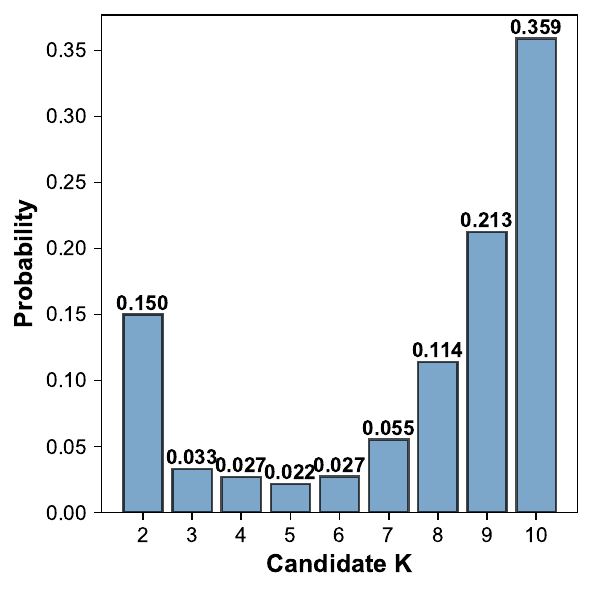}
\captionof{figure}{\textbf{PIN representations \& CIN posteriors}. Columns show $R^{(L)}$, other $K$, and $\hat p(K|\mathbf X)$.}
\label{fig:pin_cin_output}
\end{minipage}
\vspace{-10pt}
\end{figure}
Fig.~\ref{fig:representation_visualization}, comparing raw features, pretrained TabICL embeddings, ZEUS representations, and \ours{}. 
Visualizations for the full synthetic sets are provided in App.~\ref{app:representation_visualization}.
Our method has strong performance on most of the synthetic datasets.
The 1st and 2nd rows correspond to datasets where TabClustPFN outperforms ZEUS, while the last row shows a case where both methods fail. 
For our TabClustPFN, red stars denote cluster prototypes $C^{L}$ and points correspond to final representations $R^{L}$ after the last MLP layer and normalization. 
Points are colored by ground-truth labels.
In the last row, the two clusters are highly overlapping, making the dataset intrinsically difficult regardless of the clustering algorithm. 
In the 1st row, $t$-SNE applied to raw features, TabICL embeddings, and ZEUS representations fails to reveal clear cluster structure. 
In contrast, TabClustPFN produces well-separated, blob-shaped groups, with prototypes located near cluster centers. 
This indicates that TabClustPFN reshapes the data into a geometry where cluster structure becomes explicit and aligned with prototype directions.
In contrast, ZEUS representations often exhibit elongated, anisotropic cluster shapes. 
This can cause silhouette-based model selection to choose an incorrect number of clusters, as the metric implicitly favors compact, blob-like groups. 
We also observe that ZEUS performs well when the raw feature space already exhibits clear cluster structure (2nd row), but struggles when clusters are not well shaped in the input space (1st row).

We visualize the decoder representations $R^{(L)}$ produced by \pin{} under different candidate values of $K$, together with the posterior over $K$ predicted by \cin{} (Fig.~\ref{fig:pin_cin_output}). 
For the easiest case (1st row), \cin{} outputs a sharp posterior concentrated on a single $K$, and the representation shows compact, well-separated clusters. 
As difficulty increases (3rd row), the posterior becomes more diffuse, reflecting greater uncertainty in $K$. 
We also observe that $K$ values with higher $\hat{p}(K|\mX)$ yield more coherent and compact cluster geometry, while low-probability $K$ leads to fragmented or less structured layouts. 
This alignment between representation quality and posterior mass indicates that CIN captures structural cues from PIN when estimating the cluster cardinality.

\begin{wraptable}{r}{0.45\textwidth}
\centering
\vspace{-12pt}
\caption{\textbf{Calibration of the CIN posterior}.}
\label{tab:cardinality-calibration}
\vspace{-2pt}
\resizebox{\linewidth}{!}{
\begin{tabular}{lccccc}
\toprule
Nominal $\alpha$ & 80\% & 85\% & 90\% & 95\% & 99\% \\
\midrule
Empirical coverage & 0.875 & 0.907 & 0.937 & 0.968 & 0.993 \\
\bottomrule
\end{tabular}}
\vspace{-12pt}
\end{wraptable}
\textbf{Calibration}.
We evaluate the calibration of the cardinality posterior $\hat p(K | \mX)$ predicted by CIN (see App.~\ref{app:calibration}). 
We do not evaluate $\hat p(\mZ | K, \mX)$, as calibration of cluster assignments is not well-defined under label switching. Standard calibration methods require a fixed correspondence between predicted and ground-truth labels; while one could apply optimal matching (\eg Hungarian algorithm), this introduces a dataset-level dependency and does not yield a principled notion of per-sample correctness. 
Hence, calibration lacks a clean definition in the clustering setting.
Tab.~\ref{tab:cardinality-calibration} shows that the empirical coverage consistently exceeds the nominal level across all tested values of $\alpha$. 
Thus, the CIN posterior is not overconfident: prediction sets formed from $\hat p(K|\mX)$ reliably cover the true cluster cardinality. 
Moreover, the empirical coverage is close to the target at high confidence levels. 
These results suggest that \ours{} provides reliable uncertainty estimates for cluster cardinality, which is important in the unknown-$K$ setting where overconfident cardinality predictions can directly degrade downstream clustering quality.

\begin{wrapfigure}{r}{0.45\textwidth}
\centering
\vspace{-5pt}
\includegraphics[width=\linewidth]{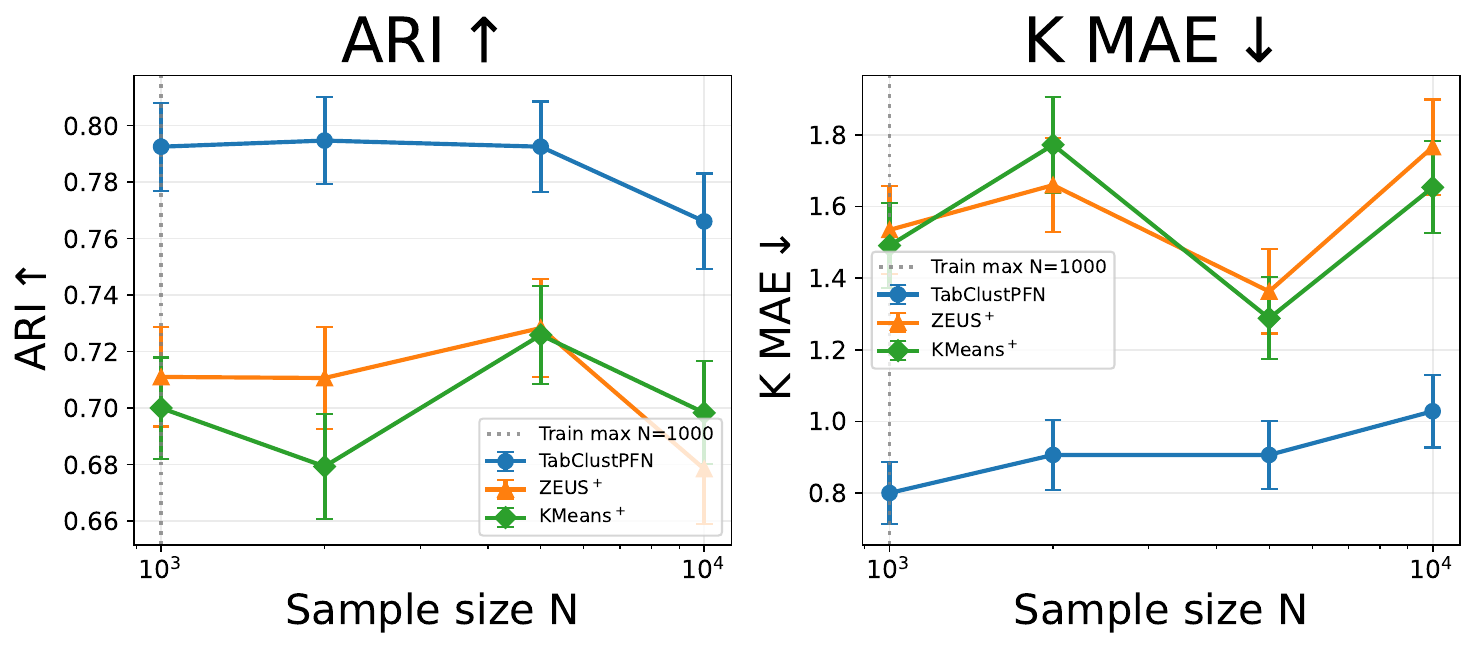}
\caption{\textbf{OOD performance under larger $N$} (fixed $D=50$).}
\label{fig:scale_ood_N}

\includegraphics[width=\linewidth]{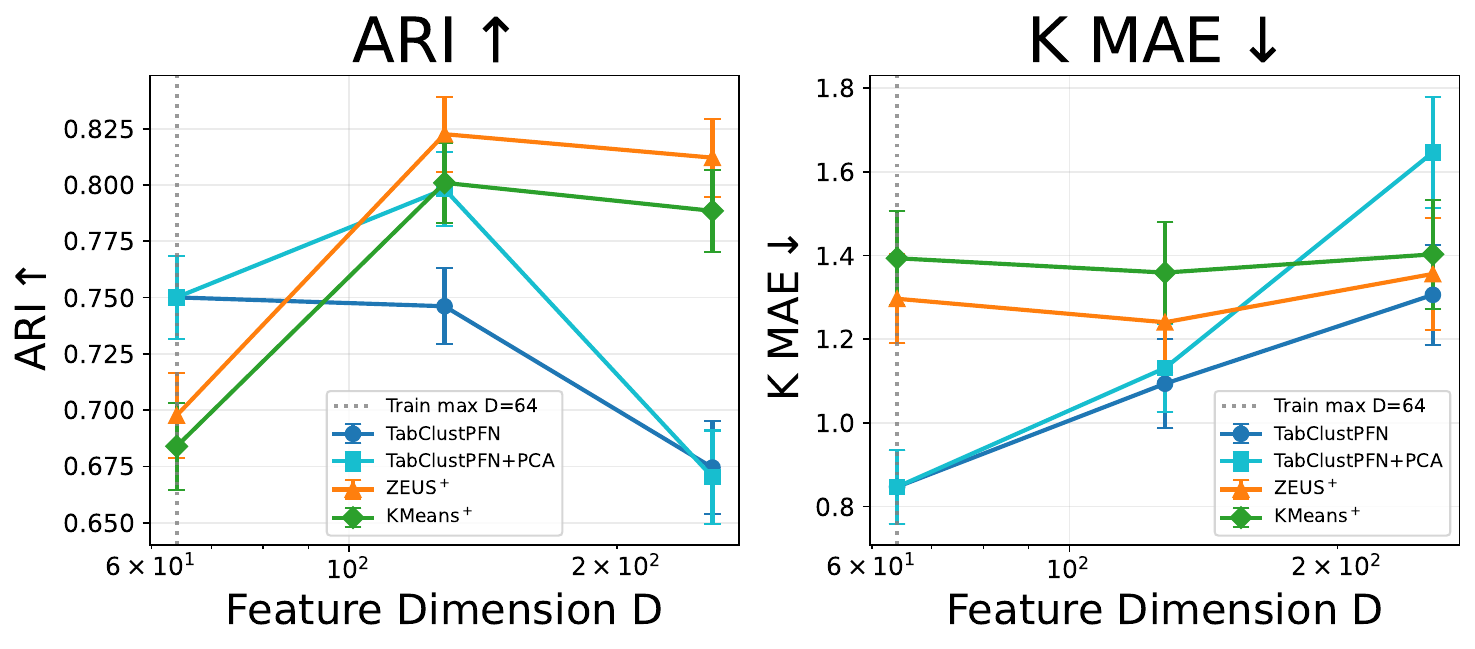}
\caption{\textbf{OOD performance with larger $D$}.}
\label{fig:scale_ood_D}

\vspace{-10pt}
\end{wrapfigure}

\textbf{Generalization for larger $N$ \& $D$}.
We further evaluate out-of-distribution scaling beyond the pretraining range ($N \leq 1000$, $D \leq 64$), comparing against ZEUS$^{+}$ and KMeans$^{+}$, the strongest PFN-style and classical baselines in our benchmark. 
TabClustPFN generalizes strongly to larger sample sizes, maintaining the best ARI and the lowest $k$-MAE up to $N=10^4$ (Fig.~\ref{fig:scale_ood_N}). 

For larger feature dimensions, direct extrapolation is more challenging, but PCA provides a simple remedy in the moderate-OOD regime, where TabClustPFN+PCA achieves comparable performance by mapping inputs back closer to the pretraining support (Fig.~\ref{fig:scale_ood_D}). 
At higher dimensions, the main degradation comes from less accurate cardinality estimation after PCA:
the projection changes the global structure used by CIN, increasing $k$-MAE, and errors in the selected $K$ directly affect unknown-$K$ ARI. 
This suggests that the current gap is mainly due to limited exposure to larger-dimensional data, and can be further reduced with continued pretraining.

\textbf{Robustness to distribution shifts}. We additionally evaluate the frozen checkpoint on cluster-size imbalance up to $100{:}1$ and Student-$t$ subpopulations. At $D=2$, TabClustPFN maintains strong ARI under severe imbalance and remains competitive for degrees of freedom $\nu\geq3$. At higher dimensions, median ARI across conditions exceeds $0.99$ under both imbalance and heavy-tailed shifts, with detailed results provided in App.~\ref{app:ood_synthetic_shifts}.

\subsection{Real-world clustering}
We further evaluate TabClustPFN on a curated real-world benchmark to assess practical performance.

\textbf{Benchmark}. The benchmark contains datasets from multiple widely used sources, including OpenML-CC18~\citep{openmlsuite}, TabArena~\citep{erickson2025tabarena}, the AutoML benchmark suite~\citep{automl}, the ZEUS evaluation datasets, and a UCI clustering collection~\citep{milaan9_clustering_datasets}. 
Duplicates across collections are removed, and synthetic datasets are excluded to ensure evaluation reflects generalization to truly unseen, real-world data rather than tasks similar to pretraining.
To match our model regime, we retain datasets with at most $10$ clusters, $64$ features, and $1,000$ samples. 
This leads to $22$ datasets from OpenML and $22$ from UCI (App.~\ref{sec:dataset_stats}).

\begin{wrapfigure}{r}{0.45\textwidth}
\centering
\vspace{-13pt}
\includegraphics[width=0.47\linewidth]{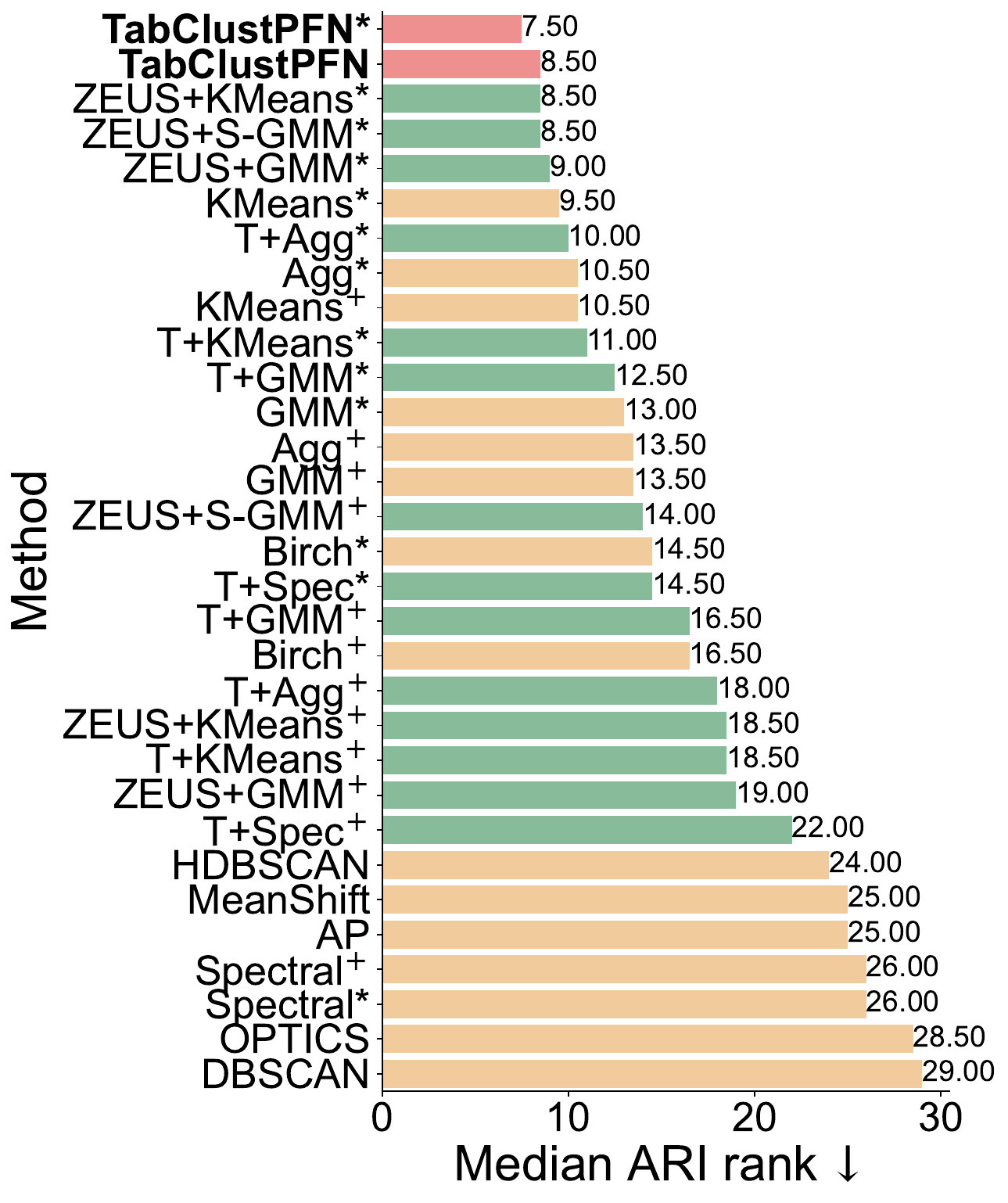}
\includegraphics[width=0.47\linewidth]{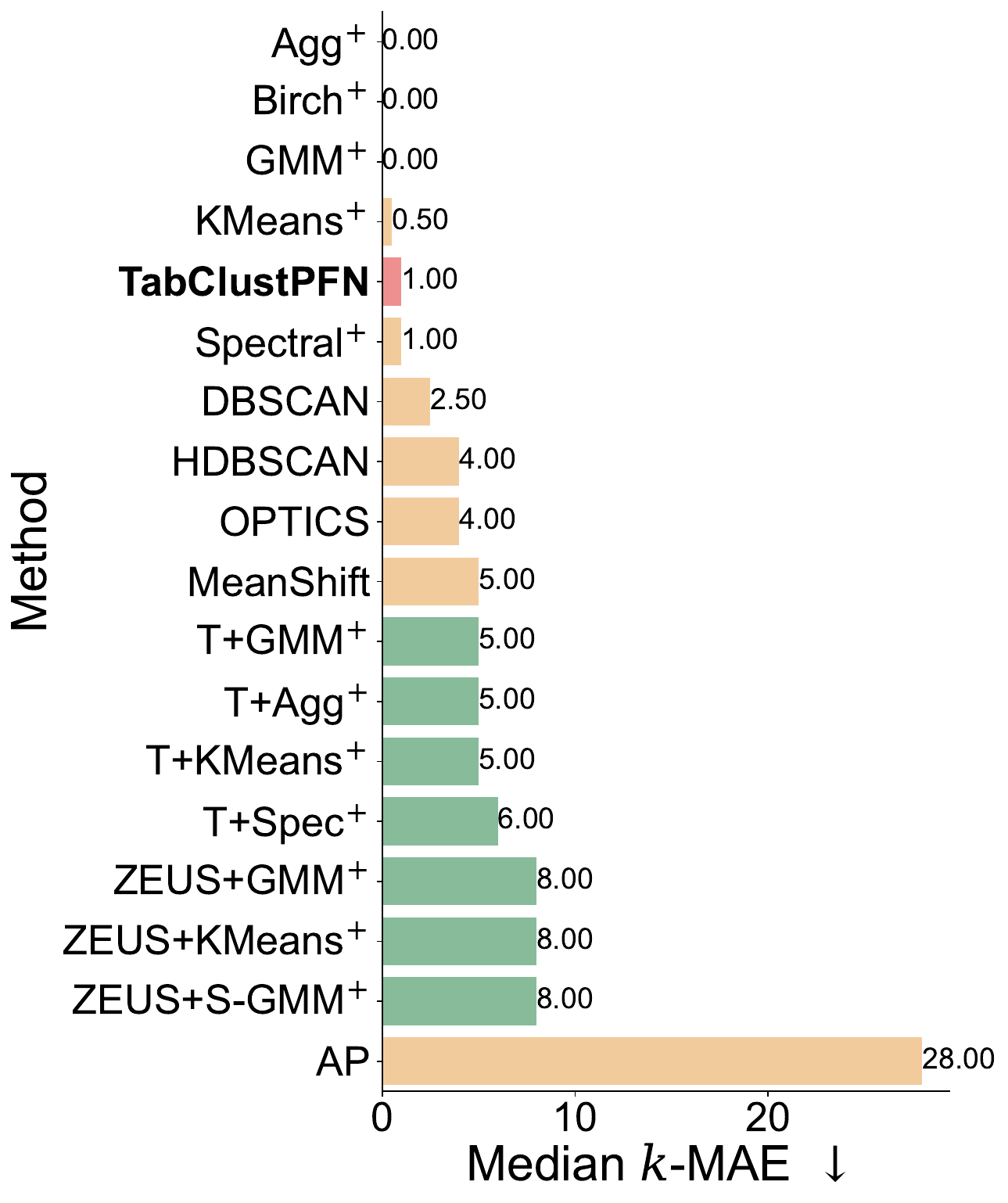}
\caption{\textbf{Benchmark performance on real-world benchmark}. Median of ARI rank (left) and MAE in the estimated number of clusters (right); lower is better for both metrics. \textcolor[HTML]{eb7575}{Red} bar denotes our \ours{}, \textcolor[HTML]{6aaa81}{green} bars denote other PFN-based methods, and \textcolor[HTML]{efbf85}{yellow} bars denote classical methods.}
\label{fig:main_exp}
\vspace{-15pt}
\end{wrapfigure}
\textbf{Results}. The median ARI rank across $31$ methods under both known-$K$ and unknown-$K$ settings is in Fig.~\ref{fig:main_exp} (left).
As expected, providing the true number of clusters improves performance. 
TabClustPFN with known $K$ achieves the best overall performance, while the variant inferring $K$ via CIN ranks second. 
Notably, \ours{} with inferred $K$ performs comparably to ZEUS followed by KMeans even when the latter is given the true $K$ in its second stage, highlighting the effectiveness of our end-to-end inference.
Fig.~\ref{fig:main_exp} (right) shows the $k$-MAE for methods that estimate $K$. 
\ours{} attains a $k$-MAE of about $1$, lower than other PFN-based approaches, highlighting the effectiveness of CIN for cardinality estimation.
We further compare with DP-means~\citep{kulis2012revisiting} and Dip-means~\citep{kalogeratos2012dipmeans}, two additional methods without a fixed upper bound on $K$, and find that TabClustPFN achieves higher median ARI and NMI on both benchmarks despite CIN's bounded support (App.~\ref{app:adaptive_k_baselines}).
More detailed comparisons are provided in App.~\ref{app:pairwise_win_rate}--\ref{app:rank_group_by_knowledge_on_k}. 
Pairwise win rates show that \ours{} is robust across real-world datasets. 
\emph{We assess statistical significance using paired $t$-tests and Wilcoxon signed-rank tests, which show that TabClustPFN significantly outperforms all baselines on synthetic data and all but Agg$^+$ and KMeans$^+$ on real-world data, where it still achieves higher ARI on most datasets.}
We further report critical-difference diagrams based on average ARI ranks, which consistently rank \ours{} first on both benchmarks (App.~\ref{app:significance}).

\textbf{Real-world OOD evaluation}. Applying the main benchmark's dataset inclusion criteria (App.~\ref{sec:dataset_stats}) within the OOD regime of $64<D\leq500$, $N<3000$, and $2\leq K\leq10$ yields 6 additional real-world datasets. We evaluate the frozen checkpoint on all 6 datasets without retraining or fine-tuning. We compare against four representative baselines spanning centroid-based, hierarchical, density-based, and PFN-based clustering. TabClustPFN achieves the highest median ARI/NMI and lowest median cardinality error. Dataset identities, and details are provided in App.~\ref{app:ood_real_world}.

\subsection{Comparison with amortized clustering}
In the main experiments, we compare primarily with classical clustering algorithms and PFN-based methods, but not with existing amortized clustering approaches such as DAC~\citep{lee2019deep} and NCP~\citep{pakman2020neural}. 
The reason is twofold. 
First, these models are typically trained under a fixed data-generating family and often generalize poorly beyond that distribution. 
Second, scaling their training to the diversity and size of datasets considered in our broader benchmarks is computationally intensive.
Nevertheless, to understand how our approach compares within the regime these methods are designed for, we follow their original evaluation protocol and conduct a controlled comparison under the same setting. 
Datasets are generated from a \textbf{2D GMM}-based process: each dataset $\gD$ is formed by sampling the number of clusters $K$, mixture weights $\pi$, and component parameters $\{(\mu_k, \Sigma_k)\}_{k=1}^K$, and then drawing instances from the resulting GMM.
Both training and test datasets are independent draws from this same distribution family (see App.~\ref{app:amortized_clustering} for details).

\begin{wraptable}{r}{0.45\textwidth}
\centering
\vspace{-12pt}
\caption{\textbf{Quantitative comparison of clustering performance and efficiency under the evaluation setting used by amortized clustering methods}. Results are reported in terms of median ARI, $k$-MAE, $k$-Acc, and clustering time. Bold values indicate the best result for each metric ($\uparrow$ higher is better, $\downarrow$ lower is better).}
\label{tab:performance_comparison}
\resizebox{\linewidth}{!}{
\begin{tabular}{lcccc}
\toprule
Method & ARI $\uparrow$ & $k$-MAE $\downarrow$ & $k$-Acc $\uparrow$ & Time (ms) $\downarrow$ \\
\midrule
DAC (Anc)            & 0.5665 & 10.0 & 0.00625 & 257.20 \\
DAC (Min)            & 0.5209 & 5.0 & 0.0375 & 235.48 \\
NCP                 & 0.6000 & 3.0 & 0.08125 & 1058.04 \\
PFN (Random-init) & 0.4222 & 1.0 & \textbf{0.4250} & \textbf{69.58} \\
PFN (TabICL-init) &0.6294 & 1.0 & \textbf{0.4250} & 71.79  \\
TabClustPFN   & \textbf{0.7893} & \textbf{1.0} & \textbf{0.4250} & 70.93 \\
\bottomrule
\end{tabular}}
\vspace{-15pt}
\end{wraptable}
Following DAC and NCP, we train all amortized baselines on $16{,}000$ synthetic training datasets with at most $10$ clusters (each used once), and evaluate on $160$ held-out datasets from the same GMM process. 
For our method, we report three models: (i) the TabClustPFN foundation model pretrained under our full prior, and two variants trained only on this GMM data: (ii) PFN (Random-init), trained from scratch without TabICL initialization, and (iii) PFN (TabICL-init), initialized with TabICL encoder weights. Results are summarized in Tab.~\ref{tab:performance_comparison}.

TabClustPFN achieves the best overall performance, showing that the PFN paradigm yields a stronger amortized clustering model even in the narrow setting tailored to previous amortized methods. 
PFN (TabICL-init) also outperforms all amortized clustering baselines and improves over PFN (Random-init), indicating that stronger initialization facilitates optimization even when the data distribution is simple and homogeneous. 
With larger-scale pretraining, the performance gap of the randomly initialized model is expected to shrink. 
In addition, our method provides a $4\times$ speedup compared to the fastest baseline.

%% file: sections/discussion.tex
\section{Discussion on design choices}
\label{sec:ablation}

\textbf{Effectiveness of iterative cross attention}.
We evaluate whether the decoder architecture drives the performance gains of TabClustPFN by keeping the encoder and training setup fixed while comparing our iterative cross-attention decoder against two non-iterative baselines.
The first baseline is a \emph{na\"ive SA+MLP decoder}, which directly predicts cluster assignments from point embeddings: it applies self-attention over the $N$ points to model point--point interactions, then uses a point-wise MLP head to output $K$-way assignment logits per point. 
Its complexity is $\gO(N^2 + NK)$. This design contains no explicit cluster representations; clusters are implicitly represented in the MLP output layer. 
The second baseline is a \emph{non-iterative Transformer decoder}. 
\begin{wraptable}{r}{0.45\textwidth}
\centering
\caption{\textbf{PIN decoder architecture ablation}. Comparison of different model architectures on synthetic and real datasets using median ARI and NMI (higher is better). Bold and underlined values indicate the best and second-best performance in each column.}
\label{tab:architecture_ablation}
\resizebox{\linewidth}{!}{
\begin{tabular}{lccccc}
\toprule
& \multicolumn{2}{c}{synthetic} & & \multicolumn{2}{c}{Real} \\
\cmidrule{2-3} \cmidrule{5-6}
Method & ARI $\uparrow$ & NMI $\uparrow$ & & ARI $\uparrow$ & NMI $\uparrow$ \\
\midrule
Naïve (SA + MLP)            & 0.8137          & 0.7446          & & \underline{0.3589} & 0.3271             \\
Non-iterative (Transformer) & \underline{0.8229} & \underline{0.7523} & & 0.3505             & \underline{0.3865} \\
Iterative (Ours)            & \textbf{0.8308} & \textbf{0.8000} & & \textbf{0.3853}    & \textbf{0.4237}    \\
\bottomrule
\end{tabular}}
\vspace{-10pt}
\end{wraptable}
This variant introduces \emph{explicit cluster prototypes} but omits iterative prototype--data co-refinement: it first encodes the points with self-attention to obtain $R^{(L)}$, then updates the $K$ prototypes using prototype self-attention (cluster--cluster interaction) and prototype--point cross-attention with keys/values from the fixed encoded points $R^{(L)}$ (cluster--point interaction). 
Its complexity remains $\gO(N^2 + NK)$. In contrast, our decoder maintains $K$ cluster prototypes and \emph{iteratively refines} point--cluster assignments and prototypes: each layer alternates (i) prototype self-attention (cluster--cluster interaction) and (ii) bidirectional prototype--point cross-attention between $N$ point embeddings and $K$ prototypes (point--cluster interaction), while avoiding full point--point self-attention in the decoder. 
Each layer costs $\gO(NK + K^2)$, less than both baselines for $K \ll N$.

Tab.~\ref{tab:architecture_ablation} shows that the iterative decoder consistently achieves the best performance across both settings. 
Despite its higher computational cost, the non-iterative Transformer underperforms our design. 
This suggests that explicit pairwise point modeling alone is insufficient; instead, \emph{iterative refinement of cluster prototypes} is key. While cluster queries already improve over the na\"ive SA+MLP decoder, the full iterative mechanism achieves the best clustering quality and efficiency.

\textbf{PIN loss ablation}. 
To assess whether SoftARI is an appropriate objective, we compare it with several natural alternatives while keeping keeping all other components fixed.
We first consider SoftNMI, based on normalized mutual information (NMI), a clustering metric that is invariant to label permutations.
We also include two matching-based baselines that explicitly address the label switching issue. These methods resolve label switching by aligning predicted clusters with ground-truth labels via either minimum-weight matching~\citep{crouse2016implementing} or a differentiable Sinkhorn relaxation~\citep{Sinkhorn1967}, and then apply a cross-entropy (CE) loss or a soft accuracy loss. 
\begin{wraptable}{r}{0.45\textwidth}
\centering
\vspace{-10pt}
\caption{\textbf{\pin{} loss ablation}. Median ARI and NMI on synthetic and real datasets.}
\label{tab:pin_loss_ablation}
\resizebox{\linewidth}{!}{
\begin{tabular}{lcccccc}
\toprule
\multirow{2}{*}{Method}& \multicolumn{2}{c}{Synthetic} & & \multicolumn{2}{c}{Real} & \multirow{2}{*}{Time(ms) $\downarrow$} \\
\cmidrule{2-3} \cmidrule{5-6}
 & ARI $\uparrow$ & NMI $\uparrow$ & & ARI $\uparrow$ & NMI $\uparrow$ \\
\midrule
Matching+SoftAcc & 0.8021          & 0.7364             & & 0.3814             & 0.4071            & 70.0346\\
Matching+CE      & \textbf{0.8627} & \underline{0.7637} & & 0.3507             & 0.4070            & 6.2625\\
SoftNMI          & 0.7637          & 0.6929             & & \textbf{0.3915}    & \textbf{0.4250}   & 3.3658\\
SoftARI (Ours)         & \underline{0.8308} & \textbf{0.8000} & & \underline{0.3853} & \underline{0.4237} &\textbf{2.9695}\\
\bottomrule
\end{tabular}}
\vspace{-13pt}
\end{wraptable}
Soft versions of NMI and accuracy are adopted to avoid discrete operations (e.g., hard labeling or hard matching). These baselines are denoted as Matching+SoftAcc and Matching+CE, respectively. 
Details of all loss formulations and an additional ablation study on CIN loss are provided in App.~\ref{app:ablation_details}.
Tab.~\ref{tab:pin_loss_ablation} reports the median ARI of \ours{} pretrained with different \pin{} losses on both synthetic and real datasets. 
SoftARI yields the best overall performance. 
While Matching+CE achieves comparable accuracy, its pretraining cost is substantially higher due to the additional computational overhead introduced by the matching step.
In the last column, we report the total time for one forward and backward pass with batch size $8$ and $1000$ samples per dataset. 
The per-iteration time of Matching+CE is approximately $2\times$ that of our approach, resulting in a doubled optimization time, which is significant given $10{,}000$ steps.

\begin{wraptable}{r}{0.45\textwidth}
\centering
\vspace{-\intextsep}
\caption{\textbf{Decoupled loss ablation}. Median ARI and NMI on synthetic and real datasets.}
\label{tab:decoupled_loss_ablation}
\resizebox{\linewidth}{!}{
\begin{tabular}{lccccc}
\toprule
& \multicolumn{2}{c}{Synthetic} & & \multicolumn{2}{c}{Real} \\
\cmidrule{2-3} \cmidrule{5-6}
Method & ARI $\uparrow$ & NMI $\uparrow$ & & ARI $\uparrow$ & NMI $\uparrow$ \\
\midrule
Additive & \underline{0.8103} & \underline{0.7227} & & \underline{0.3544} & \underline{0.3716} \\
Decoupled (Ours)  & \textbf{0.8308} & \textbf{0.8000} & & \textbf{0.3853}    & \textbf{0.4237}    \\
\bottomrule
\end{tabular}}
\vspace{-10pt}
\end{wraptable}
\textbf{Decoupled optimization ablation}.
We compare \emph{additive joint training} of \pin{} and \cin{} with our \emph{decoupled strategy}, where \cin{} operates on detached \pin{} outputs. 
As shown in Tab.~\ref{tab:decoupled_loss_ablation}, decoupling consistently improves performance on both synthetic and real datasets. 
We attribute this to reduced optimization interference: partition inference and cardinality estimation operate at different structural levels, and shared gradients can bias representations toward predicting $K$ at the expense of fine-grained partition structure. 
Decoupling stabilizes cluster representations, improving both clustering quality and estimation of $K$.

\begin{wraptable}[7]{r}{0.45\textwidth}
\centering
\vspace{-\intextsep}
\caption{\textbf{Prior data types ablation}. Median ARI on synthetic and real datasets.}
\label{tab:prior_ablation_ari}
\resizebox{\linewidth}{!}{
\begin{tabular}{lcccc}
\toprule
\multirow{2}{*}{Prior data}& \multicolumn{3}{c}{Synthetic $\uparrow$} & \multirow{2}{*}{Real $\uparrow$} \\
\cmidrule{2-4}
 & GMM & ZEUS & GMM+ZEUS & \\
\midrule
GMM             & \textbf{0.8236}    & 0.5434             & \underline{0.7428} & 0.2270             \\
ZEUS            & 0.6758             & \underline{0.8047} & 0.7405             & \underline{0.3414} \\
GMM+ZEUS (Ours) & \underline{0.8230} & \textbf{0.8334}    & \textbf{0.8308}    & \textbf{0.3853}    \\
\bottomrule
\end{tabular}}
\vspace{-6pt}
\end{wraptable}
\textbf{Different prior data mechanisms}.
To evaluate the impact of prior data diversity, we consider three prior configurations: training with GMM prior data only, ZEUS prior data only, or a combination of both. 
We then assess performance on both synthetic and real datasets. 
For the synthetic benchmarks, we further provide a fine-grained analysis by grouping datasets according to their data-generation mechanisms. 
The results are reported in Tab.~\ref{tab:prior_ablation_ari}.
Increasing the diversity of prior data consistently improves overall performance across both synthetic and real datasets. 
While the ZEUS-only prior performs best on datasets generated from the same family, its generalization to other dataset types is limited. 
These findings are consistent with~\cite{zhang2025mitra}, which show that greater prior diversity can enhance the performance of PFN models.
We report negative results from alternative prior constructions in App.~\ref{app:alternative_priors}.
\WFclear

%% file: sections/conclusion.tex
\section{Limitations \& conclusions}
\label{sec:limits}
We introduced \ours{}, an extensible PFN-based framework for tabular clustering that jointly infers cluster assignments and cardinality in a single forward pass without dataset-specific tuning. Unlike prior PFN-based clustering methods that rely on known-cardinality assumptions, low-dimensional synthetic priors, or label-ordering conventions, \ours{} combines a hybrid tabular prior, permutation-invariant SoftARI training, and a decoupled PIN--CIN architecture to support high-dimensional heterogeneous tabular data. Across 44 real-world datasets, \ours{} achieves SOTA performance over classical, deep, and amortized baselines while maintaining fast inference.

\textbf{Future work}. Due to computational constraints, our experiments focus on datasets of moderate size and dimensionality. The OOD evaluations provide evidence of generalization beyond the pretraining regime, with degradation under some shifts.
Moderately larger datasets ($N,D$) can use the same architecture with broader pretraining and more compute. Increasing $K_{\max}$ requires enlarging the PIN prototype bank and CIN input/output layers, then retraining on higher-cardinality priors.
Improving robustness to noisy or missing entries through corruption-aware priors is another important direction for future work.

%% file: sections/appendix/related.tex
\section{Related work}
\label{app:related_work}
\textbf{Classical probabilistic and distance-based clustering methods.}
Classical clustering methods remain widely used for tabular data due to their simplicity and computational efficiency. 
However, they typically operate directly on the input features and rely on rigid geometric or probabilistic assumptions to separate data. 
For example, \emph{$k$-means}~\citep{lloyd1982least} favors convex centroid-based partitions, Gaussian mixture models (GMMs) assume Gaussian component densities, \emph{DBSCAN}~\citep{ester1996density} relies on a global density scale, and hierarchical clustering~\citep{johnson1967hierarchical} is highly sensitive to the choice of distance metric and linkage criterion. 
Such inductive biases are effective in simple settings, but often fail to capture the heterogeneous structures commonly observed in real-world tabular data.

\textbf{Manifold and representation learning methods.}
To overcome the rigidity of classical methods, manifold and representation learning approaches seek transformations under which cluster structures become more separable. 
\emph{Spectral clustering}~\citep{ng2001spectral} captures non-convex structures through graph Laplacian eigenspace embeddings, while recent \emph{deep clustering} methods~\citep{rauf2025tabledc,ren2024deep} learn nonlinear representations with neural networks. 
Although these methods are more expressive, they still formulate clustering as a \emph{dataset-specific} optimization problem. 
Spectral clustering requires eigendecomposition with cubic complexity in the number of samples, and deep clustering methods require training a separate model for each dataset. 
As a result, they are often too costly for exploratory analysis and do not provide a transferable clustering inductive bias across datasets.

\textbf{Amortized clustering methods.}
A more recent paradigm replaces dataset-specific optimization with amortized neural inference. 
Amortized clustering methods~\citep{lee2019set,lee2019deep,pakman2020neural} train neural networks to infer cluster assignments in a zero-shot or iterative manner, thereby reducing the need for per-dataset search or model fitting. 
However, existing approaches are typically trained to approximate simple generative priors such as GMMs. 
Consequently, while they improve inference efficiency, their clustering behavior remains constrained by the same parametric assumptions underlying the priors they emulate.

%% file: sections/appendix/training_details.tex
\section{Pretraining details}
\label{app:training_details}
This section provides additional details on the pretraining setup. 
The prior data used for pretraining (Sec.~\ref{app:prior_data}) are presented first, followed by network architecture details (Sec.~\ref{app:network_details}).

\subsection{Prior data generation}
\label{app:prior_data}
We first sample basic dataset-level properties. 
The number of observations $N$ is drawn uniformly from integers between $500$ and $1000$. 
The feature dimension $D$ is sampled uniformly from integers between $2$ and $64$. 
The number of clusters $K$ is drawn from $\{2, \ldots, K_{\max}\}$ using a non-uniform prior: $\sP(K=2)=0.3$, while with probability $0.7$, $K$ is sampled uniformly from $\{3, \ldots, 10\}$.
This non-uniform prior over the number of clusters follows the setting of~\cite{qu2025tabicl}. 
It places greater probability mass on simpler cases, which can facilitate training by biasing the model toward learning from less complex clustering structures.

We then select the data-generating mechanism by sampling a Bernoulli variable, using the GMM prior with probability $0.4$ and the ZEUS prior with probability $0.6$. 
The selected mechanism is then used to generate a synthetic clustering dataset in $\mathbb{R}^D$ with $N$ observations, along with latent component indices serving as ground-truth cluster labels. We subsequently apply the same preprocessing steps to every generated dataset, including feature $z$-norm, feature permutation, before assembling datasets into training mini-batches. We next describe the two priors in detail.

\subsubsection{GMM prior}
Clustering difficulty in mixture models is largely governed by the degree of overlap between components: well-separated components yield near-deterministic assignments, whereas heavily overlapping components make cluster identities intrinsically ambiguous. To systematically control task difficulty during pretraining, we therefore generate datasets from Gaussian mixture models (GMMs) with an explicit constraint on component overlap.

\textbf{Overlap measure}.
For a $K$-component Gaussian mixture with parameters $\{(\pi_k, \mu_k, \Sigma_k)\}_{k=1}^K$, we quantify pairwise overlap using Bayes misclassification probabilities under the model-based decision rule. Let $\phi(x | \mu, \Sigma)$ denote the Gaussian density and $\pi_k$ the mixing weight. The one-sided overlap from component $i$ to component $j$ is defined as
\[
o_{j|i} = \mathbb{P} \left( \pi_i \phi(X | \mu_i, \Sigma_i) < \pi_j \phi(X | \mu_j, \Sigma_j) | X \sim \mathcal{N}(\mu_i, \Sigma_i) \right),
\]
which is the probability that a sample generated from component $i$ would be assigned to component $j$ by the Bayes classifier. The symmetric pairwise overlap is
\[
\omega_{ij} = o_{j | i} + o_{i | j},
\]
and the overall dataset difficulty is characterized by the maximum pairwise overlap
\[
\Omega_{\max} = \max_{i \neq j \in [K]} \omega_{ij}.
\]
Small $\Omega_{\max}$ corresponds to well-separated clusters, while large values indicate intrinsically ambiguous clustering structure.

\textbf{Controlled GMM generation}.
We generate GMM parameters using the \textsc{MixSim} package~\citep{mixsim} in R, which directly constructs mixtures satisfying a prescribed upper bound on $\Omega_{\max}$. 
This enables sampling datasets with controlled and diverse levels of clustering difficulty.
We specify the configuration of MixSim below.

For each synthetic dataset, we sample mixture parameters as follows:
\begin{itemize}[leftmargin=*]
\item \textbf{Mixture weights.}  
We sample $\pi \sim \text{Dirichlet}(\alpha \mathbf{1}_{K})$ with concentration $\alpha = 2$, encouraging moderately balanced clusters. To avoid degenerate components, we enforce a minimum proportion constraint using \texttt{PiLow} $= 10^{-1}$.

\item \textbf{Target overlap (difficulty control).}  
We first sample a target maximum pairwise overlap
\[
\Omega_{\max} \sim \text{Uniform}\!\left[0.01,\; \min\!\left(0.8,\; \frac{1.5}{D^{0.82}}\right)\right].
\]
The dimension-dependent upper bound ensures feasibility of the overlap constraint in higher dimensions, where random components tend to become more separable. 
\textsc{MixSim} then constructs $(\mu_k, \Sigma_k)$ such that the resulting mixture satisfies this target $\Omega_{\max}$.

\item \textbf{Component shape diversity.}  
To promote geometric diversity, we randomize covariance structures. For each dataset, we independently sample whether components are spherical (\texttt{sph = TRUE}) or ellipsoidal (\texttt{sph = FALSE}), and whether covariances are shared across clusters (\texttt{hom = TRUE/FALSE}). When ellipsoidal covariances are used, we allow strong anisotropy via a maximum eccentricity constraint (\texttt{ecc} $= 0.9$), producing elongated clusters.

\item \textbf{Component location range.}  
Component means are constrained to a bounded hypercube by sampling each $\mu_k$ uniformly from $[-1,1]^D$. This controls the global scale of cluster locations and prevents extreme separation driven purely by mean magnitude.
\end{itemize}

\textbf{Data sampling}.
Given the generated parameters $\{(\pi_k, \mu_k, \Sigma_k)\}_{k=1}^K$, we first draw $N$ latent component labels $Z_i \overset{\text{IID}}{\sim} \text{Categorical}(\pi)$, which serve as ground-truth cluster assignments. 
Then, conditioning on each $Z_i$, we draw samples from $\gN(\mu_{Z_i}, \Sigma_{Z_i})$.

\textbf{Visualization}.
Overall, these controls provide a scalable way to generate GMM datasets with explicitly tunable inter-cluster ambiguity while retaining substantial diversity in cluster geometry and cluster-size imbalance. 
Fig.~\ref{fig:gmm_vis} shows representative 2D mixtures generated under different $\Omega_{\max}$ values.

\begin{figure}[!t]
    \centering
    \foreach \i in {0,1,...,63} {%
        \ifnum\i<10
            \def\imgname{00\i}%
        \else
            \def\imgname{0\i}%
        \fi
        \includegraphics[width=0.11\textwidth]{figures/vis/gm_\imgname.pdf}%
        \makeatletter
        \pgfmathparse{mod(\i+1, 8) == 0 ? 1 : 0}%
        \ifnum\pgfmathresult=1
            \par\smallskip
        \else
            \hfill
        \fi
        \makeatother
    }
    \caption{\textbf{Visualization of randomly generated 2D datasets from the GMM prior}.}
    \label{fig:gmm_vis}
\end{figure}

\subsubsection{ZEUS prior}
While the GMM prior controls clustering difficulty through Gaussian overlap, the ZEUS prior generates more geometrically and statistically heterogeneous tabular data by combining latent mixture structure with nonlinear transformations and mixed feature types.

Given the total feature dimensionality $D$, we randomly partition features into a continuous subset and a categorical subset. We first sample the number of continuous features $D_{\text{cont}} \geq 2$, and treat the remaining $D - D_{\text{cont}}$ dimensions as categorical. Each categorical feature is label-encoded (rather than one-hot), and its cardinality is sampled from a predefined range.

\textbf{Continuous features}.
For the continuous feature subsets, we sample mixture weights $\pi \sim \text{Dirichlet}(\alpha\mathbf{1}_{K})$ with $\alpha=2$ and latent component labels $Z_i \sim \text{Categorical}(\pi)$, which serve as ground-truth cluster assignments.
Unlike ZEUS, which uses a smaller $\alpha$ to induce more imbalanced mixtures, we set $\alpha=2$ to avoid degenerate components and overly skewed cluster sizes, yielding more stable supervision and a broader range of moderately imbalanced clustering instances.
Conditional on $Z_i$, base continuous observations are drawn from cluster-specific distributions as described in ZEUS~\citep{marszaek2025zeus}. 
For completeness, we summarize the relevant settings below.

To increase geometric complexity beyond a simple mixture structure, ZEUS introduces nonlinear feature interactions through an invertible transformation. 
With probability $0.5$, the identity mapping is used, and with probability $0.5$, a randomly sampled invertible iResNet bijection with Lipschitz constant constrained to $[0.1, 0.9]$ is applied. 
This transformation yields nonlinearly distorted cluster shapes while preserving cluster identity due to invertibility.
The number of invertible blocks follows a log-scaled truncated normal meta-distribution. 
This design choice follows prior tabular data generation work (e.g., TabICL) rather than being specific to ZEUS, as it can increase the diversity of the data generation process.
Concretely,
\[
\log \mu \sim \text{Uniform}(\log 3, \log 8), 
\qquad
\log \sigma \sim \text{Uniform}(\log 0.01, \log 1.0),
\]
$\tilde{b} \sim \mathcal{N}(\mu, \sigma^2)$ is sampled with truncation to $[0, \infty)$, and the number of blocks is set as $b = 3 + \text{round}(\tilde{b})$. 
This results in variable transformation depth across datasets.

\textbf{Categorical features}.
Following ZEUS, categorical variables are generated conditional on the latent cluster labels. 
For each categorical feature $j$, we first sample its cardinality
\[
C_j \in \{2, \dots, C_{\max}\}, \qquad C_{\max} = 5,
\]
then draw cluster-specific categorical distributions (e.g., via Dirichlet priors) so that different clusters exhibit distinct category preferences. 
Each observation samples its categorical value from the distribution associated with its latent component $z_i$, providing discrete cluster cues complementary to the continuous geometry.

\textbf{Data assembly}.
Finally, we concatenate the transformed continuous features and categorical features to obtain a $D$-dimensional tabular dataset. 
This construction produces clusters that may be nonlinearly separable in the observed space and only partially identifiable from any single feature type, encouraging the model to learn flexible multimodal structure.

\textbf{Visualization}.
Figure~\ref{fig:zeus_vis} shows representative 2D datasets generated from the ZEUS prior, illustrating nonlinear cluster geometry and mixed-type feature structure.

\begin{figure}[!t]
    \centering
    \foreach \i in {0,1,...,63} {%
        \ifnum\i<10
            \def\imgname{00\i}%
        \else
            \def\imgname{0\i}%
        \fi
        \includegraphics[width=0.11\textwidth]{figures/vis/gmires_\imgname.pdf}%
        \makeatletter
        \pgfmathparse{mod(\i+1, 8) == 0 ? 1 : 0}%
        \ifnum\pgfmathresult=1
            \par\smallskip
        \else
            \hfill
        \fi
        \makeatother
    }
    \caption{\textbf{Visualization of randomly generated 2D datasets from the ZEUS prior}.}
    \label{fig:zeus_vis}
\end{figure}

\subsubsection{Alternative prior constructions and negative results}
\label{app:alternative_priors}
We explored several alternative prior constructions before selecting the GMM and ZEUS priors used in \ours{}.
We summarize these exploratory results to document the choices underlying the final prior.

\paragraph{SCM-derived partitions.}
We reused supervised MLP-based and tree-based structural causal model (SCM) generators and constructed cluster labels by discretizing their outputs.
These constructions performed poorly in our experiments.
In the configurations tested, the discretized targets did not reliably correspond to recoverable cluster structure in the observed features.
This observation motivated a generative construction with explicit latent cluster assignments and cluster-specific distributions, as in Eq.~\eqref{eq:generate_model}.

\paragraph{Unconstrained MLP transformations.}
We also sampled latent Gaussian components and transformed their observations using unconstrained random MLPs while retaining the component labels.
This construction likewise performed poorly.
A non-injective transformation can map distinct latent components to overlapping or indistinguishable regions, making the retained labels difficult to recover from the observed data.
This provides a possible explanation for the observed behavior and motivated the use of invertible transformations to preserve component information while introducing nonlinear geometry.

\paragraph{Other transformation families.}
We further explored affine-coupling normalizing flows and cluster-aware transformations.
These alternatives trained stably but did not improve downstream clustering performance over the iResNet-based ZEUS prior in the tested configurations.
One possible explanation is that they vary the transformation family without providing sufficiently complementary training distributions.
By contrast, the prior ablation in Tab.~\ref{tab:prior_ablation_ari} shows that combining the controlled-overlap GMM prior with the ZEUS prior improves aggregate performance over either prior alone.

These observations are specific to the constructions and configurations tested; they do not establish that the alternative prior families are ineffective in general.
They suggest that recoverable cluster structure and complementary data distributions are useful considerations when designing clustering priors.

\subsection{Network architecture details}
\label{app:network_details}

\subsubsection{Partition Inference Network (\pin{})}

Given a tabular dataset $X \in \mathbb{R}^{N \times D}$ and a candidate cluster cardinality $K \in \{2, \dots, K_{\text{max}}\}$, the \pin{} outputs a soft partition
\begin{equation*}
    \text{PIN}_{\theta} : (X, K) \mapsto P \in [0, 1]^{N \times K}, \quad P_{ik} = \hat{p}(Z_i = k | X, K),
\end{equation*}
where $Z_i$ denotes the latent cluster label of the $i$-th observation.

\textbf{Encoder}.
The encoder of our \pin{} follows the TabICL design and consists of (i) a column-wise embedding module and (ii) a row-wise interaction module. We refer to TabICL for architectural details of these two components. Formally, given a dataset, the PIN encoder produces row representations
\begin{equation*}
    R^{(0)} = \text{Enc}(X)=\text{RowInteract}(\text{ColEmbed}(X)) \in \mathbb{R}^{N \times d},
\end{equation*}
where $d$ is the representation dimension produced by the encoder. In our implementation, we use $d=512$ following TabICL.

\textbf{Decoder with learnable prototypes.}
For each candidate $K$, \pin{} instantiates $K$ learnable cluster prototypes by selecting the first $K$ rows from a global prototype matrix
\begin{equation*}
    C^{(0)} = \text{Select}_K(\mathbf{C}) \in \mathbb{R}^{K \times d}, \quad \mathbf{C} \in \mathbb{R}^{K_{\text{max}} \times d}.
\end{equation*}
The decoder then applies $L$ stacked \textit{iterative cross-attention blocks} to co-refine the data representations and prototypes.
As discussed in Sec.~\ref{sec:network}, all attention layers omit positional encodings to preserve permutation equivariance.
We set $L=6$ in all experiments, unless otherwise stated.
At layer $\ell \in \{0, \dots, L - 1\}$, the update consists of two stages:

\begin{enumerate}[leftmargin=*]
    \item \textbf{Prototype update (prototype $\to$ data cross-attention).}
    We first update the prototypes via (i) self-attention among prototypes and (ii) cross-attention where prototypes attend to data:
    \begin{equation*}
        C^{(\ell+\frac{1}{2})} = \text{SA}\!\left(C^{(\ell)}\right), \quad
        C^{(\ell+1)} = \text{CA}_{C \leftarrow R}\!\left(C^{(\ell+\frac{1}{2})}, R^{(\ell)}\right).
    \end{equation*}


    We instantiate both $\text{SA}(\cdot)$ and $\text{CA}(\cdot,\cdot)$ using the same Transformer-style attention block with multi-head attention (MHA), pre-norm layer normalization, residual connections, and a position-wise feed-forward network (MLP).
    As discussed in Sec.~\ref{sec:network}, we omit positional encodings to preserve permutation equivariance. For completeness, we detail the exact attention-block formulation shared by $\text{SA}$ and $\text{CA}$ below.

    \textit{Definition of SA/CA blocks.}
    We adopt a pre-norm variant of the Multihead Attention Block (MAB) used in Set Transformer~\citep{lee2019set}, which we denote by $\mathrm{MAB}_{\text{pre}}(\cdot\,;\cdot)$. 
    Let $H$ be the number of attention heads and $d_h=d/H$ the head dimension; we use $H=4$ in all experiments unless otherwise stated.
    Given queries $Q\in\mathbb{R}^{n_q\times d}$, keys $K\in\mathbb{R}^{n_k\times d}$, and values $V\in\mathbb{R}^{n_k\times d}$, multi-head attention (MHA) is
    \begin{equation*}
    \mathrm{MHA}(Q,K,V)
        = \mathrm{Concat}\!\left(\mathrm{head}_1,\dots,\mathrm{head}_H\right)W^O,
    \end{equation*}
    where 
    \[\mathrm{head}_h
        = \mathrm{softmax}\!\left(\frac{QW_h^Q(KW_h^K)^\top}{\sqrt{d_h}}\right)\,VW_h^V,\]
    $\{W_h^Q,W_h^K,W_h^V\}_{h=1}^H$ and $W^O$ are learnable projections.
    An attention sublayer with pre-norm and residual is defined as
    \begin{equation*}
        \mathrm{Attn}(X;Y)=X+\mathrm{MHA}\!\left(\mathrm{LN}(X),\mathrm{LN}(Y),\mathrm{LN}(Y)\right),
    \end{equation*}
    and the full pre-norm MAB applies a position-wise feed-forward network (MLP/FFN) with pre-norm and residual:
    \begin{equation}
    \label{eq:mab_pre}
        \mathrm{MAB}_{\text{pre}}(X;Y)
        = \mathrm{Attn}(X;Y)
        + \mathrm{MLP}\!\left(\mathrm{LN}(\mathrm{Attn}(X;Y))\right),
    \end{equation}
    where $\mathrm{MLP}$ is a two-layer position-wise FFN with hidden width $2d$ (i.e., $d\!\to\!2d\!\to\!d$).
    
    \textit{Self- and cross-attention operators.}
    Using $\mathrm{MAB}_{\text{pre}}$, we define self-attention on a set as
    \begin{equation}
    \label{eq:sa_def}
        \mathrm{SA}(X) := \mathrm{MAB}_{\text{pre}}(X;X),
    \end{equation}
    i.e., queries/keys/values are all drawn from $X$ (capturing within-set interactions).
    Cross-attention from a query set $X$ to a context set $Y$ is defined as
    \begin{equation}
    \label{eq:ca_def}
        \mathrm{CA}_{X\leftarrow Y}(X,Y) := \mathrm{MAB}_{\text{pre}}(X;Y),
    \end{equation}
    i.e., queries come from $X$ while keys/values come from $Y$ (aggregating information from $Y$ into $X$).

    In particular, prototype self-attention and prototype-to-data cross-attention are
    \begin{equation*}
        \mathrm{SA}(C) \in \mathbb{R}^{K\times d}, \qquad
        \mathrm{CA}_{C\leftarrow R}(C,R) \in \mathbb{R}^{K\times d},
    \end{equation*}
    where $\mathrm{SA}(C)$ captures inter-cluster interactions and $\mathrm{CA}_{C\leftarrow R}(C,R)$ aggregates dataset evidence into prototype summaries.

    \item \textbf{Representation update (data $\to$ prototype cross-attention).}
    We then update data representations by attending to the updated prototypes:
    \begin{equation*}
        R^{(\ell+1)} = \text{CA}_{R \leftarrow C}\!\left(R^{(\ell)}, C^{(\ell+1)}\right),
    \end{equation*}
    where $\text{CA}_{R \leftarrow C}(R,C):=\mathrm{Block}(R;C)$, i.e., $Q=R$ and $K=V=C$, injecting cluster-level structure back into point representations.
\end{enumerate}

\textbf{Cosine-similarity logits}.
After the final decoder layer, we compute logits via normalized cosine similarity between row representations and prototypes. Let
\begin{equation*}
    \bar{r}_i = \frac{g\left(R_{i:}^{(L)}\right)}{\left\|g\left(R_{i:}^{(L)}\right)\right\|_2}, \quad \bar{c}_j = \frac{g\left(C_{j:}^{(L)}\right)}{\left\|g\left(C_{j:}^{(L)}\right)\right\|_2},
\end{equation*}
where $A_{i:}$ denotes the $i$th row of matrix $A$ and $g(\cdot)$ is a shared MLP projection applied to both rows and prototypes. 
We then compute the logits as
\begin{equation*}
    \ell_{ij} = \tau \bar{r}_i^\top \bar{c}_j, \quad \tau > 0,
\end{equation*}
where $\tau$ is a learnable temperature. Finally, the soft partition is obtained by a row-wise softmax:
\begin{equation*}
    P_{i:} = \text{softmax}(\ell_{i:}) \in [0, 1]^K.
\end{equation*}

PIN is trained with the permutation-invariant SoftARI loss on $P$.

\subsubsection{Cardinality Inference Network (\cin{})}
\label{app:cin}

To infer the cluster numbers, we run \pin{} for each $K \in \{2, \dots, K_{\text{max}}\}$ to obtain $\{P^{(K)}\}$. For each $K$, we compute the normalized Gram matrix
\begin{equation*}
    G^{(K)} = \frac{1}{N}(P^{(K)})^\top P^{(K)} \in \mathbb{R}^{K \times K},
\end{equation*}
which summarizes pairwise cluster co-assignment statistics. We then form a permutation-invariant fingerprint by extracting the diagonal entries and the strictly upper-triangular off-diagonal entries, sorting each in descending order, and concatenating them:
\begin{equation*}
    g^{(K)} = \left[ \text{sort}(\text{diag}(G^{(K)})) \,;\, \text{sort}(\text{triu}(G^{(K)}, 1)) \right] \in \mathbb{R}^{K(K+1)/2}.
\end{equation*}

Concatenating $\{g^{(K)}\}_{K=2}^{K_{\text{max}}}$ yields a fixed-length vector, which is passed through an MLP to output logits over $K$:
\begin{equation*}
    \text{CIN}_{\phi}(X) = \text{MLP}\left(\text{concat}(g^{(2)}, \dots, g^{(K_{\text{max}})})\right) \in \mathbb{R}^{K_{\text{max}}-1}.
\end{equation*}

In our implementation, \cin{} is a $3$-layer MLP with hidden width $256$, which is trained with cross-entropy on the ground-truth $K^*$. Following our decoupled optimization design, gradients from \cin{} do not backpropagate into \pin{}.

%% file: sections/appendix/exp_details.tex
\section{Experimental details}
\label{app:experimental_details}
This section provides additional details on the evaluation setup. 
Statistics of the synthetic and real-world benchmarks (Sec.~\ref{sec:dataset_stats}) are presented first, followed by descriptions of the baseline methods (Sec.~\ref{app:baselines}), data preprocessing details (Sec.~\ref{app:data_preprocessing}), and the experiment setting for amortized clustering (Sec.~\ref{app:amortized_clustering}).

\subsection{Statistics of synthetic and real-world benchmarks}
\label{sec:dataset_stats}

\textbf{Synthetic evaluation datasets}.
To assess the generalization ability of \ours{} on unseen data drawn from the same distribution families as the pretraining data, we construct a held-out synthetic benchmark, constructed by setting a different random seed. 
Following the two classes of priors introduced in Sec.~\ref{sec:prior_data}, we refer to them as the \emph{GMM prior} and the \emph{ZEUS prior}.  
The benchmark contains 25 synthetic datasets generated from the GMM prior and 24 synthetic datasets generated from the ZEUS prior.
Detailed statistics of these evaluation datasets are reported in Tab.~\ref{tab:synthetic_stats_gmm} (GMM prior) and Tab.~\ref{tab:synthetic_stats_zeus} (ZEUS prior). 
In both tables, the \texttt{ID} column provides a reference index for each synthetic dataset.
\begin{table}[!ht]
\centering
\caption{\textbf{Statistics of synthetic datasets for evaluation generated with the \textit{GMM} prior}.}
\label{tab:synthetic_stats_gmm}
\small
\begin{tabular}{lccccc}
\toprule
ID & \# Instances & \# Numerical features & \# Categorical features & Dimension & \# Classes \\
\midrule
0  & 511  & 55 & 0  & 55 & 8  \\
1  & 741  & 33 & 0  & 33 & 2  \\
2  & 827  & 8  & 0  & 8  & 7  \\
3  & 993  & 47 & 0  & 47 & 10 \\
4  & 740  & 23 & 0  & 23 & 6  \\
5  & 675  & 39 & 0  & 39 & 5  \\
6  & 718  & 10 & 0  & 10 & 7  \\
7  & 794  & 44 & 0  & 44 & 2  \\
8  & 724  & 16 & 0  & 16 & 2  \\
9 & 984  & 55 & 0  & 55 & 4  \\
10 & 588  & 4  & 0  & 4  & 2  \\
11 & 595  & 20 & 0  & 20 & 5  \\
12 & 850  & 16 & 0  & 16 & 5  \\
13 & 700  & 31 & 0  & 31 & 7  \\
14 & 869  & 7  & 0  & 7  & 3  \\
15 & 722  & 8  & 0  & 8  & 2  \\
16 & 693  & 55 & 0  & 55 & 4  \\
17 & 989  & 19 & 0  & 19 & 4  \\
18 & 892  & 35 & 0  & 35 & 2  \\
19 & 633  & 5  & 0  & 5  & 2  \\
20 & 825  & 46 & 0  & 46 & 7  \\
21 & 691  & 17 & 0  & 17 & 2  \\
22 & 853  & 58 & 0  & 58 & 7  \\
23 & 707  & 43 & 0  & 43 & 9  \\
24 & 753  & 56 & 0  & 56 & 6  \\
\bottomrule
\end{tabular}
\end{table}

\begin{table}[!ht]
\centering
\caption{\textbf{Statistics of synthetic datasets for evaluation generated with the \textit{ZEUS} prior}.}
\label{tab:synthetic_stats_zeus}
\small
\begin{tabular}{lccccc}
\toprule
ID & \# Instances & \# Numerical features & \# Categorical features & Dimension & \# Classes \\
\midrule
0  & 649  & 3  & 4  & 7  & 2  \\
1  & 910  & 18 & 14 & 32 & 7  \\
2  & 948  & 6  & 43 & 49 & 6  \\
3  & 905  & 13 & 17 & 30 & 2  \\
4  & 863  & 14 & 1  & 15 & 2  \\
5  & 898  & 6  & 0  & 6  & 2  \\
6  & 589  & 11 & 9  & 20 & 10 \\
7  & 568  & 12 & 9  & 21 & 10 \\
8 & 523  & 38 & 10 & 48 & 10 \\
9 & 785  & 2  & 27 & 29 & 2  \\
10 & 658  & 2  & 17 & 19 & 6  \\
11 & 628  & 11 & 41 & 52 & 5  \\
12 & 862  & 21 & 3  & 24 & 2  \\
13 & 632  & 35 & 17 & 52 & 2  \\
14 & 906  & 5  & 57 & 62 & 2  \\
15 & 688  & 4  & 25 & 29 & 10 \\
16 & 757  & 10 & 17 & 27 & 8  \\
17 & 705  & 19 & 31 & 50 & 2  \\
18 & 590  & 14 & 3  & 17 & 3  \\
19 & 535  & 3  & 4  & 7  & 2  \\
20 & 635  & 15 & 22 & 37 & 2  \\
21 & 546  & 38 & 6  & 44 & 2  \\
22 & 757  & 2  & 3  & 5  & 6  \\
23 & 633  & 19 & 3  & 22 & 9  \\
\bottomrule
\end{tabular}
\end{table}

\textbf{Real-world benchmark.}
We construct a real-world tabular clustering benchmark by aggregating datasets from several widely used sources: OpenML-CC18~\citep{openmlsuite}, TabArena~\citep{erickson2025tabarena}, the AutoML benchmark suite~\citep{automl}, the ZEUS evaluation dataset group, and a curated UCI clustering collection~\citep{milaan9_clustering_datasets}. 
Tab.~\ref{tab:openml_stats} and Tab.~\ref{tab:uci_stats} report dataset-level statistics and identifiers. 
The OpenML table lists datasets by OpenML ID, while the UCI table follows the original dataset names. 
For each dataset, we report the number of instances, the counts of numerical and categorical features, the total feature dimensionality (numerical + categorical), and the number of clusters. 
\begin{table}[!ht]
\centering
\caption{\textbf{Statistics of OpenML datasets used in the real-world benchmark}.}
\label{tab:openml_stats}
\begin{tabular}{lccccc}
\toprule
ID & \# Instances & \# Numerical features & \# Categorical features & Dimension & \# Classes \\
\midrule
16    & 2000 & 64 & 0 & 64 & 10 \\
18    & 2000 & 6  & 0 & 6  & 10 \\
22    & 2000 & 47 & 0 & 47 & 10 \\
51    & 294  & 6  & 7 & 13 & 2  \\
53    & 270  & 13 & 0 & 13 & 2  \\
481   & 209  & 7  & 1 & 8  & 2  \\
694   & 310  & 8  & 0 & 8  & 9  \\
721   & 200  & 10 & 0 & 10 & 2  \\
733   & 209  & 6  & 0 & 6  & 2  \\
756   & 159  & 15 & 0 & 15 & 2  \\
796   & 209  & 6  & 1 & 7  & 2  \\
820   & 235  & 12 & 0 & 12 & 2  \\
840   & 205  & 17 & 8 & 25 & 2  \\
854   & 158  & 5  & 2 & 7  & 2  \\
1063  & 522  & 21 & 0 & 21 & 2  \\
1495  & 250  & 0  & 6 & 6  & 2  \\
1510  & 569  & 30 & 0 & 30 & 2  \\
1523  & 310  & 6  & 0 & 6  & 3  \\
40496 & 500  & 7  & 0 & 7  & 10 \\
40705 & 959  & 42 & 2 & 44 & 2  \\
42585 & 344  & 4  & 2 & 6  & 3  \\
40981 & 690  & 6  & 8 & 14 & 2  \\
\bottomrule
\end{tabular}
\end{table}

\begin{table}[!ht]
\centering
\caption{\textbf{Statistics of UCI datasets used in the real-world benchmark}.}
\label{tab:uci_stats}
\begin{tabular}{lccccc}
\toprule
Name & \# Instances & \# Numerical features & \# Categorical features & Dimension & \# Classes \\
\midrule
haberman    & 306   & 2  & 1  & 3  & 2  \\
banknote    & 1372  & 4  & 0  & 4  & 2  \\
iris        & 150   & 4  & 0  & 4  & 3  \\
thyroid     & 215   & 5  & 0  & 5  & 3  \\
ecoli       & 336   & 7  & 0  & 7  & 8  \\
seeds       & 210   & 7  & 0  & 7  & 3  \\
pima        & 768   & 8  & 0  & 8  & 2  \\
breast      & 277   & 9  & 0  & 9  & 2  \\
wisc        & 699   & 9  & 0  & 9  & 2  \\
glass       & 214   & 9  & 0  & 9  & 6  \\
wine        & 178   & 13 & 0  & 13 & 3  \\
heart       & 303   & 13 & 0  & 13 & 2  \\
vote        & 435   & 16 & 0  & 16 & 2  \\
zoo         & 101   & 16 & 0  & 16 & 7  \\
vehicle     & 846   & 18 & 0  & 18 & 4  \\
segment     & 2310  & 19 & 0  & 19 & 7  \\
wdbc        & 569   & 31 & 0  & 31 & 2  \\
iono        & 351   & 34 & 0  & 34 & 2  \\
dermatology & 366   & 1  & 33 & 34 & 6  \\
landsat     & 2000  & 36 & 0  & 36 & 6  \\
biodeg      & 1055  & 41 & 0  & 41 & 2  \\
spambase    & 4601  & 57 & 0  & 57 & 2  \\
\bottomrule
\end{tabular}
\end{table}

To ensure compatibility with the modeling assumptions and computational regime of \ours{}, we first filter all candidate datasets by feature dimensionality and cluster cardinality, retaining only those with $D \leq 64$ and $K \leq 10$. 
We then control dataset scale through source-dependent handling of the number of instances $N$.
For large benchmark suites with many candidate datasets (e.g., OpenML-CC18, TabArena, and the AutoML benchmark), we further restrict the pool by keeping only datasets whose original size satisfies $N \leq 1000$. This prevents these large collections from dominating the benchmark.
In contrast, the curated UCI collection and the ZEUS dataset group (a subset of OpenML) contain a more limited number of eligible datasets after the $D$ and $K$ filtering step. 
For these sources, we therefore do not exclude datasets based on size. 
Instead, when $N > 1000$, we uniformly subsample $1000$ instances to match the evaluation scale.
The subsampling is performed with a fixed random seed for reproducibility. 
We verify that the subsampled set preserves the original number of clusters in each dataset.
This procedure preserves dataset diversity across sources while ensuring a controlled and comparable sample size in evaluation.

To characterize task difficulty, we stratify datasets according to empirical clustering performance measured by ARI. 
Let $\max_{m} \mathrm{ARI}_{m}(\mathcal{D})$ denote the maximum ARI achieved on dataset $\mathcal{D}$ across all compared methods $m$. 
Datasets are categorized as \emph{Low Difficulty} if $\max_{m} \mathrm{ARI}_{m}(\mathcal{D}) \geq 0.4$, and \emph{High Difficulty} if $0.1 \leq \max_{m} \mathrm{ARI}_{m}(\mathcal{D}) < 0.4$. 
Datasets with $\max_{m} \mathrm{ARI}_{m}(\mathcal{D}) < 0.1$ are excluded, as such cases provide little evidence of recoverable cluster structure and would confound interpretation of method comparisons.
The threshold $\max_{m}\mathrm{ARI} \geq 0.4$ follows the ZEUS evaluation protocol, where only datasets above this separability level are included. 
Our benchmark subsumes these ZEUS datasets and further extends the evaluation to more challenging regimes by incorporating datasets with lower but non-trivial ARI ($0.1$–$0.4$), thereby covering a broader spectrum of realistic clustering difficulty.

\subsection{Baseline details}
\label{app:baselines}
We first describe the baseline methods. 
Their acronyms are listed below together with brief descriptions.

\begin{itemize}[leftmargin=*]
    \item \textbf{AP}. This method corresponds to \emph{Affinity Propagation}, which forms clusters by iteratively passing messages between pairs of samples until convergence to a set of exemplars. 
    We use the implementation in the \texttt{scikit-learn} package with default parameter settings on all datasets.

    \item \textbf{KMeans}. This method corresponds to the classical \emph{$k$-means} clustering algorithm, which partitions data into $K$ clusters by minimizing the within-cluster sum of squared Euclidean distances. 
    We use the \texttt{scikit-learn} implementation with default settings, where the number of clusters is specified according to the evaluation protocol.

    \item \textbf{GMM}. This method refers to the \emph{Gaussian Mixture Model} with full covariance matrices, where each component is associated with its own general covariance matrix. 
    Model parameters estimates are computed via the expectation–maximization (EM) algorithm. We use the \texttt{scikit-learn} implementation with default configuration unless otherwise specified.

    \item \textbf{S-GMM}. This method denotes a simplified Gaussian mixture model in which all components share the same identity covariance matrix. 
    We adopt the implementation in the official implementation of ZEUS~\citep{marszaek2025zeus}.

    \item \textbf{Agg}. This method corresponds to \emph{Agglomerative Clustering}, a hierarchical bottom-up approach that iteratively merges clusters according to a linkage criterion. 
    We use the \texttt{scikit-learn} implementation with default linkage and distance metric settings.

    \item \textbf{Spec/Spectral}. This method refers to \emph{Spectral Clustering}, which constructs a similarity graph over samples and performs clustering in the space spanned by the leading eigenvectors of the graph Laplacian. 
    We use the \texttt{scikit-learn} implementation with default affinity and eigen-solver settings.

    \item \textbf{Birch}. This method corresponds to \emph{BIRCH} (Balanced Iterative Reducing and Clustering using Hierarchies), which incrementally builds a clustering feature tree to summarize the data and performs clustering based on this compressed representation. 
    We use the \texttt{scikit-learn} implementation with default threshold and branching factor settings.

    \item \textbf{DBSCAN}. This method corresponds to \emph{Density-Based Spatial Clustering of Applications with Noise (DBSCAN)}, which groups together points that lie in high-density regions and labels low-density points as noise based on neighborhood radius and minimum density criteria. 
     We use the \texttt{scikit-learn} implementation with default neighborhood and minimum-sample settings.

    \item \textbf{HDBSCAN}. This method refers to \emph{Hierarchical Density-Based Spatial Clustering of Applications with Noise (HDBSCAN)}, an extension of DBSCAN that constructs a hierarchy of density-based clusters and extracts a stable flat clustering based on cluster persistence. 
    We use the standard \texttt{scikit-learn} library implementation with default parameter settings.

    \item \textbf{OPTICS}. This method corresponds to \emph{Ordering Points To Identify the Clustering Structure (OPTICS)}, which produces an ordering of samples reflecting the underlying density-based cluster structure and enables extraction of clusters at multiple density levels. 
    We use the \texttt{scikit-learn} implementation with default reachability and minimum-sample parameters.

    \item \textbf{MeanShift}. This method refers to \emph{Mean Shift Clustering}, a non-parametric approach that seeks modes of a kernel density estimate by iteratively shifting points toward regions of higher density. 
    We use the \texttt{scikit-learn} implementation with automatic bandwidth estimation and default settings.

    \item \textbf{ZEUS+X}. This class of methods uses the pretrained checkpoint provided by ZEUS~\citep{marszaek2025zeus} to extract feature embeddings, followed by applying a downstream clustering algorithm X in the embedding space. 
    We consider X $\in$ \{\text{$k$-means}, \text{GMM}, \text{S-GMM}\}, following the evaluation protocol adopted in ZEUS.

    \item \textbf{T+X}. 
    The method is proposed in~\cite{li2025clustering}. It follows an iterative pseudo-labeling scheme built on TabPFN v2. First, a subset of samples is drawn from the dataset, and a clustering algorithm X is applied to obtain initial cluster assignments. These samples and their cluster labels are then used as context to fine-tune TabPFN v2, which subsequently predicts cluster labels for all samples, producing a fully pseudo-labeled dataset. Next, a new subset is sampled from this pseudo-labeled dataset and again used as context to fine-tune TabPFN v2, which re-predicts cluster labels for the full dataset. This subsample–condition–predict procedure is repeated iteratively, progressively refining the pseudo labels. We implement this algorithm following the descriptions in the paper.
\end{itemize}

\subsection{Data preprocessing details}
\label{app:data_preprocessing}
Since the datasets contain both categorical and numerical features, preprocessing differs between \ours{} and the baseline methods.

For ZEUS and all baselines implemented in \texttt{scikit-learn}, categorical variables are converted to one-hot representations and concatenated with the numerical features along the feature dimension. 
In contrast, \ours{} does not apply explicit one-hot encoding. During both pretraining and evaluation, categorical variables are processed directly by the model’s encoder, following the treatment used in TabICL.

For numerical features, we apply feature-wise standardization for all methods: each variable is centered by subtracting its sample mean and scaled by its sample standard deviation, resulting in zero mean and unit variance per feature. 
For ZEUS, we additionally follow their preprocessing protocol by applying a MinMax scaling step after standardization, mapping each numerical feature to the range $[0, 1]$.

\subsection{Amortized clustering experiments details}
\label{app:amortized_clustering}

\textbf{Data generation for the amortized clustering comparison}. 
We follow the common 2D synthetic setup in amortized clustering, but use a more diverse overlap-controlled generator to systematically vary cluster separation (see App.~\ref{app:prior_data} for details). Within each minibatch, different datasets may have different ground-truth numbers of clusters $K$. In addition, Gaussian components are sampled with full covariance matrices rather than being restricted to diagonal (or isotropic) covariances, yielding rotated and correlated clusters while keeping the input as an unordered set of points.

\textbf{Training and evaluation protocol}.
We train and evaluate all methods on the same synthetic datasets: at each training step we sample a batch of datasets from the generator above, and evaluation is performed on independently sampled held-out datasets using a different random seed. We train for $1{,}000$ iterations with batch size $16$ and evaluate for $100$ iterations. For NCP and DAC, we use their default optimizer and learning-rate schedules as specified in the original papers; we refer readers to those works for the complete optimization details.

%% file: sections/appendix/additional_results.tex
\section{Additional results}
\label{app:additional_results}
This section provides additional experimental results.
We first report median ARI and NMI values on the evaluation sets
(Sec.~\ref{app:median_ari_nmi_group_by_difficulty}) and ranks of methods with known $K$
on synthetic data (Sec.~\ref{app:synthetic_rank_true_k}).
We then provide a controlled comparison with Cluster-PFN
(Sec.~\ref{app:cluster-pfn}), representation visualizations
(Sec.~\ref{app:representation_visualization}), statistical significance tests
(Sec.~\ref{app:significance}), pairwise real-world comparisons
(Sec.~\ref{app:pairwise_win_rate}), cardinality calibration results
(Sec.~\ref{app:calibration}), and real-world results grouped by knowledge of $K$
(Sec.~\ref{app:rank_group_by_knowledge_on_k}).
We additionally report synthetic distribution shifts and real-world feature-OOD evaluations in Sec.~\ref{app:ood_generalization}.

\subsection{Median ARI and NMI values grouped by levels of difficulty}
\label{app:median_ari_nmi_group_by_difficulty}
In the main paper, we report median ARI and NMI \emph{ranks} to account for varying dataset difficulty. 
Here, we additionally report the median ARI and NMI \emph{values} of each method. 
As described in Sec.~\ref{sec:dataset_stats}, datasets are grouped into two levels of difficulty (low and high), enabling a more controlled comparison. 
Tables~\ref{tab:synthetic_difficulty_complete} and~\ref{tab:real_difficulty_complete} present the corresponding results on synthetic and real-world benchmarks.
Variants marked $^{\dagger}$ use method-specific selection criteria:
\begin{itemize}[leftmargin=*]
\item \textbf{DBSCAN$^{\dagger}$} maximizes density-based clustering validation (DBCV), $\sum_c \frac{|C_c|}{N}\frac{b_c-a_c}{\max(a_c,b_c)}$, where $C_c$ are non-noise clusters, $N$ counts all observations, and $a_c$ and $b_c$ denote within-cluster density sparseness and minimum between-cluster density separation, respectively.
We search 41 log-spaced values of $\varepsilon\in[0.1,10]$ and $\texttt{min\_samples}\in\{3,5,10,20\}$.
\item \textbf{GMM$^{\dagger}$} selects $K\in\{2,\ldots,10\}$ by minimizing the MAP approximation to integrated completed likelihood, $\mathrm{ICL}_{\mathrm{BIC}}(K)=\mathrm{BIC}(K)-2\sum_i\log\max_k\tau_{ik}$, where $\tau_{ik}$ is the posterior component probability.
We use full covariance matrices and one initialization per candidate.
\item \textbf{Spectral$^{\dagger}$} selects $\hat K=\arg\max_{K\in\{2,\ldots,10\}}(\lambda_{K+1}-\lambda_K)$, where $\lambda_1\leq\lambda_2\leq\cdots$ are the eigenvalues of the normalized graph Laplacian constructed using an RBF affinity.
\end{itemize}

\begin{table}[ht]
\centering
\caption{\textbf{Comprehensive performance on synthetic datasets by difficulty}. Results show median [IQR] for ARI and NMI (higher is better), and $k$-MAE. $^{\dagger}$ denotes method-specific selection criteria described in the text.}
\label{tab:synthetic_difficulty_complete}
\setlength{\tabcolsep}{3pt}
\footnotesize
\begin{tabular}{lcccccc}
\toprule
Method & \multicolumn{2}{c}{ARI [IQR] $\uparrow$} & \multicolumn{2}{c}{NMI [IQR] $\uparrow$} & \multicolumn{2}{c}{$k$-MAE $\downarrow$} \\
\cmidrule(lr){2-3} \cmidrule(lr){4-5} \cmidrule(lr){6-7}
& Low & High & Low & High & Low & High \\
\midrule
TabClustPFN$^*$    & \underline{0.8348 [0.22]} & 0.1125 [0.10]  & 0.8012 [0.28]          & 0.1571 [0.14]          & -          & - \\
TabClustPFN        & \textbf{0.8360 [0.24]} & 0.1321 [0.08] & \underline{0.8046 [0.32]} & \underline{0.1940 [0.10]} & \textbf{0} & 5 \\
\midrule
AP                 & 0.0551 [0.10]          & 0.0243 [0.02]          & 0.3169 [0.28]          & 0.1440 [0.06]          & 44         & 45 \\
Agg$^*$            & 0.6586 [0.39]          & 0.1536 [0.07]          & 0.6513 [0.35]          & 0.1562 [0.01]          & -          & - \\
Agg$^+$            & 0.6081 [0.46]          & 0.1278 [0.09]          & 0.6006 [0.40]          & 0.1087 [0.04]          & 1          & 3 \\
Birch$^*$          & 0.6489 [0.41]          & 0.0886 [0.00]          & 0.6536 [0.36]          & 0.1422 [0.02]          & -          & - \\
Birch$^+$          & 0.5836 [0.48]          & 0.0628 [0.03]          & 0.5355 [0.41]          & 0.0947 [0.03]          & 1          & 3 \\
DBSCAN             & 0.0000 [0.00]          & 0.0000 [0.00]          & 0.0000 [0.00]          & 0.0000 [0.02]          & 5          & 6 \\
DBSCAN$^{\dagger}$ & 0.0129 [0.18] & 0.0019 [0.00] & 0.1215 [0.21] & 0.0476 [0.02] & 3 & 10 \\
GMM$^*$            & 0.6990 [0.44]          & \textbf{0.1769 [0.08]} & 0.7442 [0.36]          & \textbf{0.1956 [0.02]} & -          & - \\
GMM$^+$            & 0.7510 [0.48]          & 0.1538 [0.10]          & 0.7142 [0.41]          & 0.1618 [0.05]          & 1          & 4 \\
GMM$^{\dagger}$ & 0.4189 [0.42] & 0.1538 [0.10] & 0.5094 [0.37] & 0.1618 [0.05] & 3 & 4 \\
HDBSCAN            & 0.0000 [0.16]          & -0.0071 [0.01]          & 0.0881 [0.25]          & 0.0056 [0.01]          & 3          & 5 \\
KMeans$^*$         & 0.6838 [0.49]          & 0.1688 [0.10]          & 0.7386 [0.39]          & 0.1883 [0.04]          & -          & - \\
KMeans$^+$         & 0.7350 [0.46]          & \underline{0.1749 [0.09]} & 0.7054 [0.42]          & 0.1886 [0.04]          & 1          & 4 \\
MeanShift          & 0.0000 [0.00]          & -0.0002 [0.00]          & 0.0000 [0.00]          & 0.0018 [0.00]          & 4          & 4 \\
OPTICS             & 0.0000 [0.01]          & -0.0003 [0.00]          & 0.0110 [0.08]          & 0.0209 [0.02]          & 4          & 8 \\
Spectral$^*$       & 0.0682 [0.37]          & 0.1092 [0.11]         & 0.0867 [0.48]          & 0.1326 [0.09]          & -          & - \\
Spectral$^+$       & 0.0192 [0.38]          & 0.1100 [0.11]          & 0.0539 [0.49]          & 0.1143 [0.11]          & 2          & 4 \\
Spectral$^{\dagger}$ & 0.2366 [0.45] & 0.0162 [0.02] & 0.2956 [0.44] & 0.0506 [0.01] & 3 & 3 \\
\midrule
T+Agg$^*$          & 0.7879 [0.35]          & 0.0326 [0.07]          & \textbf{0.8160 [0.33]} & 0.1481 [0.03]          & -          & - \\
T+Agg$^+$          & 0.7326 [0.57]          & -0.0085 [0.03]         & 0.6875 [0.45]          & 0.1077 [0.01]          & 1          & \textbf{2} \\
T+GMM$^*$          & 0.7651 [0.43]          & 0.1644 [0.07]          & 0.7389 [0.37]          & 0.1827 [0.02]          &  -          & - \\
T+GMM$^+$          & 0.7711 [0.59]          & 0.1421 [0.09]          & 0.7013 [0.40]          & 0.1582 [0.04]          & 1 & 4\\
T+KMeans$^*$       & 0.7413 [0.54]          & 0.1642 [0.08]          & 0.7217 [0.45]          & 0.1809 [0.03]          & -          & - \\
T+KMeans$^+$       & 0.7413 [0.54]          & 0.1572 [0.08]          & 0.6457 [0.45]          & 0.1734 [0.03]          & 1          & \textbf{2} \\
T+Spec$^*$         & 0.7939 [0.47]          & 0.1282 [0.05]          & 0.7659 [0.39]          & 0.1505 [0.01]          & -          & - \\
T+Spec$^+$         & 0.7850 [0.66]          & 0.1243 [0.05]          & 0.7131 [0.48]          & 0.1286 [0.13]          & 1          & 4 \\
\midrule
Zeus+GMM$^*$       & 0.7624 [0.47]          & 0.0917 [0.06]          & 0.6914 [0.38]          & 0.1534 [0.13]          & -          & - \\
Zeus+GMM$^+$       & 0.3771 [0.44]          & 0.0908 [0.06]          & 0.4821 [0.30]          & 0.1785 [0.11]          & 6          & 5 \\
Zeus+KMeans$^*$    & 0.7696 [0.47]          & 0.0897 [0.06]          & 0.6939 [0.37]          & 0.1538 [0.13]          & -          & - \\
Zeus+KMeans$^+$    & 0.4041 [0.51]          & 0.0935 [0.06]          & 0.5035 [0.34]          & 0.1771 [0.12]          & 6          & 5 \\
Zeus+S-GMM$^*$     & 0.7684 [0.46]          & 0.0912 [0.06]          & 0.6920 [0.38]          & 0.1534 [0.13]          & -          & - \\
Zeus+S-GMM$^+$     & 0.5602 [0.54]          & 0.0876 [0.07]          & 0.6129 [0.35]          & 0.1781 [0.12]          & 6          & 5 \\
\bottomrule
\end{tabular}
\end{table}

In terms of ARI, \ours{} achieves the strongest overall performance in most settings, with the main exception being high-difficulty synthetic datasets. 
For NMI, \ours{} remains among the top-performing methods across nearly all groups, and stays within the top three even in its weakest regime (low-difficulty real datasets).
Difficulty-stratified analysis reveals several trends. 
First, performance degradation on high-difficulty datasets is universal: even with oracle $K$, classical methods such as GMM$^*$ and KMeans$^*$ exhibit low ARI, indicating that the dominant challenge is recovering assignments under heavy cluster overlap and complex structure, rather than model selection alone. 
Second, in the practical unknown-$K$ setting, \ours{} achieves the best or near-best ARI/NMI on low-difficulty synthetic datasets while accurately recovering $K$, demonstrating good calibration of model complexity. 
On real datasets, \ours{}$^*$ attains the best ARI across difficulty levels, while the automatic-$K$ variant remains competitive, suggesting graceful degradation under distribution shift. 
Third, the performance gap between oracle and inferred $K$ is generally smaller for \ours{} than for classical pipelines, indicating reduced sensitivity to exact model order. 
Finally, in settings where ARI is not strictly best but NMI remains high, errors tend to preserve coarse cluster structure rather than producing highly fragmented partitions.
Overall, these results suggest that the primary strength of \ours{} lies in stable assignment quality and robust $K$ inference in realistic exploratory settings, rather than dependence on favorable parametric assumptions. 
Further gains may be achievable by broadening the pretraining prior or extending training.

\begin{table}[ht]
\centering
\caption{\textbf{Comprehensive performance on real-world benchmark by difficulty}. Results show median [IQR] for ARI and NMI (higher is better), and $k$-MAE. $^{\dagger}$ denotes method-specific selection criteria described in the text.}
\label{tab:real_difficulty_complete}
\setlength{\tabcolsep}{3pt}
\footnotesize
\begin{tabular}{lcccccc}
\toprule
Method & \multicolumn{2}{c}{ARI [IQR] $\uparrow$} & \multicolumn{2}{c}{NMI [IQR] $\uparrow$} & \multicolumn{2}{c}{$k$-MAE $\downarrow$} \\
\cmidrule(lr){2-3} \cmidrule(lr){4-5} \cmidrule(lr){6-7}
& Low & High & Low & High & Low & High \\
\midrule
TabClustPFN$^*$        & \textbf{0.5754 [0.31]} & \textbf{0.1897 [0.21]}    & 0.5737 [0.34] & \underline{0.1614 [0.13]} & - & - \\
TabClustPFN    & 0.4431 [0.29]          & \underline{0.1801 [0.14]} & 0.5311 [0.28]          & \textbf{0.1843 [0.16]} & 1          & 1 \\
\midrule
AP                 & 0.1217 [0.18]          & 0.0298 [0.05]          & 0.3306 [0.29]          & 0.1408 [0.13]          & 27         & 40 \\
Agg$^*$            & 0.5104 [0.27]          & 0.1404 [0.06]          & 0.5665 [0.32]          & 0.0786 [0.05]          & -          & - \\
Agg$^+$            & 0.5019 [0.37]          & 0.1359 [0.11]          & 0.4544 [0.34]          & 0.0915 [0.06]          & \textbf{0} & 2 \\
Birch$^*$          & 0.4157 [0.31]          & 0.1435 [0.08]          & 0.5132 [0.40]          & 0.0920 [0.07]          & -          & - \\
Birch$^+$          & 0.2980 [0.45]          & 0.1435 [0.07]          & 0.4245 [0.43]          & 0.0965 [0.08]          & \textbf{0} & 1 \\
DBSCAN             & 0.0000 [0.20]          & -0.0029 [0.02]         & 0.1158 [0.29]          & 0.0303 [0.06]          & 3          & 2 \\
DBSCAN$^{\dagger}$ & 0.0691 [0.24] & 0.0003 [0.10] & 0.2059 [0.32] & 0.0549 [0.07] & 4 & 3 \\
GMM$^*$            & 0.4453 [0.28]          & 0.1313 [0.19]          & 0.5147 [0.38]          & 0.1068 [0.19]          & -          & - \\
GMM$^+$            & 0.3772 [0.40]          & 0.1430 [0.16]          & 0.4028 [0.38]          & 0.1156 [0.22]          & \textbf{0} & 1 \\
GMM$^{\dagger}$ & 0.3407 [0.27] & 0.1265 [0.15] & 0.4222 [0.30] & 0.1528 [0.22] & 3 & 4 \\
HDBSCAN            & 0.1663 [0.30]          & 0.0042 [0.04]          & 0.2628 [0.28]          & 0.0843 [0.11]          & 4          & 6 \\
KMeans$^*$         & 0.5106 [0.22]          & 0.1362 [0.13]          & 0.5485 [0.33]          & 0.1012 [0.20]          & -          & - \\
KMeans$^+$         & 0.5005 [0.32]          & 0.1179 [0.13]          & 0.4942 [0.34]          & 0.1168 [0.22]          & \textbf{0} & 1 \\
MeanShift          & 0.0707 [0.31]          & 0.0365 [0.21]          & 0.1352 [0.35]          & 0.0535 [0.13]          & 5          & 8 \\
OPTICS             & 0.0168 [0.04]          & -0.0195 [0.04]         & 0.1788 [0.19]          & 0.0979 [0.09]          & 3          & 9 \\
Spectral$^*$       & 0.0578 [0.37]          & 0.0081 [0.06]          & 0.0828 [0.53]          & 0.0190 [0.04]          & -          & - \\
Spectral$^+$       & 0.0786 [0.38]          & 0.0055 [0.01]          & 0.1276 [0.44]          & 0.0142 [0.03]          & 1          & \textbf{0} \\
Spectral$^{\dagger}$ & 0.3012 [0.43] & 0.0751 [0.14] & 0.2849 [0.42] & 0.0628 [0.14] & 4 & 4 \\
\midrule
T+Agg$^*$          & 0.5119 [0.31]          & 0.0867 [0.15]          & 0.5536 [0.29]          & 0.0534 [0.13]          & -          & - \\
T+Agg$^+$          & 0.3200 [0.27]          & 0.0903 [0.07]          & 0.4120 [0.32]          & 0.1424 [0.17]          & 5          & 6 \\
T+GMM$^*$          & 0.4841 [0.27]          & 0.0859 [0.13]          & 0.5535 [0.25]          & 0.0701 [0.16]          & -          & - \\
T+GMM$^+$          & 0.2980 [0.25]          & 0.1063 [0.08]          & 0.4091 [0.32]          & 0.1249 [0.17]          & 5          & 6 \\
T+KMeans$^*$       & 0.5148 [0.29]          & 0.0919 [0.13]          & 0.5183 [0.29]          & 0.0712 [0.12]          & -          & - \\
T+KMeans$^+$       & 0.2741 [0.27]          & 0.0882 [0.12]          & 0.3913 [0.32]          & 0.1217 [0.19]          & 5 & 5 \\
T+Spec$^*$         & 0.4124 [0.26]          & 0.0024 [0.17]          & 0.5141 [0.30]          & 0.0510 [0.14]          & -          & - \\
T+Spec$^+$         & 0.2421 [0.25]          & 0.0489 [0.12]          & 0.3813 [0.32]          & 0.1037 [0.20]          & 6 & 8 \\
\midrule
Zeus+GMM$^*$       & 0.5292 [0.43]          & 0.0501 [0.17]          & \underline{0.6192 [0.40]}          & 0.0713 [0.12]          & -          & - \\
Zeus+GMM$^+$       & 0.3155 [0.38]          & 0.0698 [0.07]          & 0.4377 [0.39]          & 0.1070 [0.08]          & 8 & 8 \\
Zeus+KMeans$^*$    & 0.5209 [0.42]          & 0.0700 [0.26]          & 0.6199 [0.39]          & 0.1126 [0.15]          & -          & - \\
Zeus+KMeans$^+$    & 0.3282 [0.36]          & 0.0615 [0.08]          & 0.4561 [0.39]          & 0.1089 [0.08]          & 8 & 8 \\
Zeus+S-GMM$^*$     & 0.5186 [0.38]          & 0.0834 [0.24]          & \textbf{0.6289 [0.39]}          & 0.1242 [0.14]          & -          & - \\
Zeus+S-GMM$^+$     & 0.4145 [0.48]          & 0.0800 [0.07]          & 0.4670 [0.35] & 0.1081 [0.40]          & 8 & 8 \\
\bottomrule
\end{tabular}
\end{table}

\subsection{Median ARI and NMI values on the synthetic for methods with known $K$}
\label{app:synthetic_rank_true_k}
In the main paper, we report median ARI and NMI ranks when the number of clusters $K$ is inferred from the data. 
Here, we additionally evaluate the setting where the true $K$ is provided to all methods. Tab.~\ref{tab:truek_gen_rank_comparison} reports the median ARI and NMI ranks on synthetic datasets under this oracle-$K$ scenario.
\begin{table}[ht]
\centering
\caption{\textbf{Performance on synthetic datasets}. Median ARI/NMI ranks of methods with known $K$. Brackets show IQR. Bold and underline indicate best and second-best per column.}
\label{tab:truek_gen_rank_comparison}
\begin{tabular}{lcc}
\toprule
Method & ARI Rank [IQR] $\downarrow$ & NMI Rank [IQR] $\downarrow$ \\
\midrule
TabClustPFN*     & \textbf{1 [4]}    & \textbf{3 [5]}    \\
\midrule
Agg*             & 8 [4]             & 9 [4]             \\
Birch*           & 8 [5]             & 9 [4]             \\
GMM*             & 6 [7]             & 5 [5]             \\
KMeans*          & 6 [6]             & 6 [6]             \\
Spectral*        & 13 [2]            & 13 [1]            \\
\midrule
T+Agg*           & \underline{4 [5]} & \textbf{3 [5]}    \\
T+GMM*           & 6 [5]             & 6 [6]             \\
T+KMeans*        & 6 [6]             & 6 [7]             \\
T+Spec*          & \underline{4 [5]} & \underline{4 [6]} \\
\midrule
Zeus+GMM*        & 7 [6]             & 6 [5]             \\
Zeus+KMeans*     & 6 [7]             & 6 [6]             \\
Zeus+S-GMM*      & 6 [7]             & 6 [5]             \\
\bottomrule
\end{tabular}
\end{table}
When the true number of clusters is given, \ours{} also achieves the best median ARI rank and NMI rank overall, indicating that its advantage is not solely due to accurate $K$ inference but also to improved assignment quality.

\subsection{Representation visualization on all synthetic datasets}
\label{app:representation_visualization}
\begin{figure}[!ht]
\centering
\includegraphics[width=0.1\columnwidth]{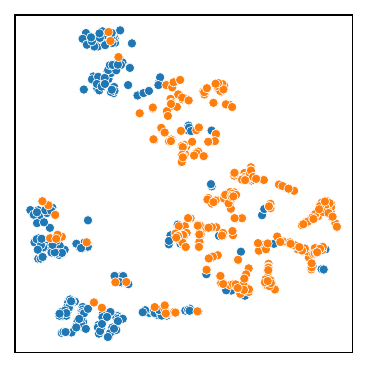}
\includegraphics[width=0.1\columnwidth]{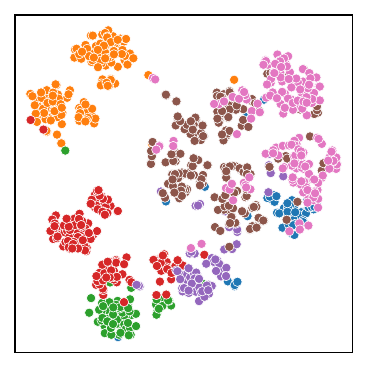}
\includegraphics[width=0.1\columnwidth]{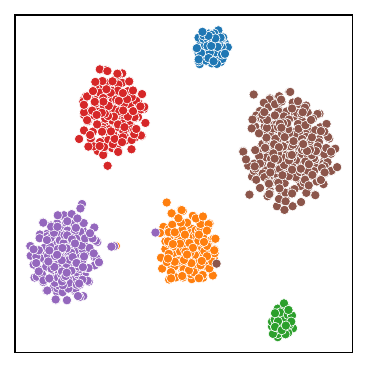}
\includegraphics[width=0.1\columnwidth]{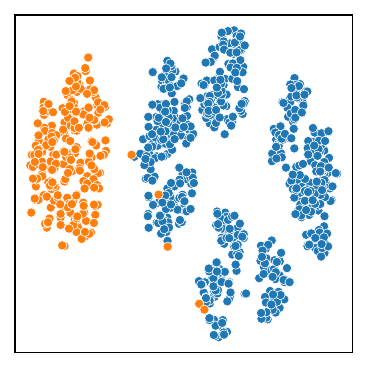}
\includegraphics[width=0.1\columnwidth]{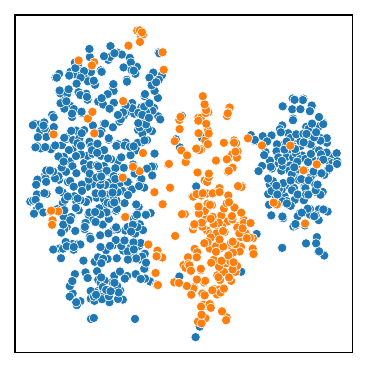}
\includegraphics[width=0.1\columnwidth]{figures/zeus_representations/zeus_zeus_feature_5_original.pdf}
\includegraphics[width=0.1\columnwidth]{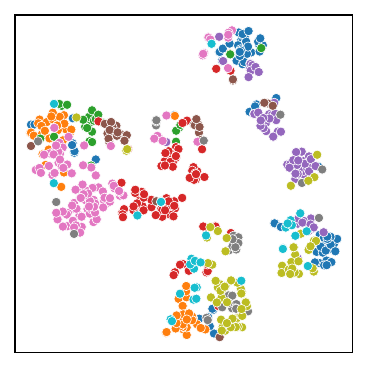}
\includegraphics[width=0.1\columnwidth]{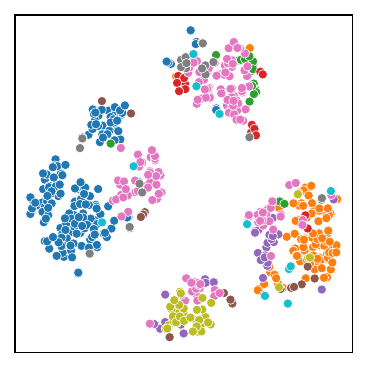}\\
\includegraphics[width=0.1\columnwidth]{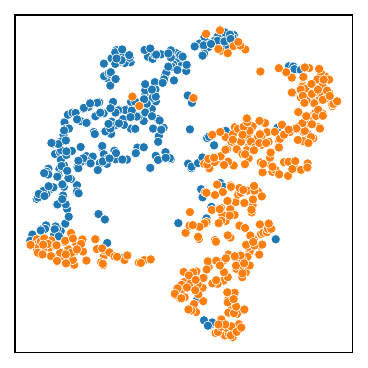}
\includegraphics[width=0.1\columnwidth]{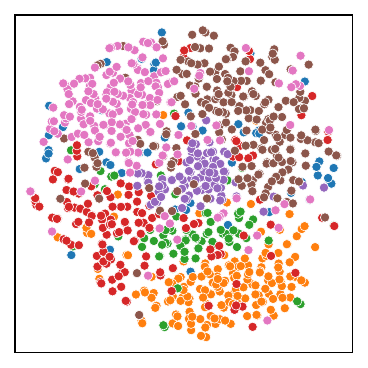}
\includegraphics[width=0.1\columnwidth]{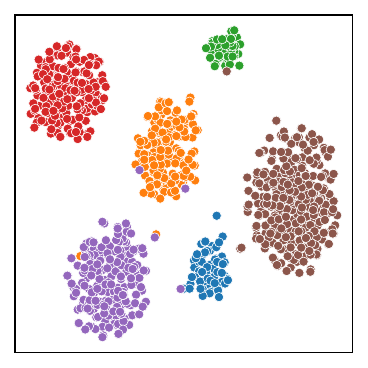}
\includegraphics[width=0.1\columnwidth]{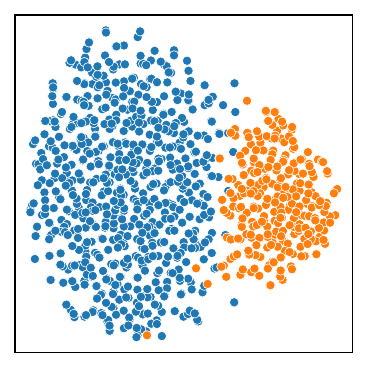}
\includegraphics[width=0.1\columnwidth]{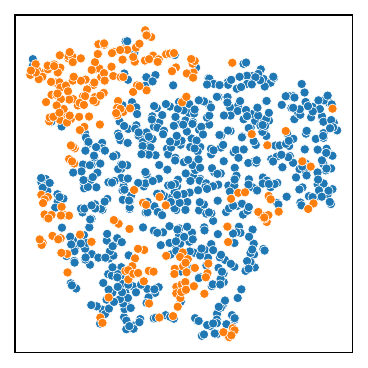}
\includegraphics[width=0.1\columnwidth]{figures/tabclustpfn_representation/zeus_generated_5_tabicl_gt.pdf}
\includegraphics[width=0.1\columnwidth]{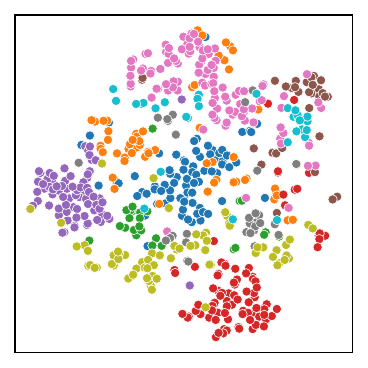}
\includegraphics[width=0.1\columnwidth]{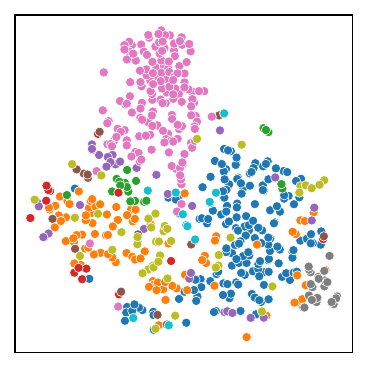}\\
\includegraphics[width=0.1\columnwidth]{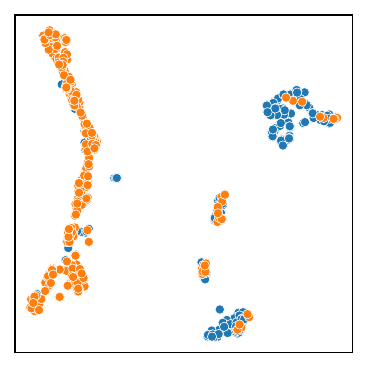}
\includegraphics[width=0.1\columnwidth]{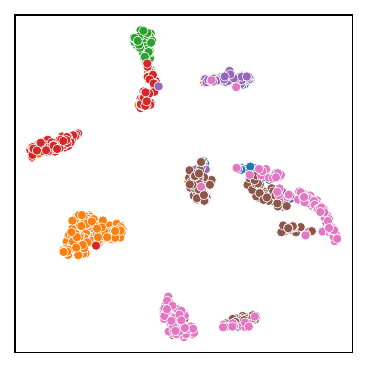}
\includegraphics[width=0.1\columnwidth]{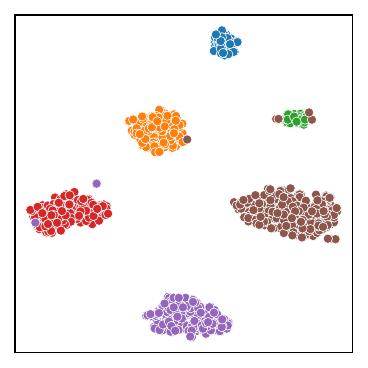}
\includegraphics[width=0.1\columnwidth]{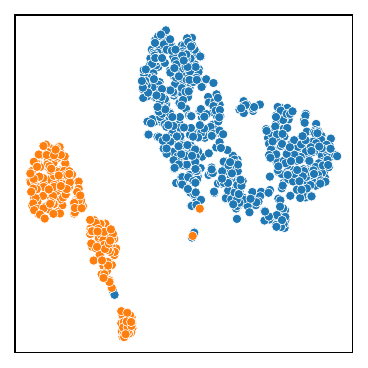}
\includegraphics[width=0.1\columnwidth]{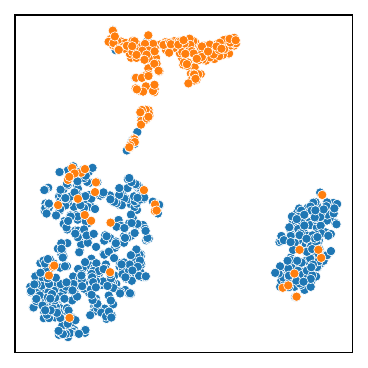}
\includegraphics[width=0.1\columnwidth]{figures/zeus_representations/zeus_zeus_feature_5_repr.pdf}
\includegraphics[width=0.1\columnwidth]{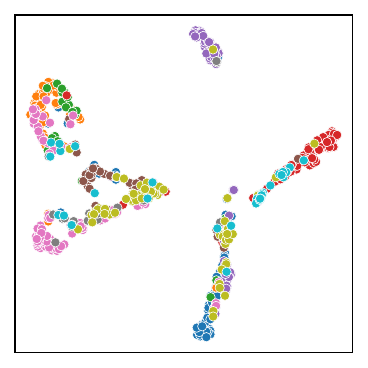}
\includegraphics[width=0.1\columnwidth]{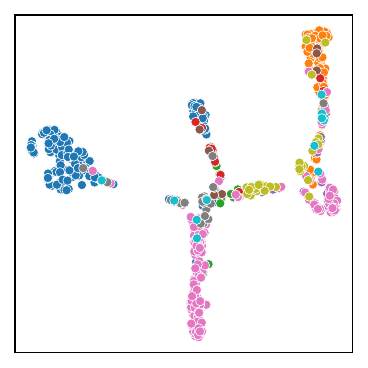}\\
\includegraphics[width=0.1\columnwidth]{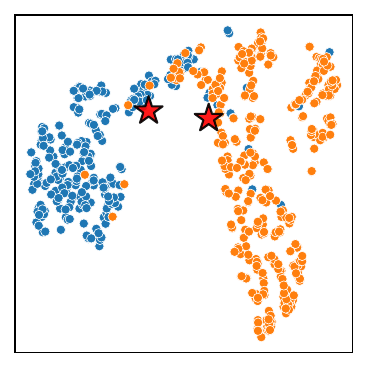}
\includegraphics[width=0.1\columnwidth]{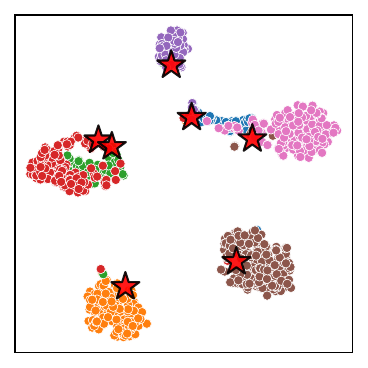}
\includegraphics[width=0.1\columnwidth]{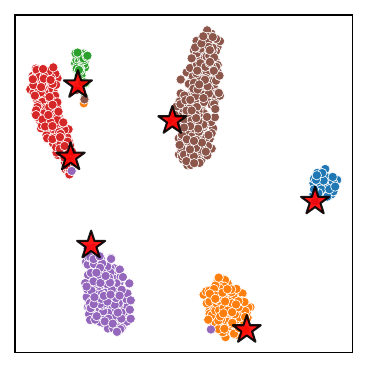}
\includegraphics[width=0.1\columnwidth]{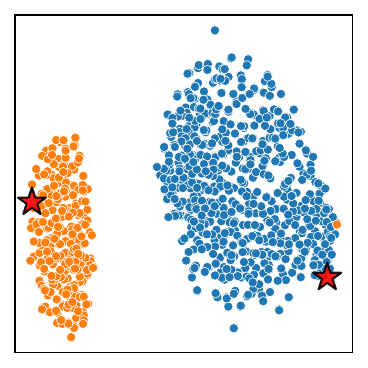}
\includegraphics[width=0.1\columnwidth]{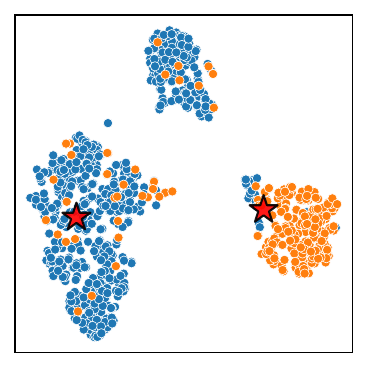}
\includegraphics[width=0.1\columnwidth]{figures/tabclustpfn_representation/zeus_generated_5_ours_gt.pdf}
\includegraphics[width=0.1\columnwidth]{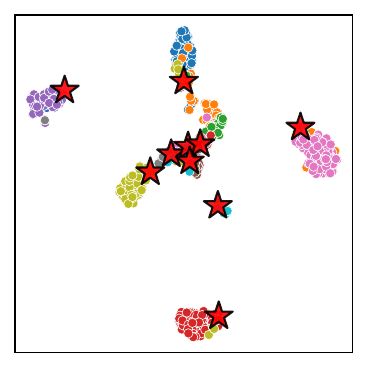}
\includegraphics[width=0.1\columnwidth]{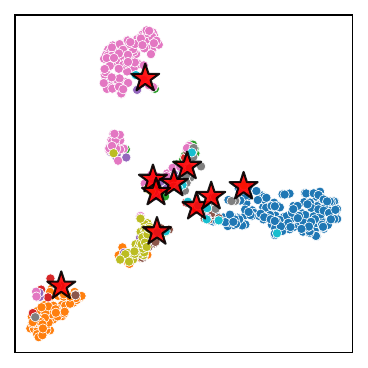}\\
\hrule height 0.4pt width \textwidth
\includegraphics[width=0.1\columnwidth]{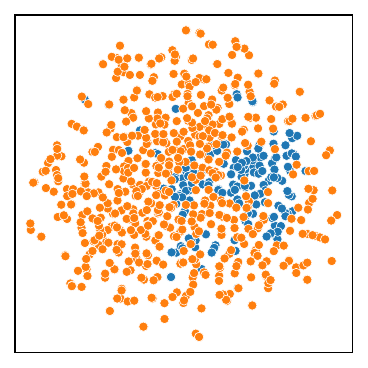}
\includegraphics[width=0.1\columnwidth]{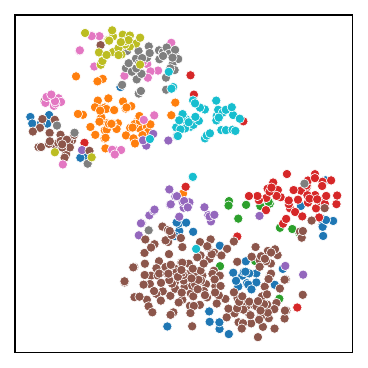}
\includegraphics[width=0.1\columnwidth]{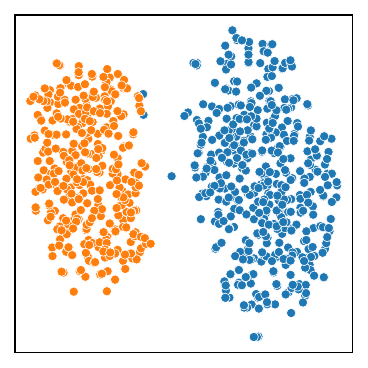}
\includegraphics[width=0.1\columnwidth]{figures/zeus_representations/zeus_zeus_feature_11_original.pdf}
\includegraphics[width=0.1\columnwidth]{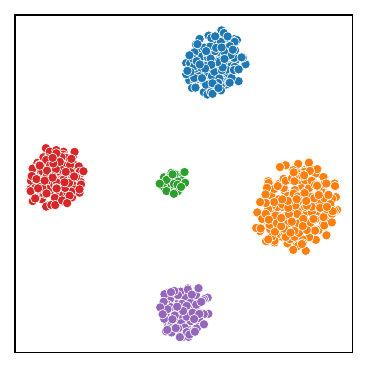}
\includegraphics[width=0.1\columnwidth]{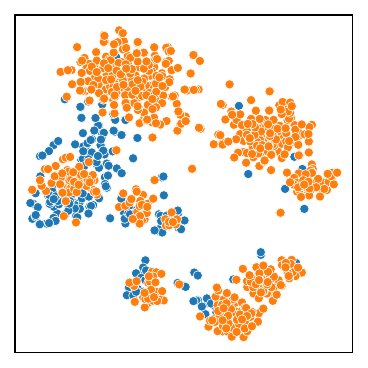}
\includegraphics[width=0.1\columnwidth]{figures/zeus_representations/zeus_zeus_feature_14_original.pdf}
\includegraphics[width=0.1\columnwidth]{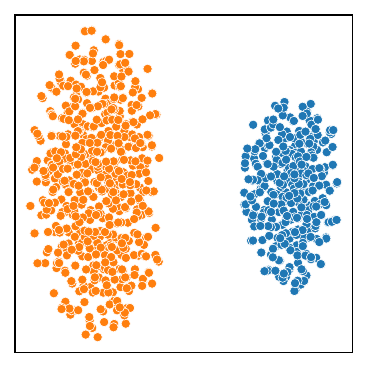}\\
\includegraphics[width=0.1\columnwidth]{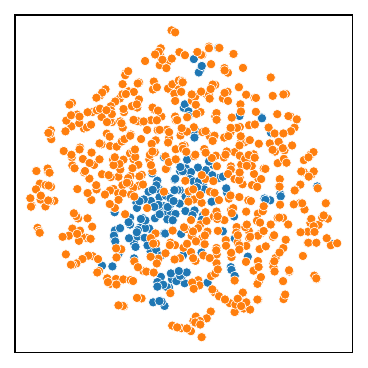}
\includegraphics[width=0.1\columnwidth]{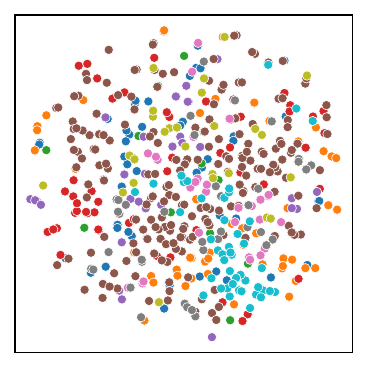}
\includegraphics[width=0.1\columnwidth]{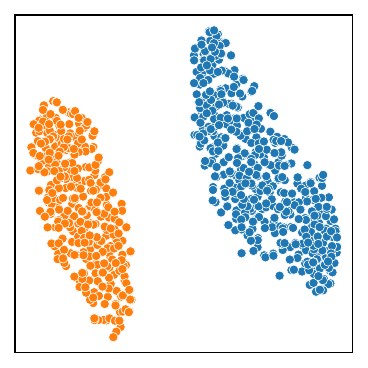}
\includegraphics[width=0.1\columnwidth]{figures/tabclustpfn_representation/zeus_generated_11_tabicl_gt.pdf}
\includegraphics[width=0.1\columnwidth]{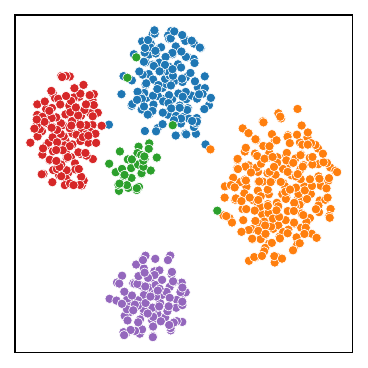}
\includegraphics[width=0.1\columnwidth]{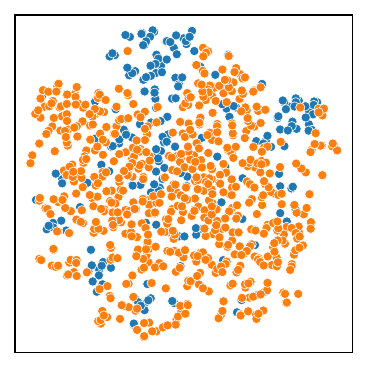}
\includegraphics[width=0.1\columnwidth]{figures/tabclustpfn_representation/zeus_generated_14_tabicl_gt.pdf}
\includegraphics[width=0.1\columnwidth]{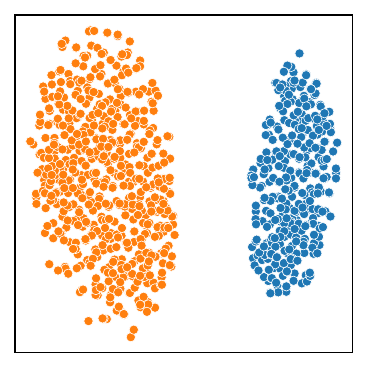}\\
\includegraphics[width=0.1\columnwidth]{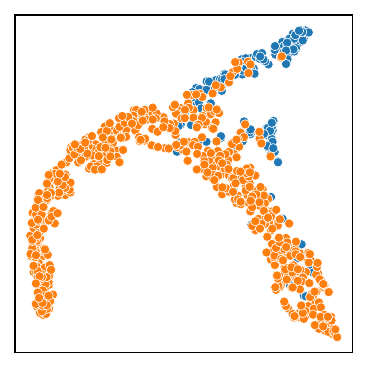}
\includegraphics[width=0.1\columnwidth]{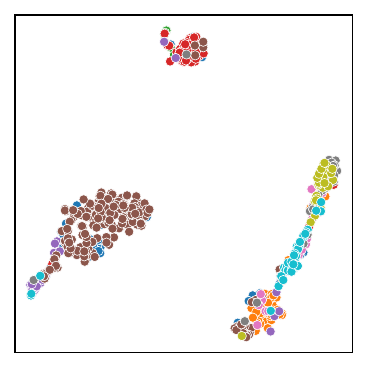}
\includegraphics[width=0.1\columnwidth]{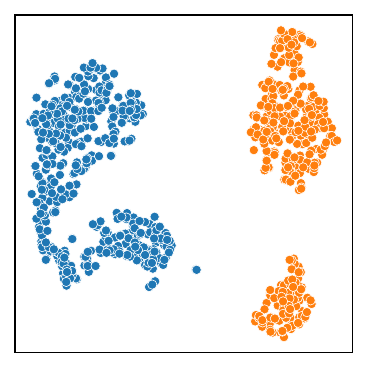}
\includegraphics[width=0.1\columnwidth]{figures/zeus_representations/zeus_zeus_feature_11_repr.pdf}
\includegraphics[width=0.1\columnwidth]{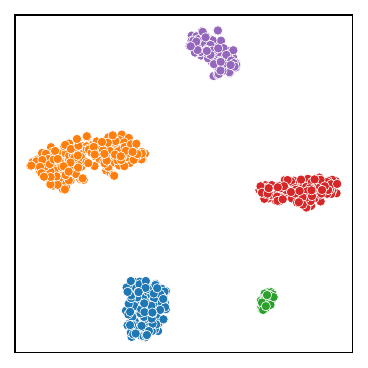}
\includegraphics[width=0.1\columnwidth]{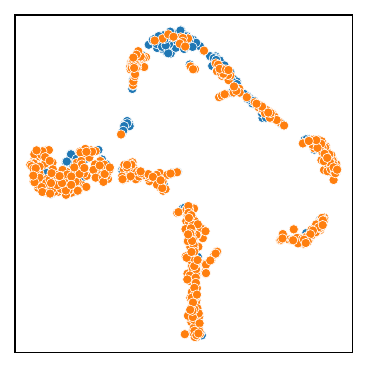}
\includegraphics[width=0.1\columnwidth]{figures/zeus_representations/zeus_zeus_feature_14_repr.pdf}
\includegraphics[width=0.1\columnwidth]{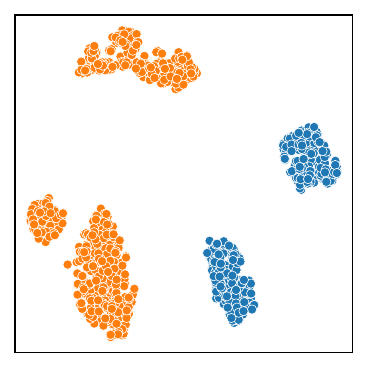}\\
\includegraphics[width=0.1\columnwidth]{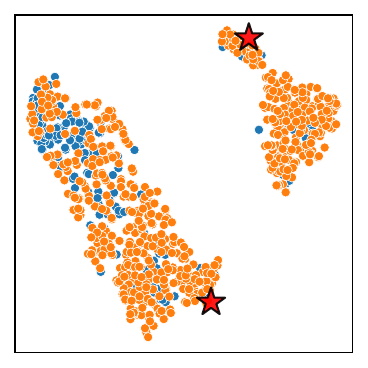}
\includegraphics[width=0.1\columnwidth]{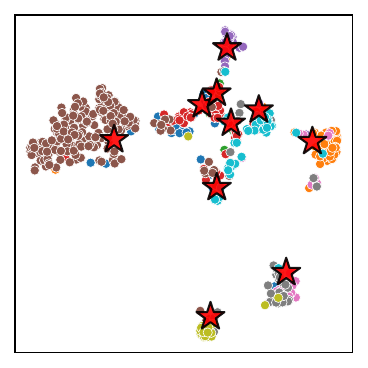}
\includegraphics[width=0.1\columnwidth]{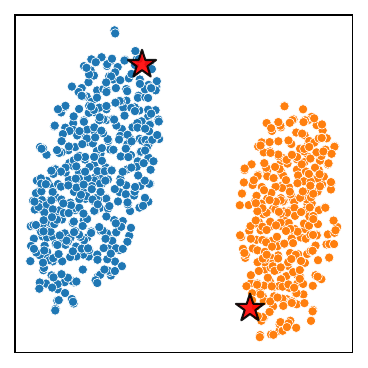}
\includegraphics[width=0.1\columnwidth]{figures/tabclustpfn_representation/zeus_generated_11_ours_gt.pdf}
\includegraphics[width=0.1\columnwidth]{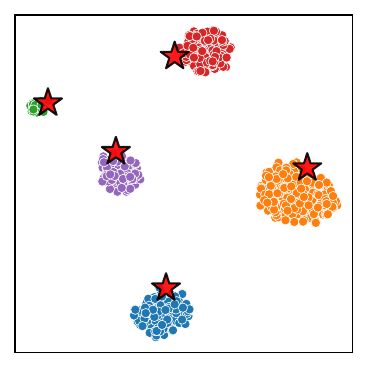}
\includegraphics[width=0.1\columnwidth]{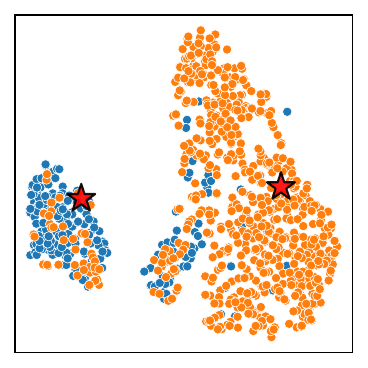}
\includegraphics[width=0.1\columnwidth]{figures/tabclustpfn_representation/zeus_generated_14_ours_gt.pdf}
\includegraphics[width=0.1\columnwidth]{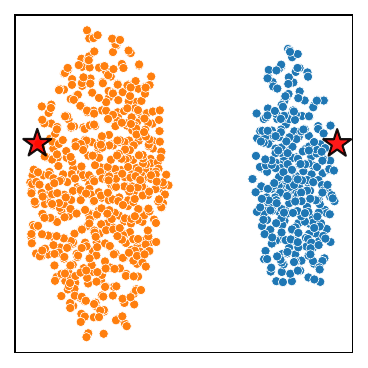}\\
\hrule height 0.4pt width \textwidth
\includegraphics[width=0.1\columnwidth]{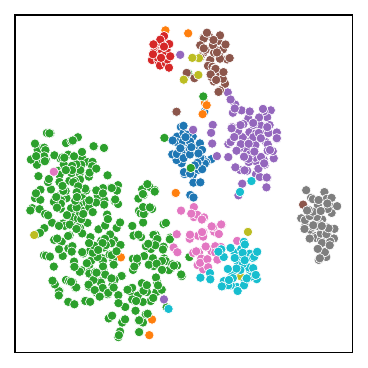}
\includegraphics[width=0.1\columnwidth]{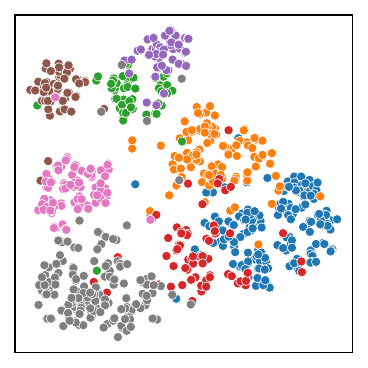}
\includegraphics[width=0.1\columnwidth]{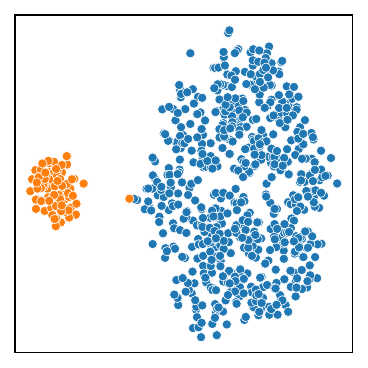}
\includegraphics[width=0.1\columnwidth]{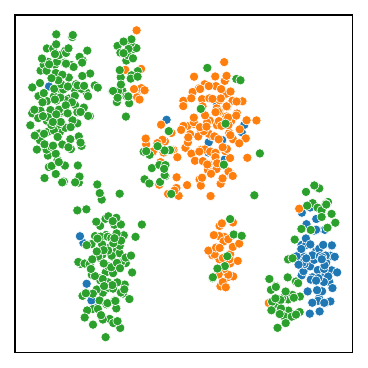}
\includegraphics[width=0.1\columnwidth]{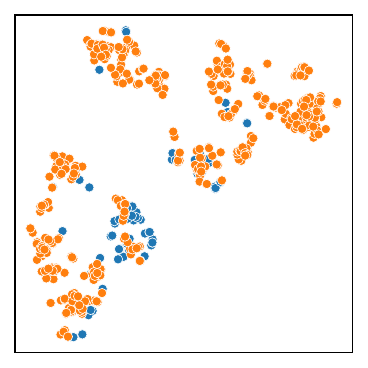}
\includegraphics[width=0.1\columnwidth]{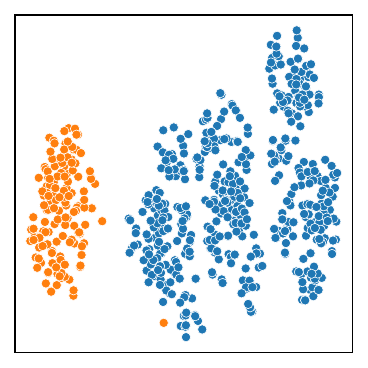}
\includegraphics[width=0.1\columnwidth]{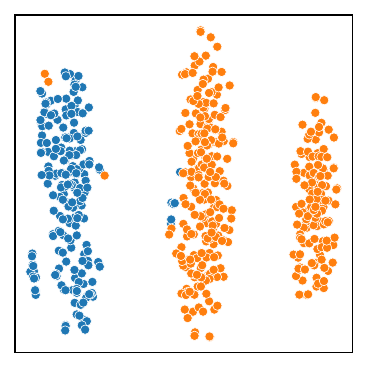}
\includegraphics[width=0.1\columnwidth]{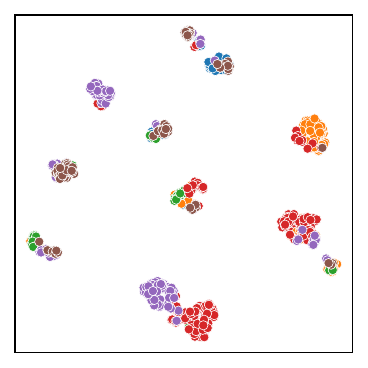}\\
\includegraphics[width=0.1\columnwidth]{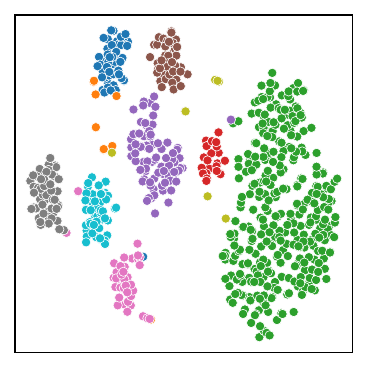}
\includegraphics[width=0.1\columnwidth]{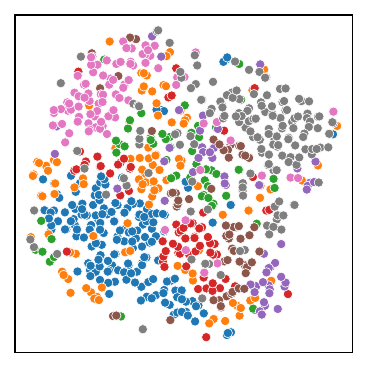}
\includegraphics[width=0.1\columnwidth]{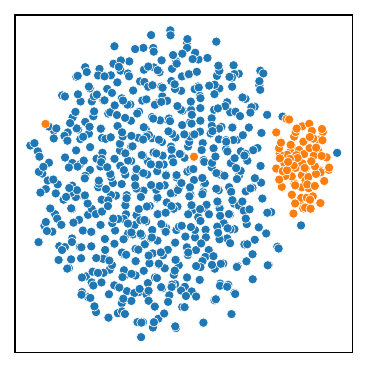}
\includegraphics[width=0.1\columnwidth]{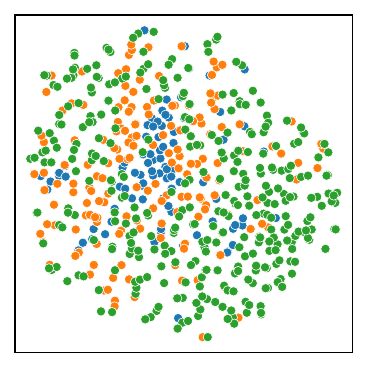}
\includegraphics[width=0.1\columnwidth]{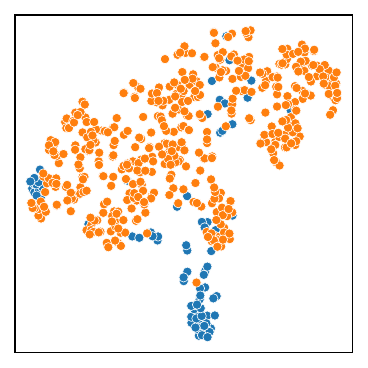}
\includegraphics[width=0.1\columnwidth]{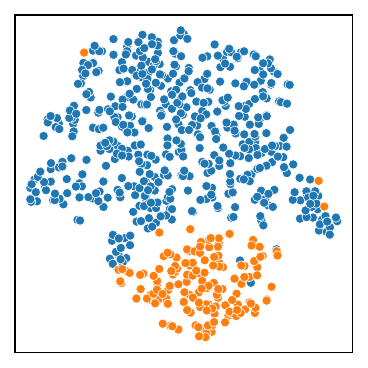}
\includegraphics[width=0.1\columnwidth]{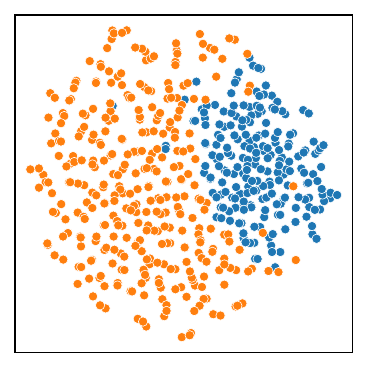}
\includegraphics[width=0.1\columnwidth]{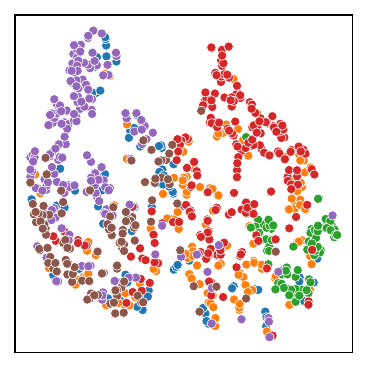}\\
\includegraphics[width=0.1\columnwidth]{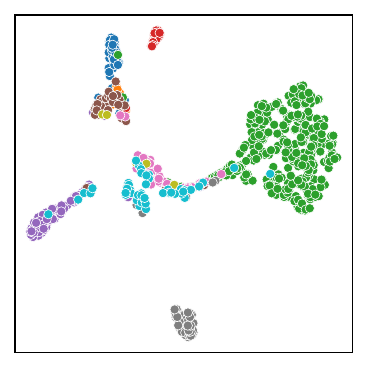}
\includegraphics[width=0.1\columnwidth]{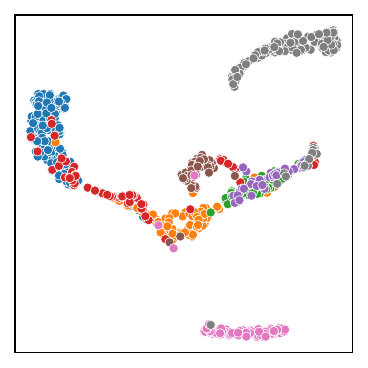}
\includegraphics[width=0.1\columnwidth]{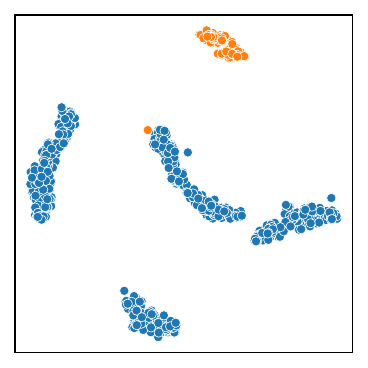}
\includegraphics[width=0.1\columnwidth]{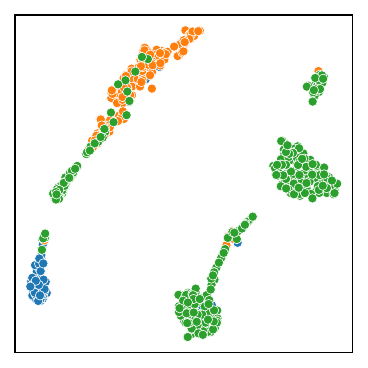}
\includegraphics[width=0.1\columnwidth]{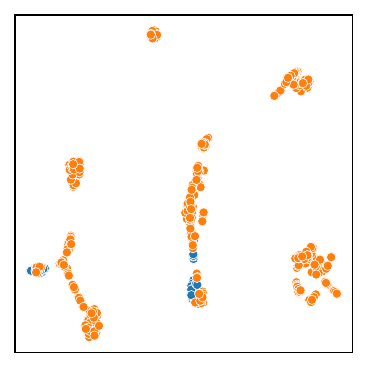}
\includegraphics[width=0.1\columnwidth]{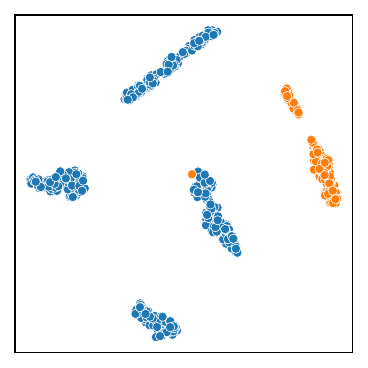}
\includegraphics[width=0.1\columnwidth]{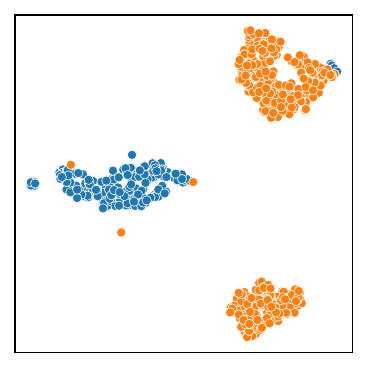}
\includegraphics[width=0.1\columnwidth]{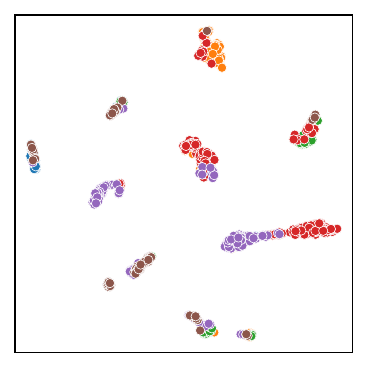}\\
\includegraphics[width=0.1\columnwidth]{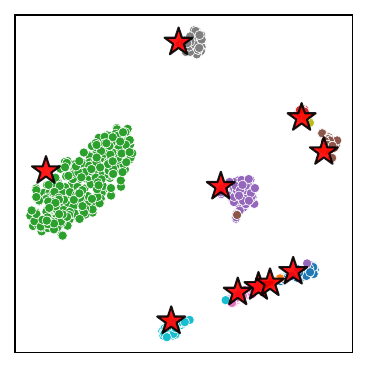}
\includegraphics[width=0.1\columnwidth]{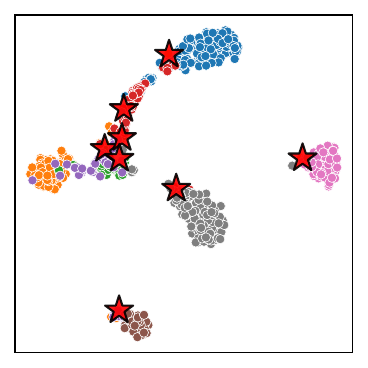}
\includegraphics[width=0.1\columnwidth]{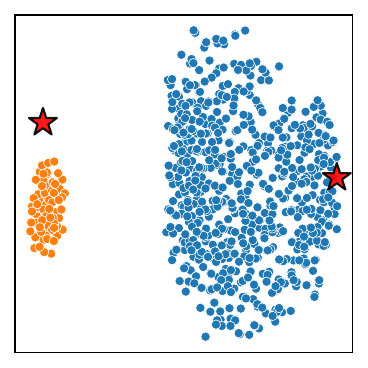}
\includegraphics[width=0.1\columnwidth]{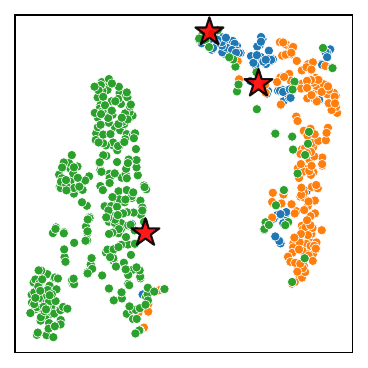}
\includegraphics[width=0.1\columnwidth]{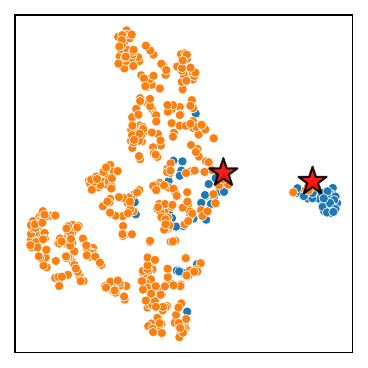}
\includegraphics[width=0.1\columnwidth]{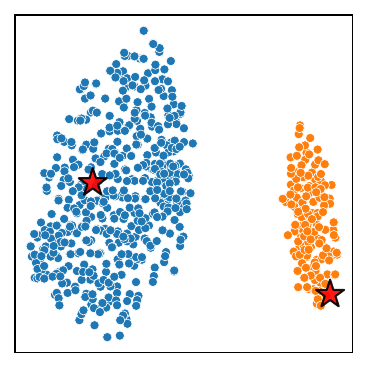}
\includegraphics[width=0.1\columnwidth]{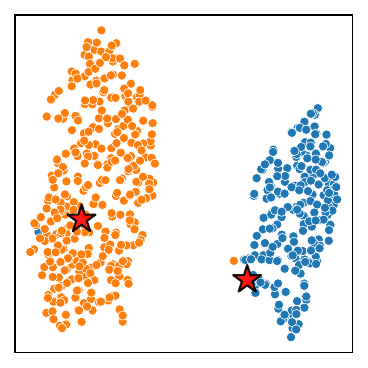}
\includegraphics[width=0.1\columnwidth]{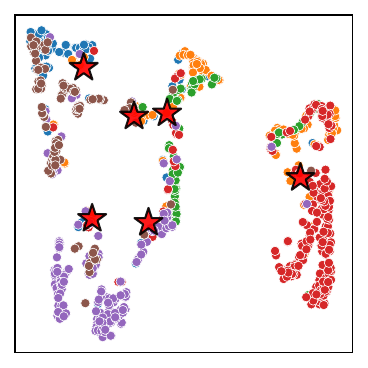}
\caption{\textbf{Representation $t$-SNE visualizations on GMM prior}. Rows are organized in groups of four; within each group, columns correspond to different datasets. From top to bottom, rows display raw features, TabICL encoder outputs, ZEUS representations, and TabClustPFN representations.}
\label{fig:representation_visualization_gmm_all}
\end{figure}

\begin{figure}[!htbp]
\centering
\includegraphics[width=0.1\columnwidth]{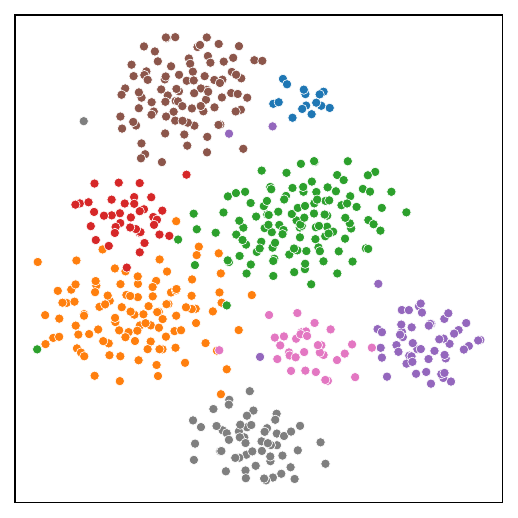}
\includegraphics[width=0.1\columnwidth]{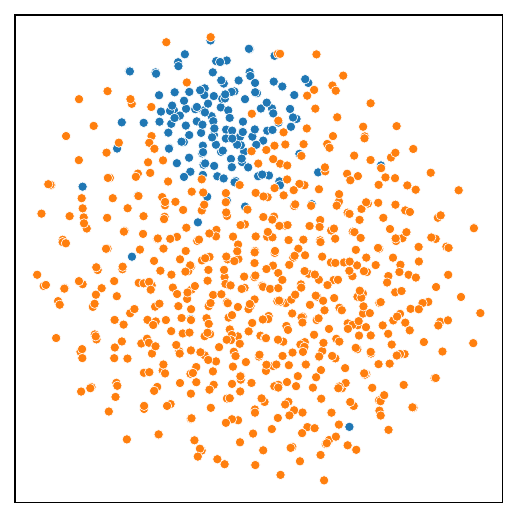}
\includegraphics[width=0.1\columnwidth]{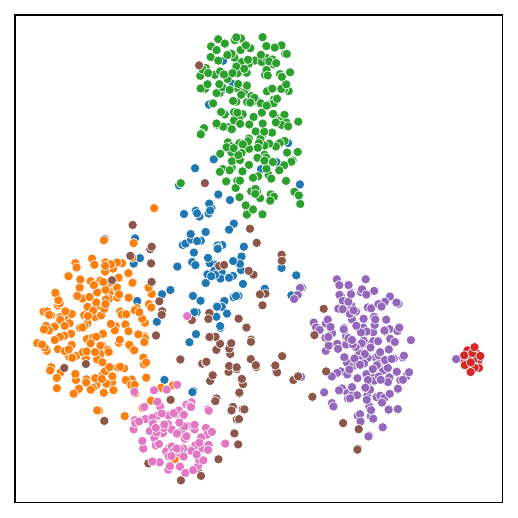}
\includegraphics[width=0.1\columnwidth]{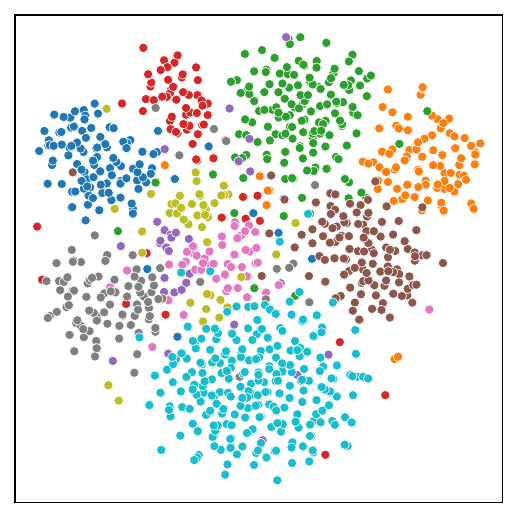}
\includegraphics[width=0.1\columnwidth]{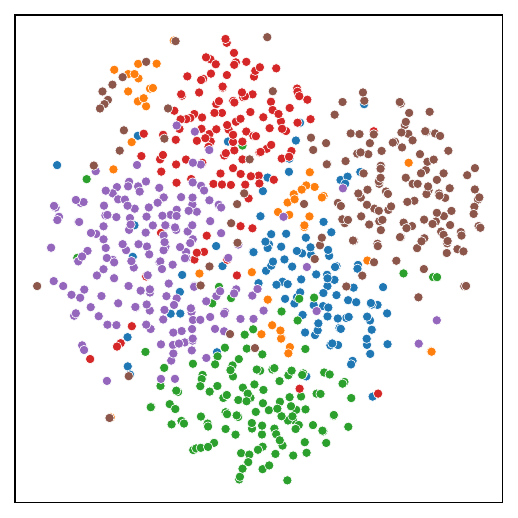}
\includegraphics[width=0.1\columnwidth]{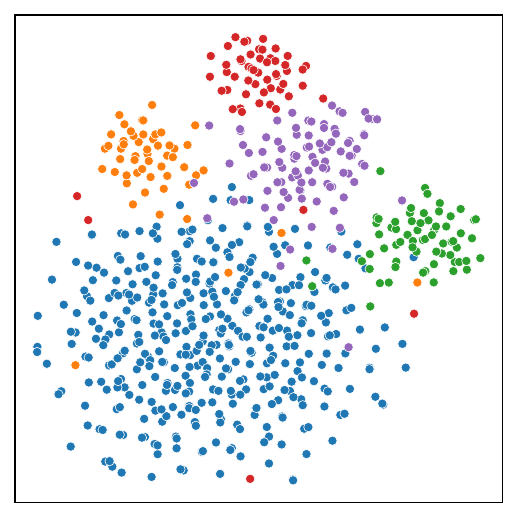}
\includegraphics[width=0.1\columnwidth]{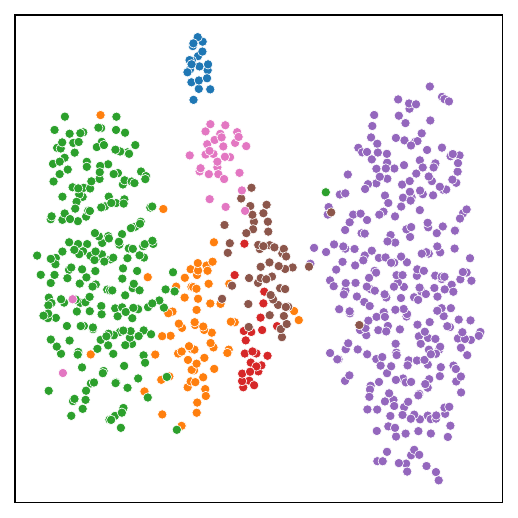}
\includegraphics[width=0.1\columnwidth]{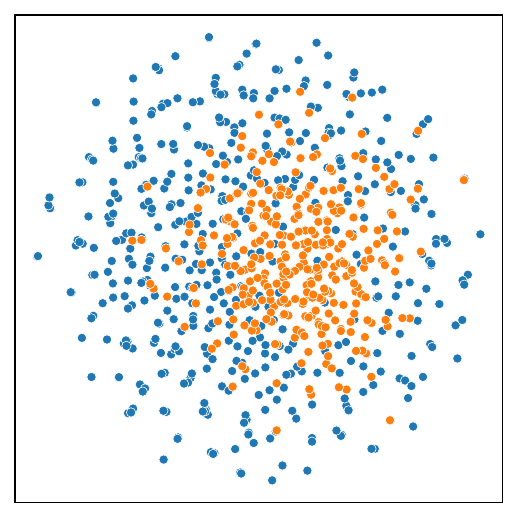}\\
\includegraphics[width=0.1\columnwidth]{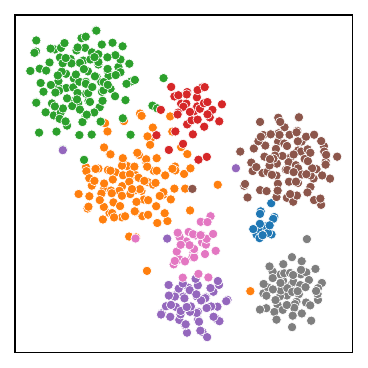}
\includegraphics[width=0.1\columnwidth]{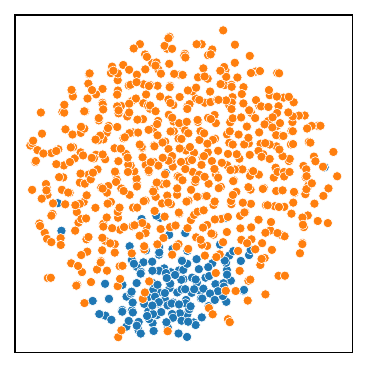}
\includegraphics[width=0.1\columnwidth]{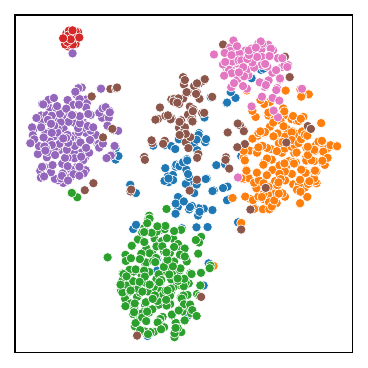}
\includegraphics[width=0.1\columnwidth]{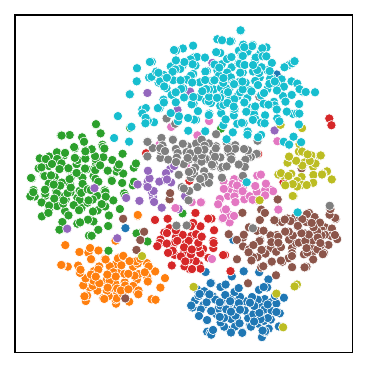}
\includegraphics[width=0.1\columnwidth]{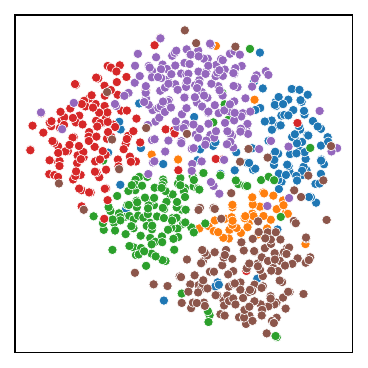}
\includegraphics[width=0.1\columnwidth]{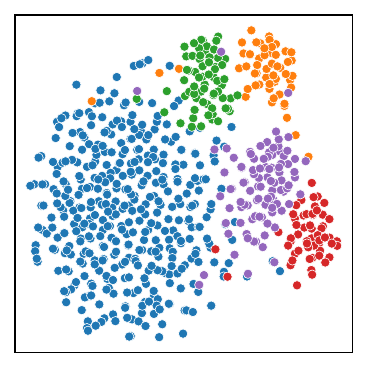}
\includegraphics[width=0.1\columnwidth]{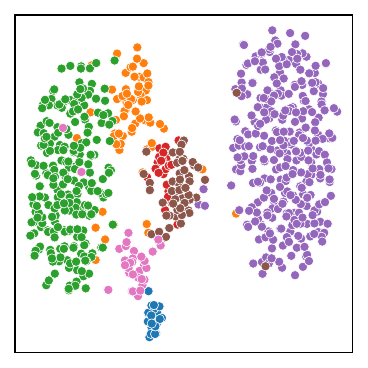}
\includegraphics[width=0.1\columnwidth]{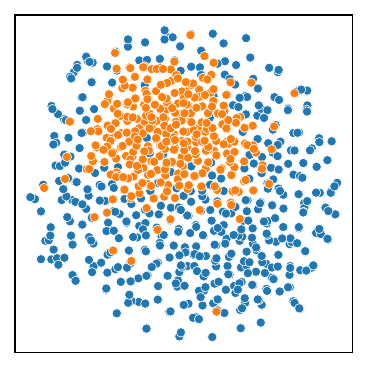}\\
\includegraphics[width=0.1\columnwidth]{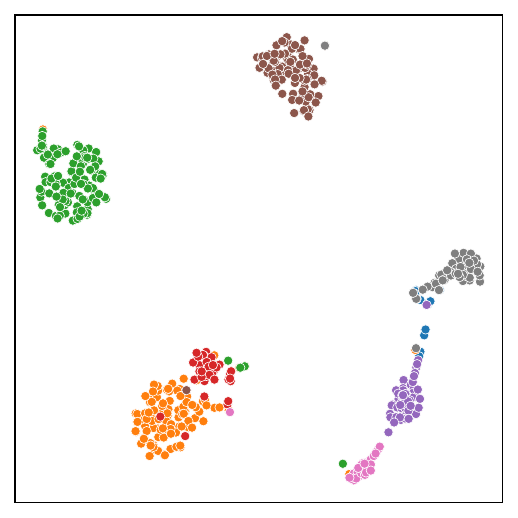}
\includegraphics[width=0.1\columnwidth]{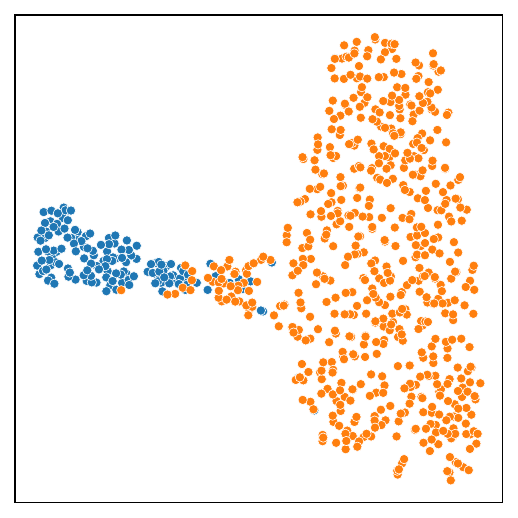}
\includegraphics[width=0.1\columnwidth]{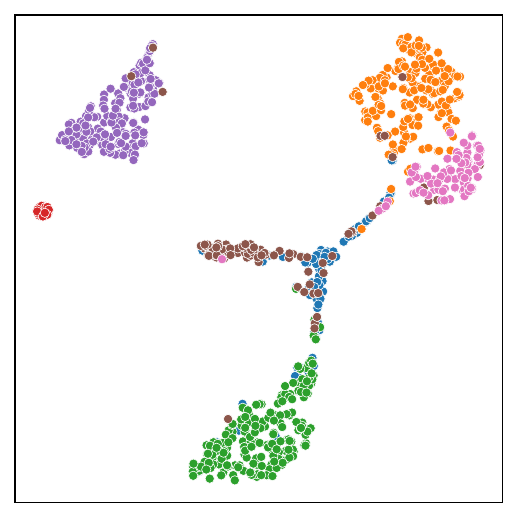}
\includegraphics[width=0.1\columnwidth]{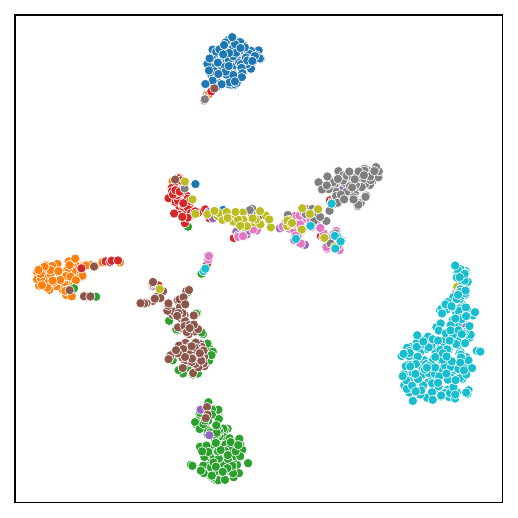}
\includegraphics[width=0.1\columnwidth]{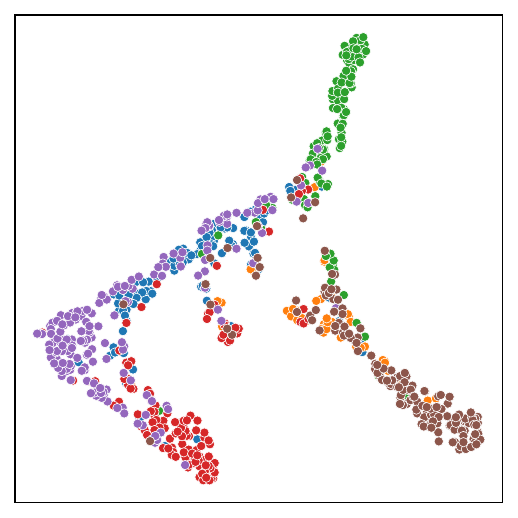}
\includegraphics[width=0.1\columnwidth]{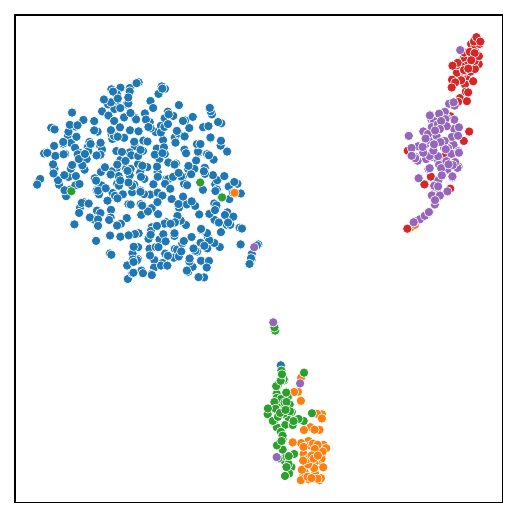}
\includegraphics[width=0.1\columnwidth]{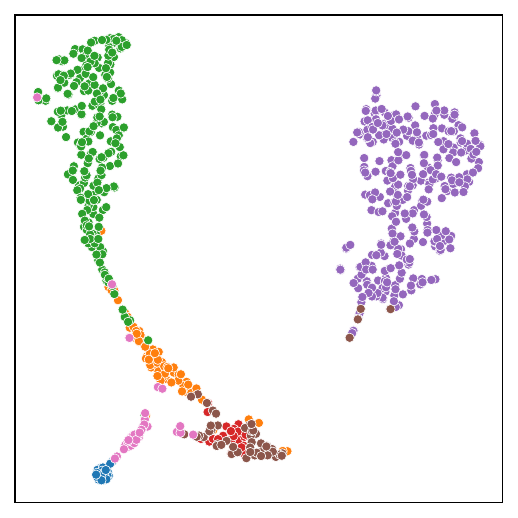}
\includegraphics[width=0.1\columnwidth]{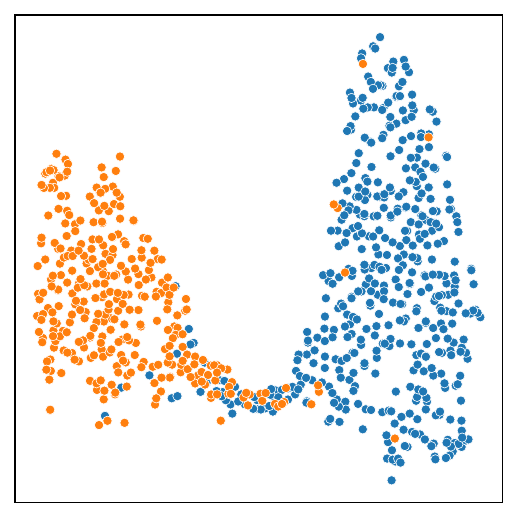}\\
\includegraphics[width=0.1\columnwidth]{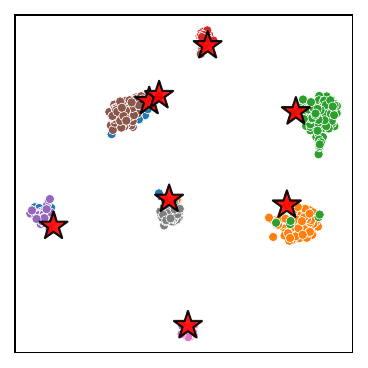}
\includegraphics[width=0.1\columnwidth]{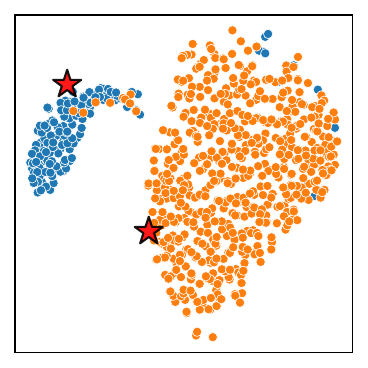}
\includegraphics[width=0.1\columnwidth]{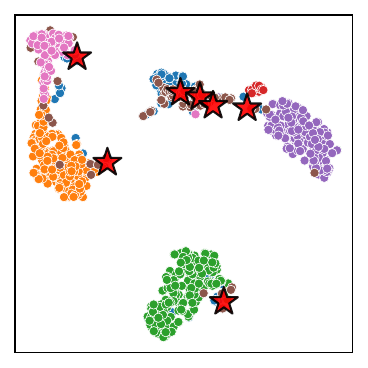}
\includegraphics[width=0.1\columnwidth]{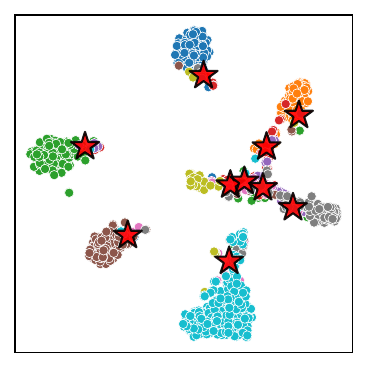}
\includegraphics[width=0.1\columnwidth]{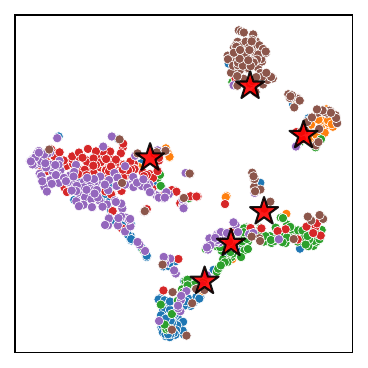}
\includegraphics[width=0.1\columnwidth]{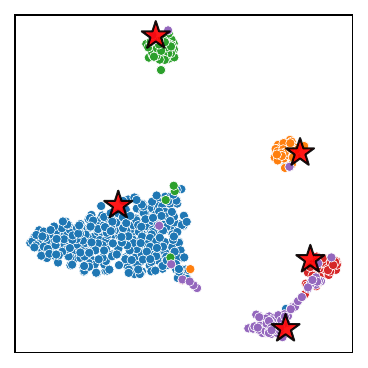}
\includegraphics[width=0.1\columnwidth]{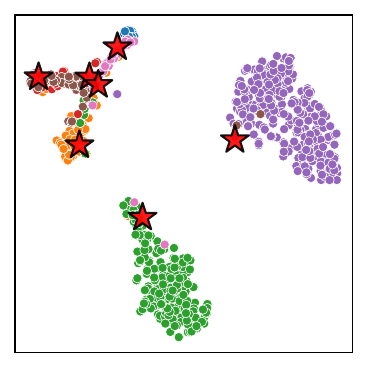}
\includegraphics[width=0.1\columnwidth]{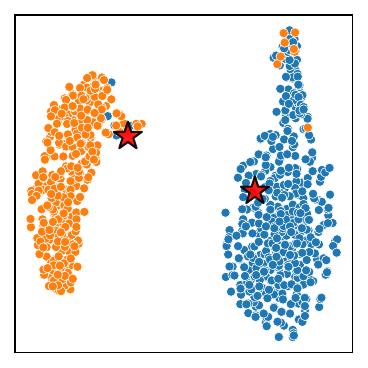}\\
\hrule height 0.4pt width \textwidth
\includegraphics[width=0.1\columnwidth]{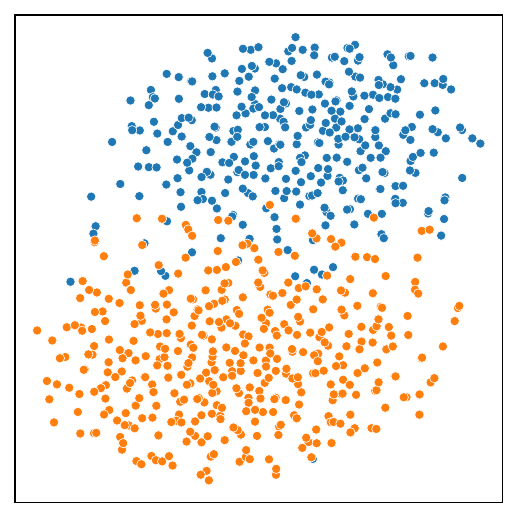}
\includegraphics[width=0.1\columnwidth]{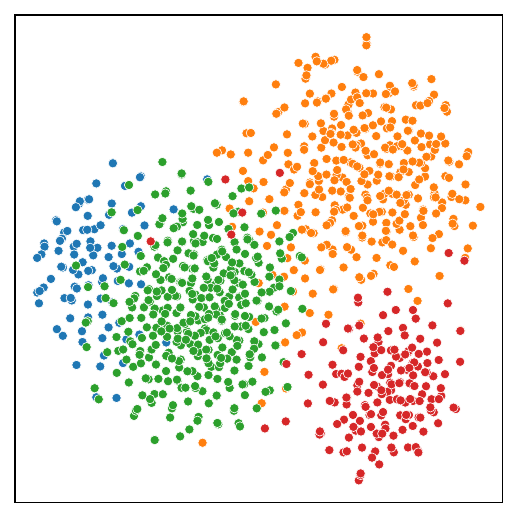}
\includegraphics[width=0.1\columnwidth]{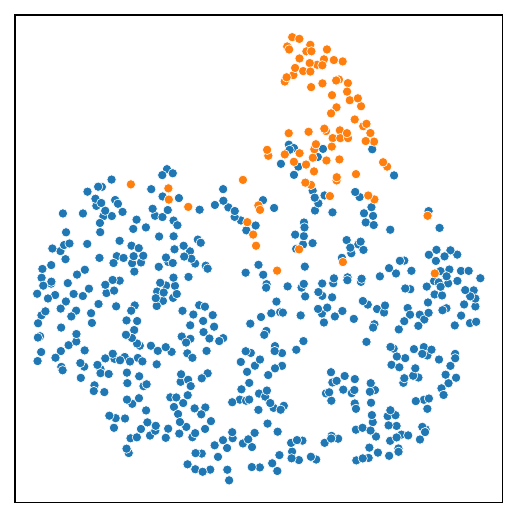}
\includegraphics[width=0.1\columnwidth]{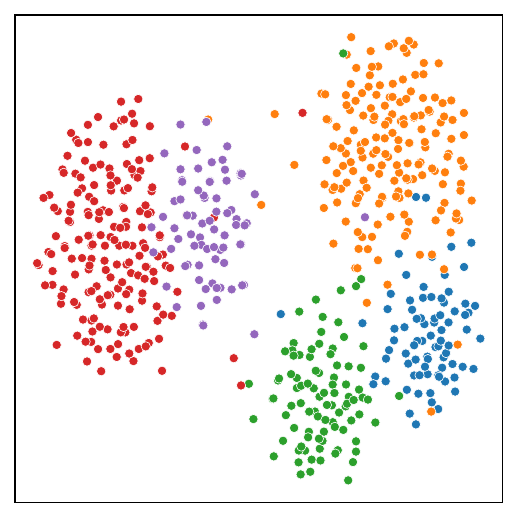}
\includegraphics[width=0.1\columnwidth]{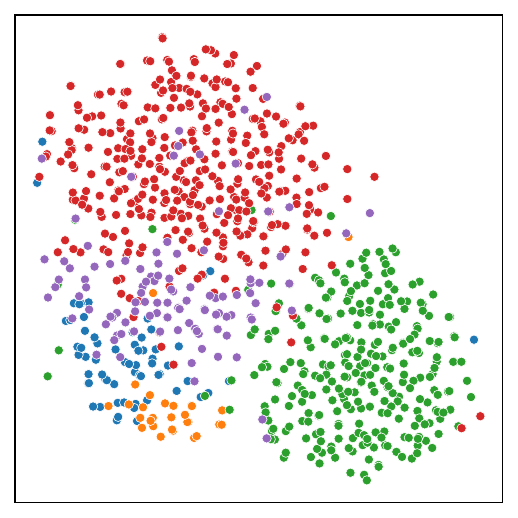}
\includegraphics[width=0.1\columnwidth]{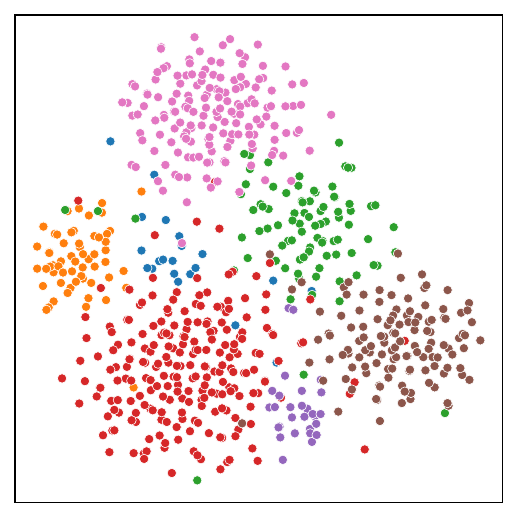}
\includegraphics[width=0.1\columnwidth]{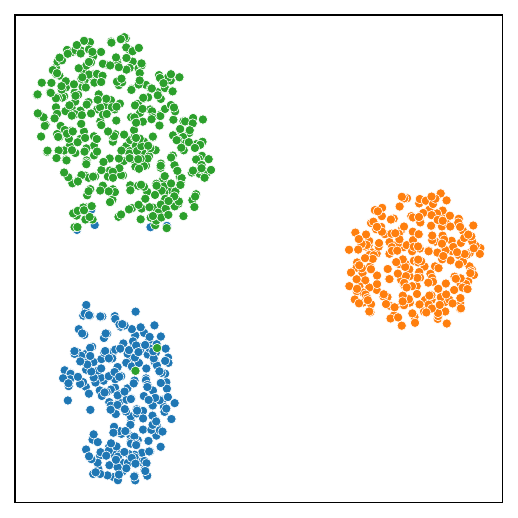}
\includegraphics[width=0.1\columnwidth]{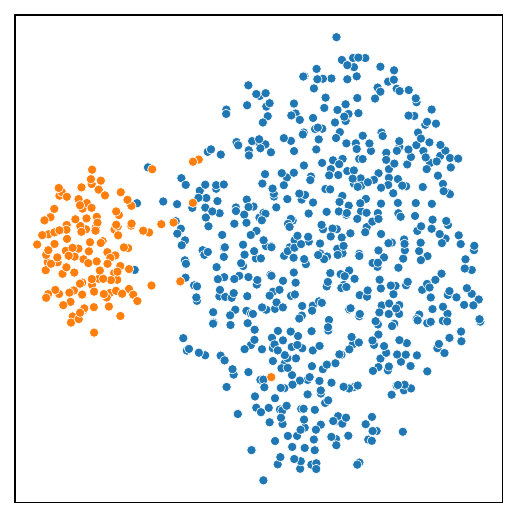}\\
\includegraphics[width=0.1\columnwidth]{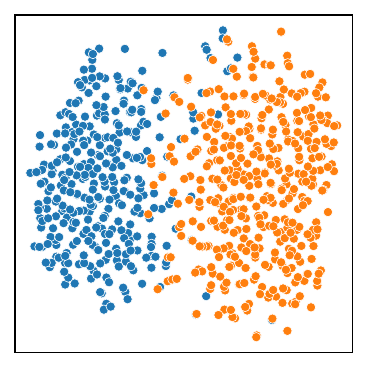}
\includegraphics[width=0.1\columnwidth]{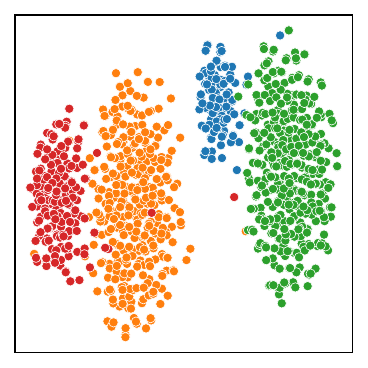}
\includegraphics[width=0.1\columnwidth]{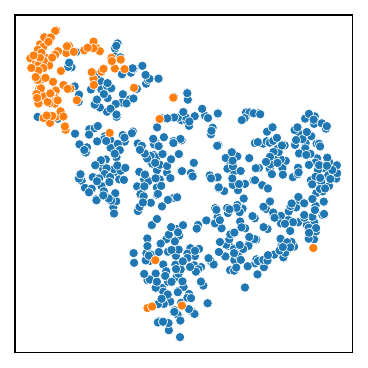}
\includegraphics[width=0.1\columnwidth]{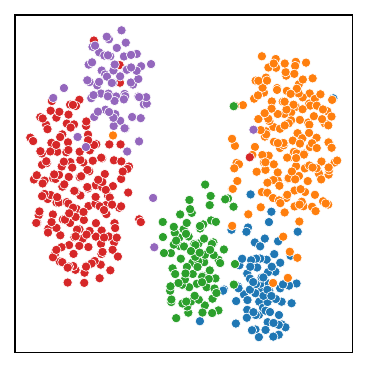}
\includegraphics[width=0.1\columnwidth]{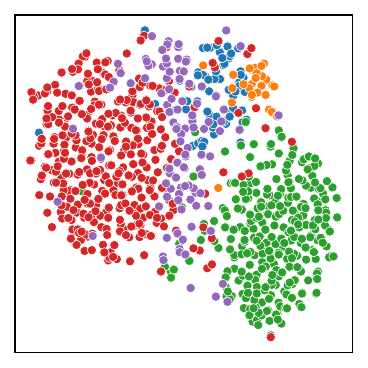}
\includegraphics[width=0.1\columnwidth]{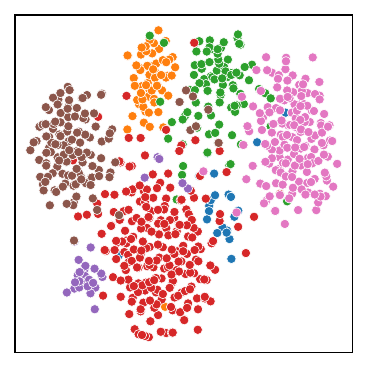}
\includegraphics[width=0.1\columnwidth]{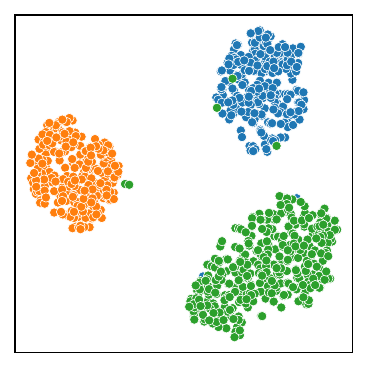}
\includegraphics[width=0.1\columnwidth]{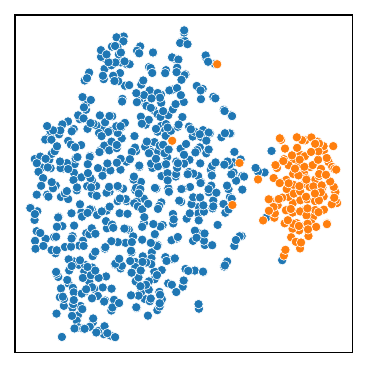}\\
\includegraphics[width=0.1\columnwidth]{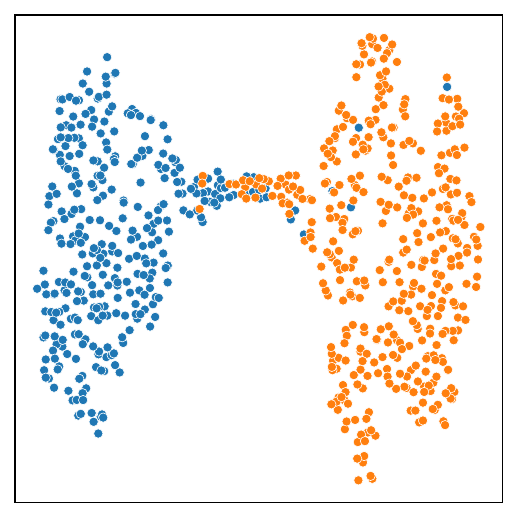}
\includegraphics[width=0.1\columnwidth]{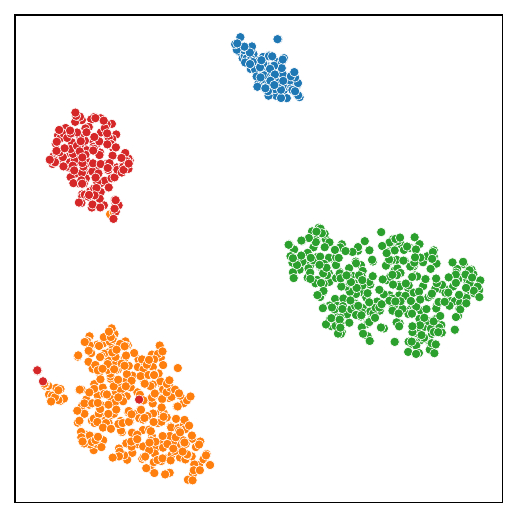}
\includegraphics[width=0.1\columnwidth]{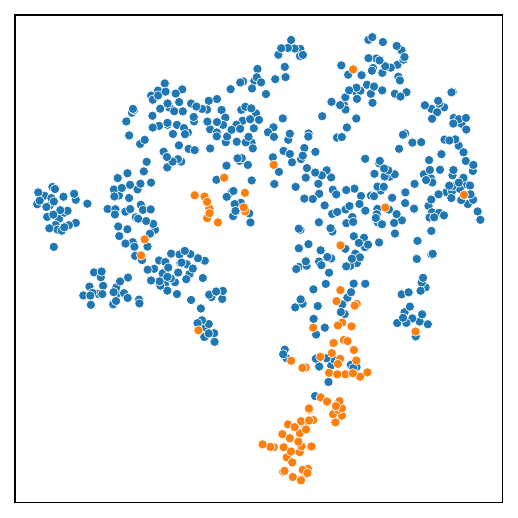}
\includegraphics[width=0.1\columnwidth]{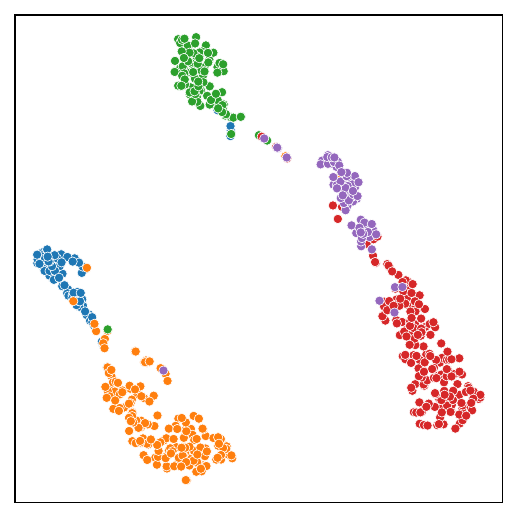}
\includegraphics[width=0.1\columnwidth]{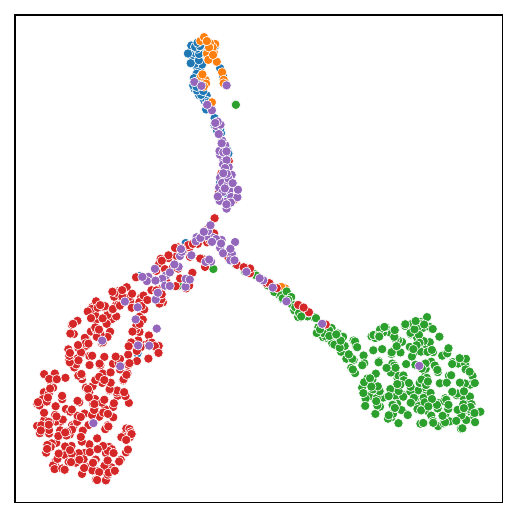}
\includegraphics[width=0.1\columnwidth]{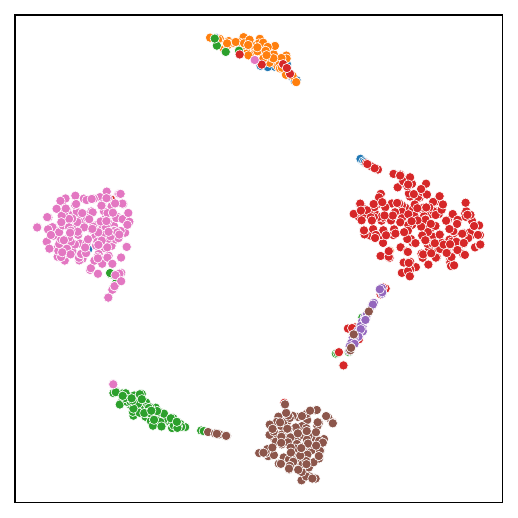}
\includegraphics[width=0.1\columnwidth]{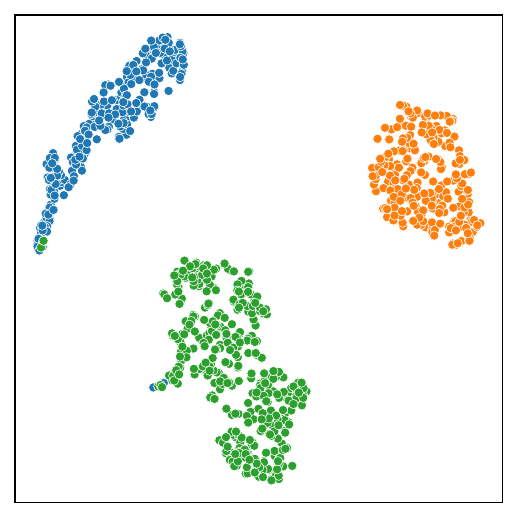}
\includegraphics[width=0.1\columnwidth]{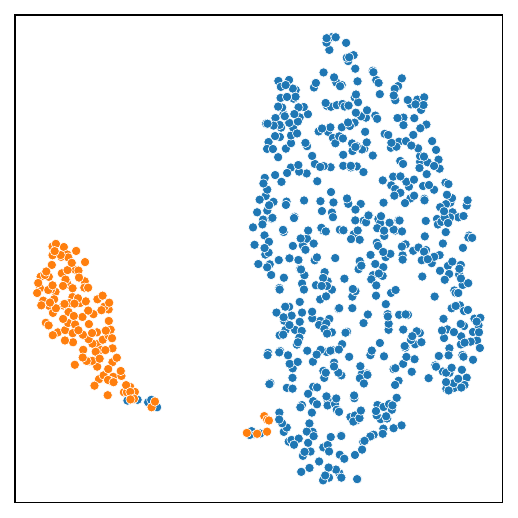}\\
\includegraphics[width=0.1\columnwidth]{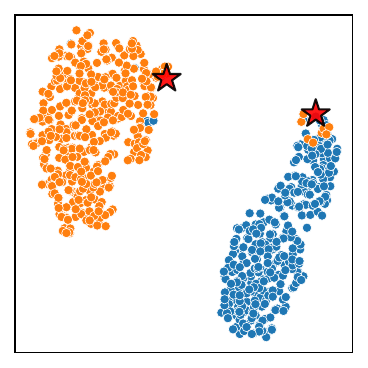}
\includegraphics[width=0.1\columnwidth]{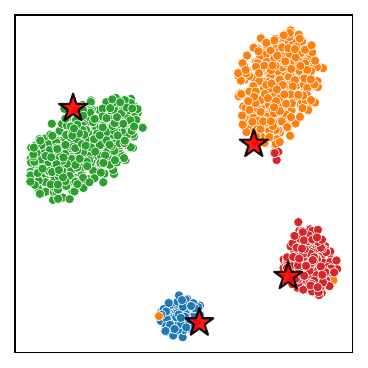}
\includegraphics[width=0.1\columnwidth]{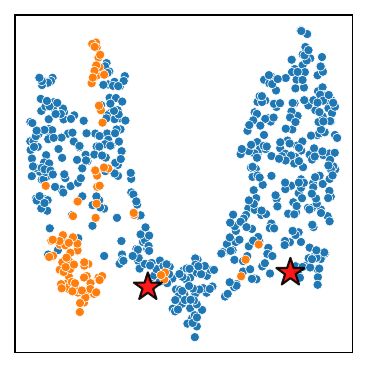}
\includegraphics[width=0.1\columnwidth]{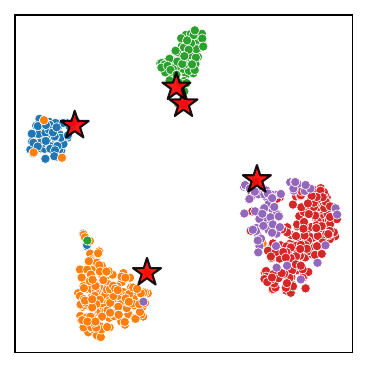}
\includegraphics[width=0.1\columnwidth]{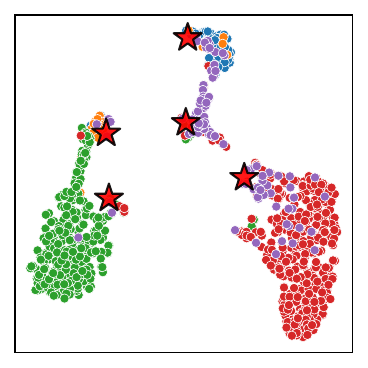}
\includegraphics[width=0.1\columnwidth]{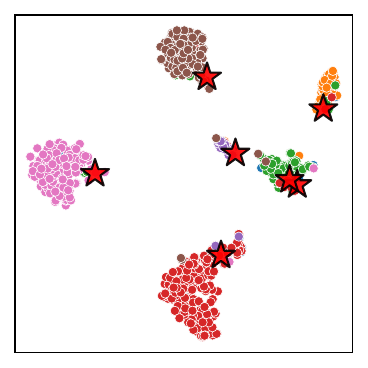}
\includegraphics[width=0.1\columnwidth]{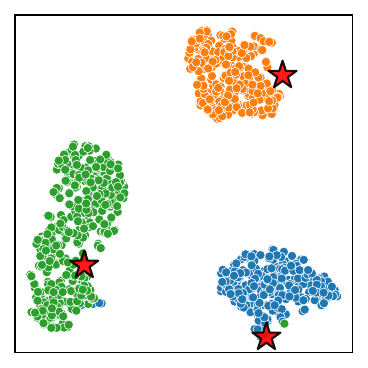}
\includegraphics[width=0.1\columnwidth]{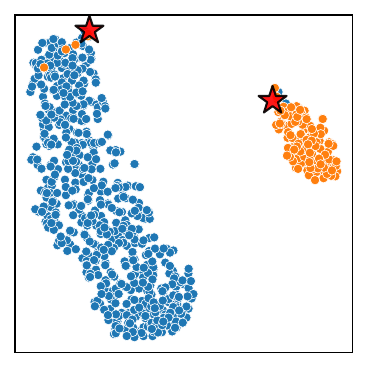}\\
\hrule height 0.4pt width \textwidth
\includegraphics[width=0.1\columnwidth]{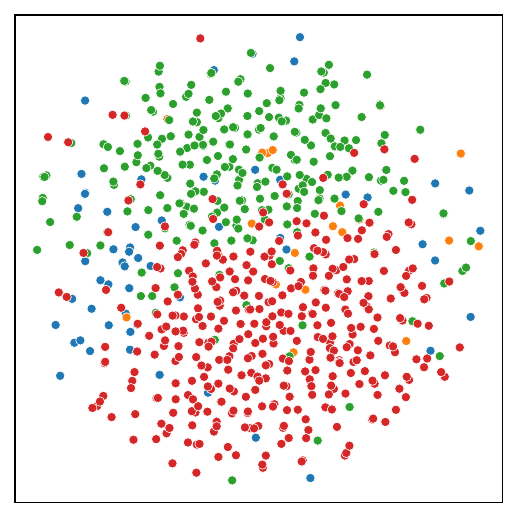}
\includegraphics[width=0.1\columnwidth]{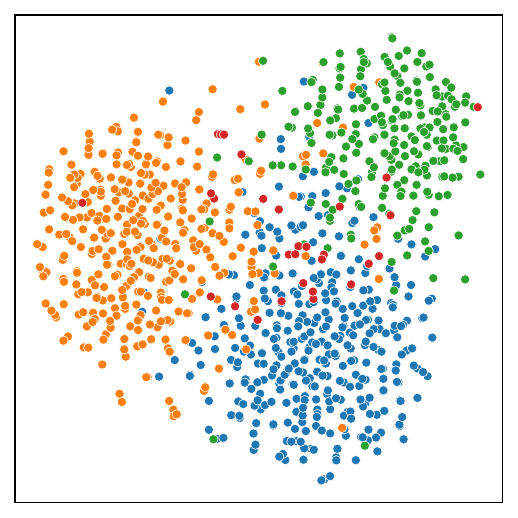}
\includegraphics[width=0.1\columnwidth]{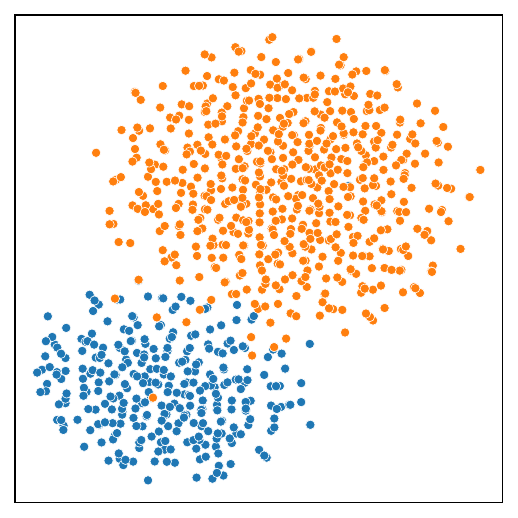}
\includegraphics[width=0.1\columnwidth]{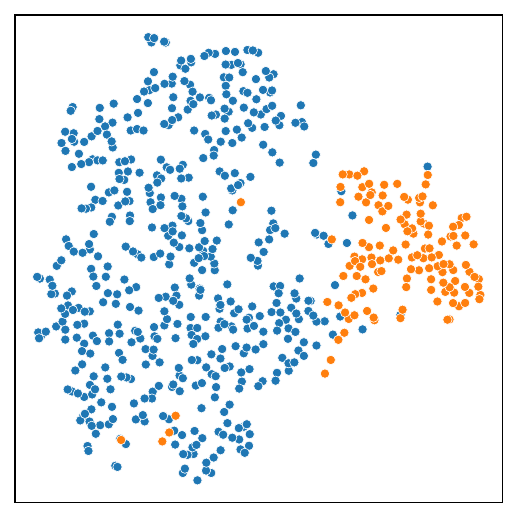}
\includegraphics[width=0.1\columnwidth]{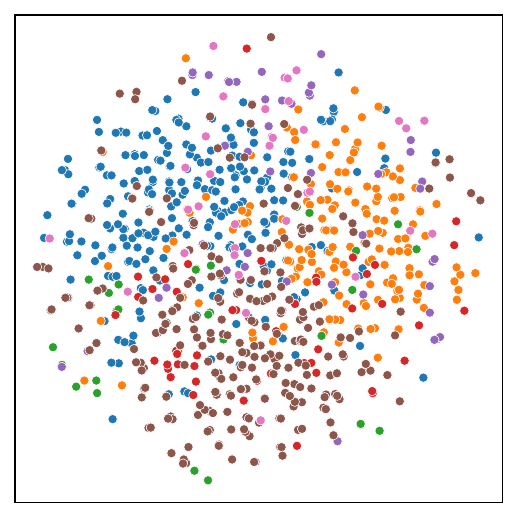}
\includegraphics[width=0.1\columnwidth]{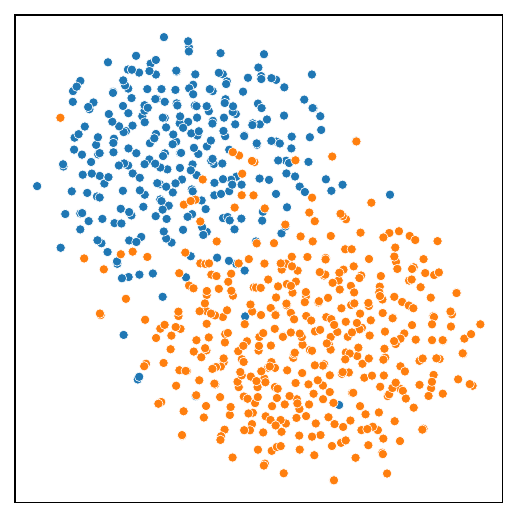}
\includegraphics[width=0.1\columnwidth]{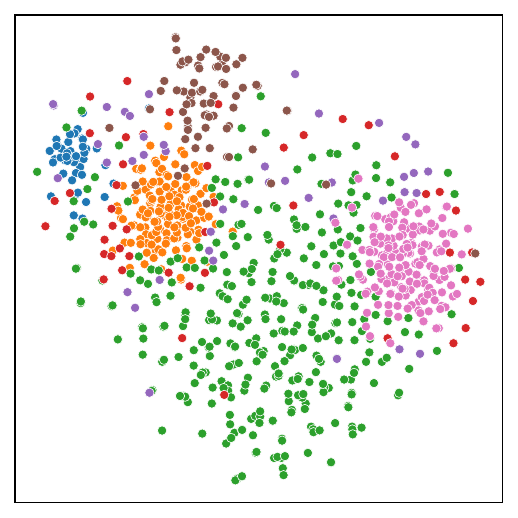}
\includegraphics[width=0.1\columnwidth]{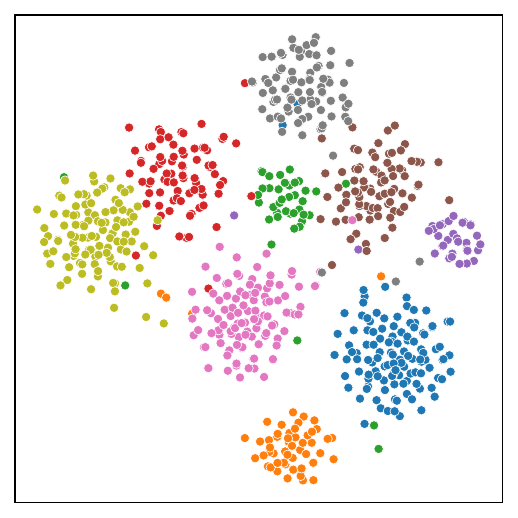}\\
\includegraphics[width=0.1\columnwidth]{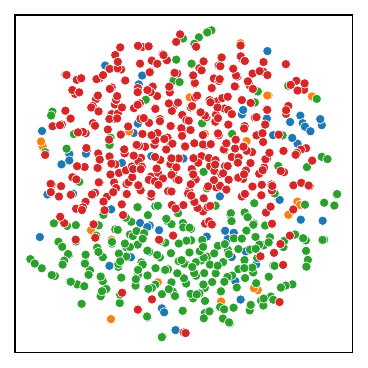}
\includegraphics[width=0.1\columnwidth]{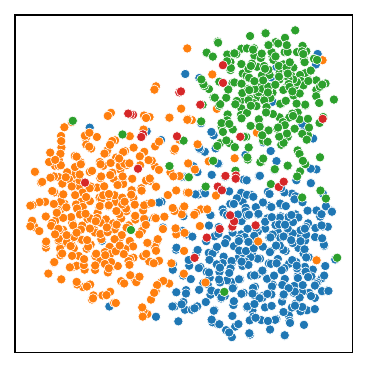}
\includegraphics[width=0.1\columnwidth]{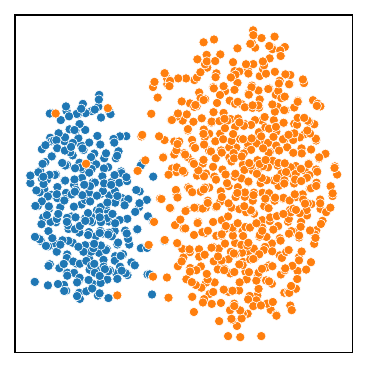}
\includegraphics[width=0.1\columnwidth]{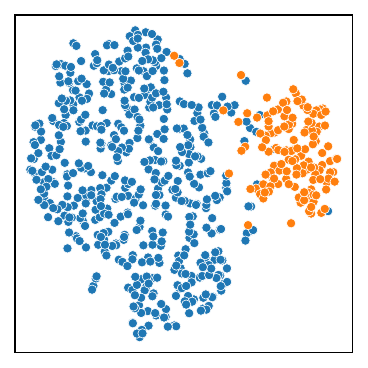}
\includegraphics[width=0.1\columnwidth]{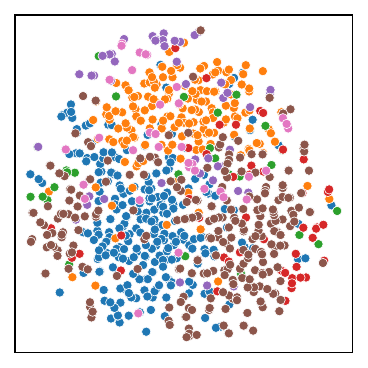}
\includegraphics[width=0.1\columnwidth]{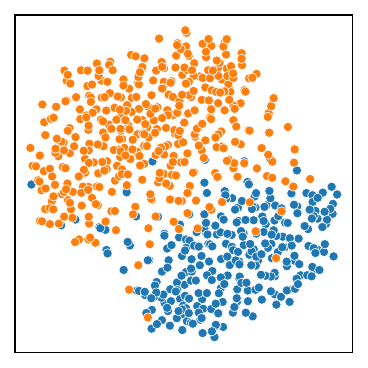}
\includegraphics[width=0.1\columnwidth]{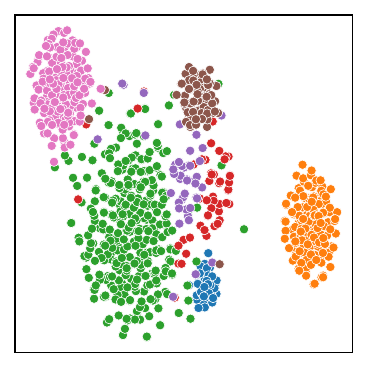}
\includegraphics[width=0.1\columnwidth]{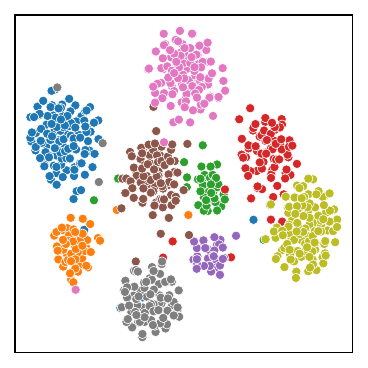}\\
\includegraphics[width=0.1\columnwidth]{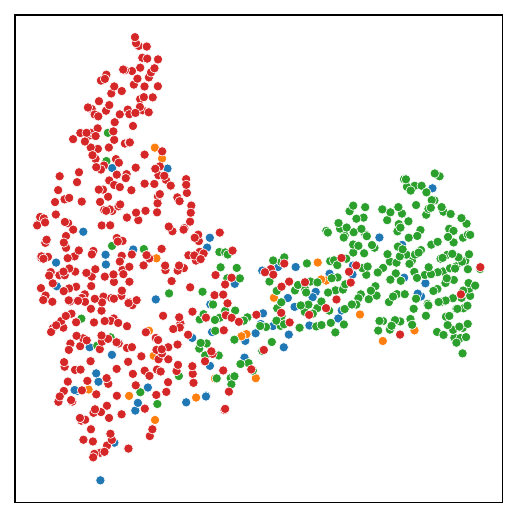}
\includegraphics[width=0.1\columnwidth]{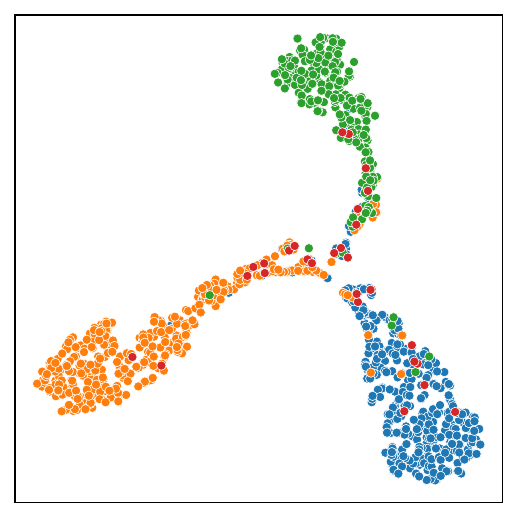}
\includegraphics[width=0.1\columnwidth]{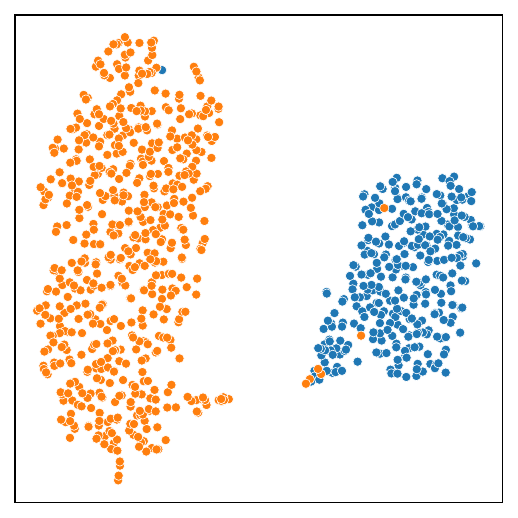}
\includegraphics[width=0.1\columnwidth]{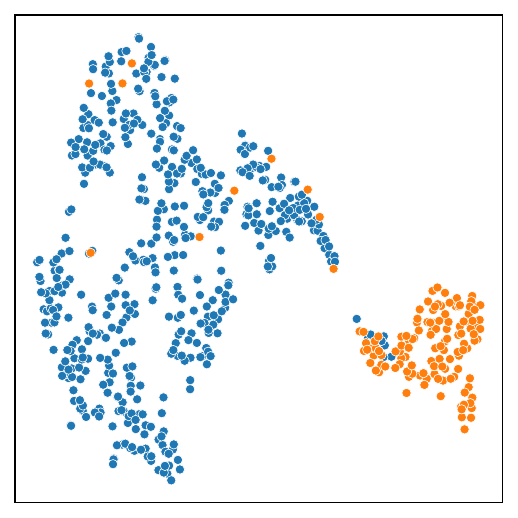}
\includegraphics[width=0.1\columnwidth]{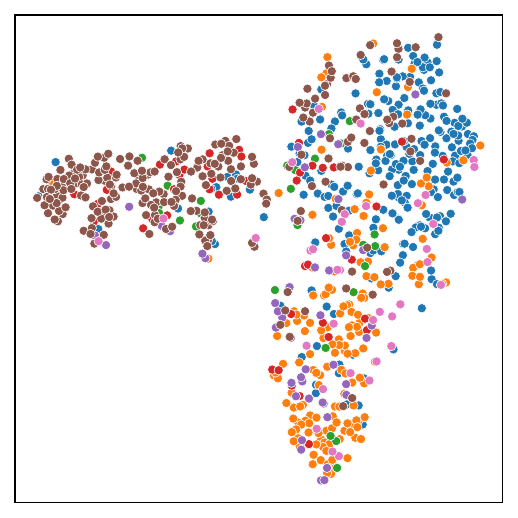}
\includegraphics[width=0.1\columnwidth]{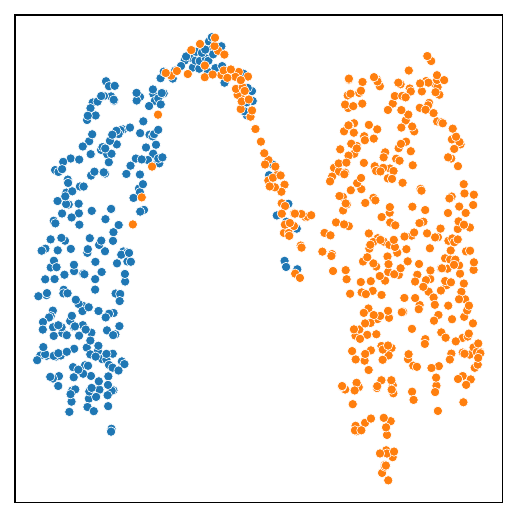}
\includegraphics[width=0.1\columnwidth]{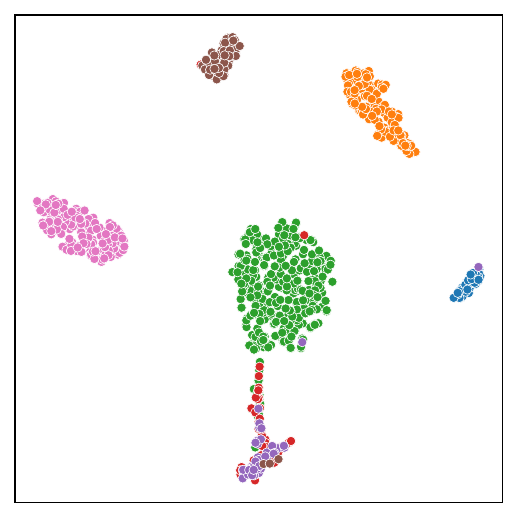}
\includegraphics[width=0.1\columnwidth]{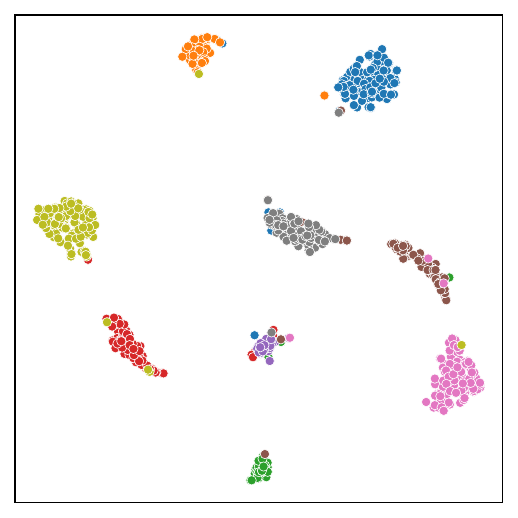}\\
\includegraphics[width=0.1\columnwidth]{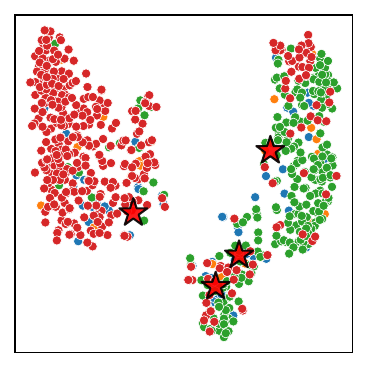}
\includegraphics[width=0.1\columnwidth]{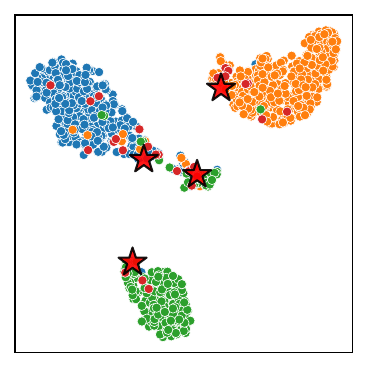}
\includegraphics[width=0.1\columnwidth]{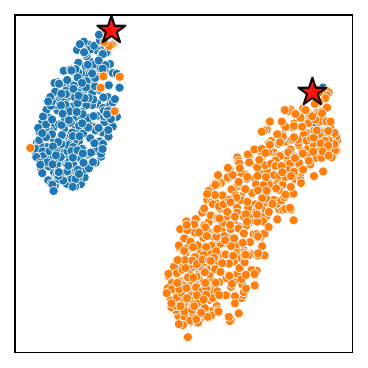}
\includegraphics[width=0.1\columnwidth]{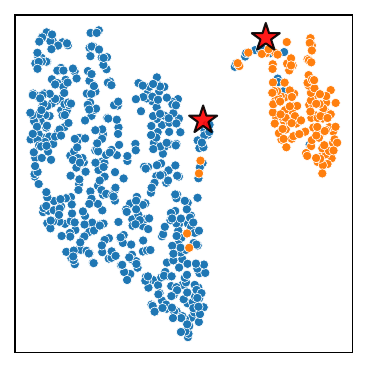}
\includegraphics[width=0.1\columnwidth]{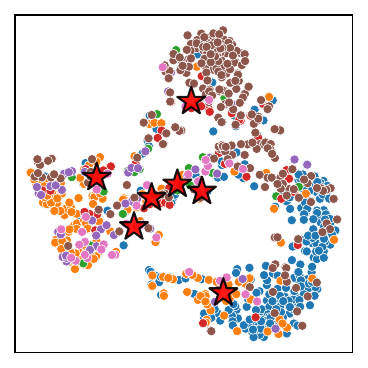}
\includegraphics[width=0.1\columnwidth]{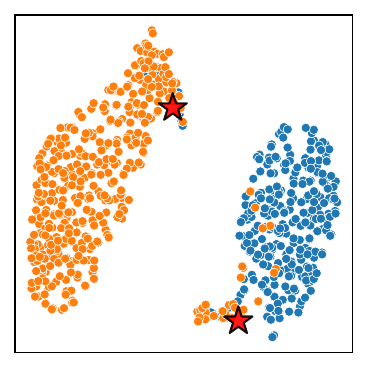}
\includegraphics[width=0.1\columnwidth]{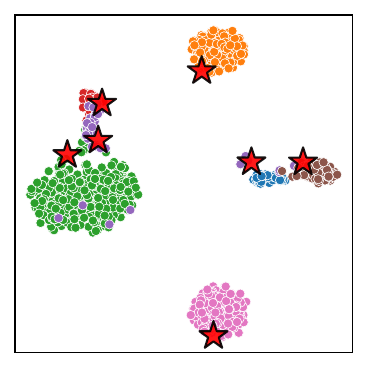}
\includegraphics[width=0.1\columnwidth]{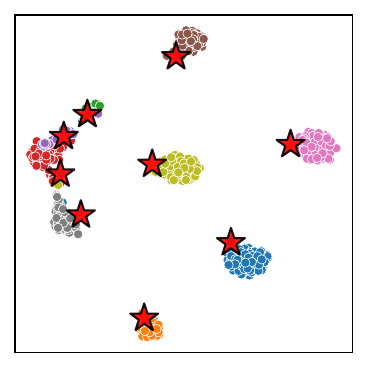}
\caption{\textbf{Representation $t$-SNE visualizations on ZEUS prior}. Rows are organized in groups of four; within each group, columns correspond to different datasets. From top to bottom, rows display raw features, TabICL encoder outputs, ZEUS representations, and TabClustPFN representations.}
\label{fig:representation_visualization_zeus_all}
\end{figure}

In the main paper (Fig.~\ref{fig:representation_visualization}), we presented representative examples to illustrate how different models shape the representation space. 
Here, we extend that analysis by providing $t$-SNE visualizations for \emph{all} synthetic datasets used in our evaluation. 
Figures~\ref{fig:representation_visualization_gmm_all} and~\ref{fig:representation_visualization_zeus_all} show results for datasets generated from GMM priors and the ZEUS prior, respectively.
The same protocol is followed as in the main text: we compare raw input features, pretrained TabICL embeddings, ZEUS representations, and \ours{} representations after the final projection and normalization. 
Points are colored by ground-truth labels, and for \ours{}, red stars denote learned cluster prototypes. 
This comprehensive view confirms that the geometric patterns highlighted in the main paper—such as the tendency of \ours{} to form more compact and prototype-aligned clusters, and the anisotropic or elongated structures often observed in ZEUS representations—are consistent across datasets rather than isolated cases.
It also becomes evident that ZEUS tends to perform better when the raw feature space already exhibits blob-like cluster structure, and comparatively worse when such structure is absent.

\subsection{Comparison with Cluster-PFN}
\label{app:cluster-pfn}

While Cluster-PFN~\citep{bhaskaran2025transformer} shares the objective of utilizing in-context learning to predict both cluster cardinality and assignments, its scope is restricted to a narrower low-dimensional GMM regime (2 to 5 feature dimensions). Consequently, to ensure a rigorous and fair evaluation, we compare the two approaches both methodologically and empirically under this specific controlled setting.

\textbf{Methodological comparison.}
Table~\ref{tab:cluster-pfn-comparison} summarizes the main differences between Cluster-PFN and \ours{}. 
Although both methods follow the PFN paradigm, they differ substantially in the prior, the assignment objective, the assignment architecture, and the mechanism for cardinality inference. 
Cluster-PFN is primarily designed around finite GMM priors and resolves label-switching through deterministic relabeling before applying a supervised cross-entropy loss. 
In contrast, \ours{} is pretrained on a broader prior family combining controlled-overlap GMMs and the ZEUS prior, and optimizes a permutation-invariant SoftARI objective. 
Architecturally, \ours{} introduces explicit cluster prototypes and iterative prototype--data cross-attention in the Partition Inference Network (PIN), while the Cardinality Inference Network (CIN) estimates $p(K\mid X)$ from assignment Gram matrices produced across candidate values of $K$. 
These differences make the cluster structure more explicit and provide a different route for scaling amortized clustering beyond very low-dimensional GMMs.

\begin{table}[!ht]
\centering
\caption{
Methodological comparison between Cluster-PFN and \ours{}.
Both methods perform amortized clustering, but differ in their prior, objective, architecture, and cardinality-inference mechanism.
}
\label{tab:cluster-pfn-comparison}
\begingroup
\small
\setlength{\tabcolsep}{4pt}
\renewcommand{\arraystretch}{1.18}
\begin{tabular}{
>{\raggedright\arraybackslash}p{0.22\textwidth}
>{\raggedright\arraybackslash}p{0.34\textwidth}
>{\raggedright\arraybackslash}p{0.36\textwidth}
}
\toprule
\textbf{Aspect} & \textbf{Cluster-PFN} & \textbf{\ours{}} \\
\midrule
Prior 
& Finite GMM prior in a low-dimensional setting ($D \leq 5$). 
& Controlled-overlap GMM prior combined with the ZEUS prior, covering both Gaussian and nonlinear non-Gaussian cluster geometries ($D \leq 64$). \\
\midrule
Cluster-assignment objective 
& Uses deterministic relabeling to break label permutation symmetry, followed by cross-entropy. 
& Optimizes a permutation-invariant SoftARI loss, avoiding deterministic relabeling conventions that can be sensitive to scaling and dimensionality. \\
\midrule
Cluster-assignment architecture 
& Uses standard Transformer layers followed by an MLP output head. 
& Uses a TabICL encoder followed by an iterative prototype--data cross-attention decoder in PIN. \\
\midrule
Cluster-assignment mechanism 
& Outputs membership probabilities directly from point embeddings. 
& Maintains $K$ cluster prototypes and computes assignments via cosine similarities between point representations and prototypes. \\
\midrule
Cardinality-inference architecture 
& Constructs a special object $\rho$ from a fixed all-$-1$ vector, concatenates it with dataset embeddings, and maps the decoded representation of $\rho$ to $p(K\mid X)$. 
& Uses CIN to map assignment Gram matrices $\{P_K^\top P_K\}_{K=2}^{K_{\max}}$ to a posterior over $K$. \\
\midrule
Conditioning on $K$ 
& Conditions the model through an embedding of the candidate value of $K$. 
& Conditions PIN by selecting the first $K$ cluster prototypes, so the active prototypes directly define the candidate cardinality. \\
\bottomrule
\end{tabular}
\endgroup
\end{table}

\textbf{Controlled GMM evaluation.}
To provide a fair empirical comparison within the regime where Cluster-PFN is designed to operate, we follow its low-dimensional GMM setting and evaluate all methods on newly generated held-out datasets from a GMM prior. 
For each feature dimension $D \in \{2,3,4,5\}$, we generate 320 independent datasets, resulting in 1,280 datasets in total. 
Each dataset contains $N=500$ observations and has an unknown number of clusters with maximum cluster count $K_{\max}=10$. 
All methods are evaluated in the unknown-$K$ setting, where the true cluster number is not provided at test time. 
In addition to Cluster-PFN and \ours{}, we include GMM$^+$ as a classical baseline, where the number of mixture components is selected by maximizing the silhouette score. 
This baseline is particularly relevant because the test data are themselves generated from a GMM prior, making the evaluation favorable to GMM-based methods.

\begin{figure}[ht]
    \centering
    \includegraphics[width=\textwidth]{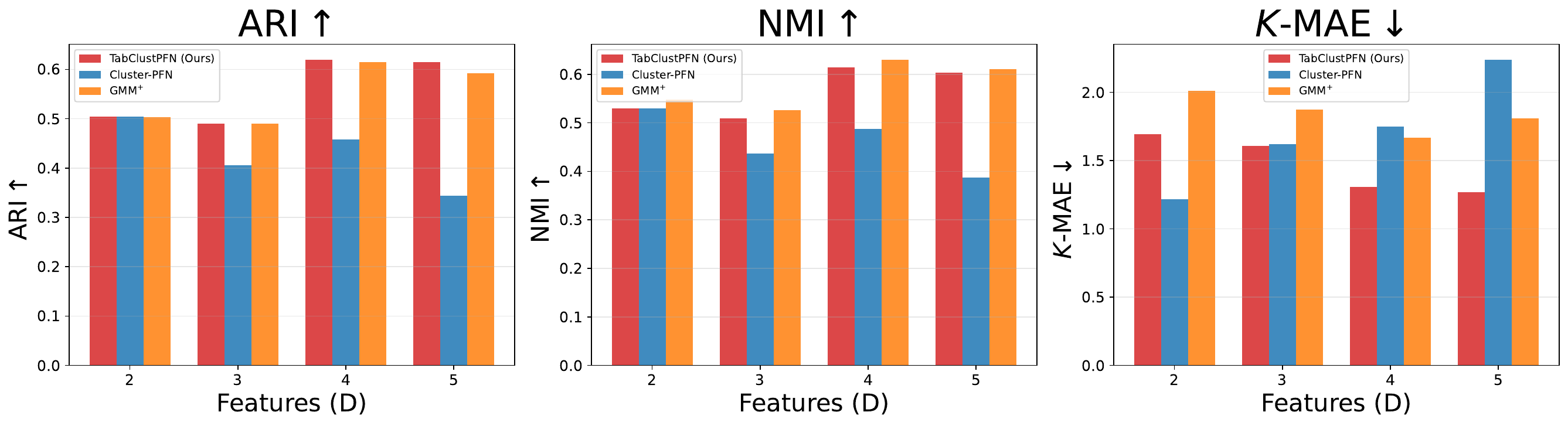}
    \caption{
    Comparison with Cluster-PFN under the low-dimensional GMM regime. 
    All methods are evaluated in the unknown-$K$ setting. 
    Higher ARI/NMI and lower $k$-MAE indicate better performance. 
    \ours{} consistently outperforms Cluster-PFN for $D>2$.
    }
    \label{fig:cluster-pfn-comparison}
\end{figure}

Figure~\ref{fig:cluster-pfn-comparison} reports ARI, NMI, and $k$-MAE across feature dimensions. 
At $D=2$, the three methods perform comparably, which is expected since this is the simplest setting and is well aligned with the assumptions of both Cluster-PFN and GMM$^+$. 
As the feature dimension increases, however, the performance of Cluster-PFN decreases in both clustering quality and cardinality estimation. 
In contrast, \ours{} remains stable and outperforms Cluster-PFN for all dimensions $D>2$ across ARI, NMI, and $k$-MAE. 
The gap is most significant for $D=4$ and $D=5$, where \ours{} achieves substantially higher ARI and NMI while also producing more accurate estimates of the number of clusters.

\subsection{Comparison with adaptive-cardinality baselines}
\label{app:adaptive_k_baselines}

CIN infers a posterior over a fixed set of candidate cardinalities, with $K_{\max}=10$ in the current checkpoint. To examine this design trade-off, we compare with DP-means~\citep{kulis2012revisiting} and Dip-means~\citep{kalogeratos2012dipmeans}, which can adapt the number of clusters without a fixed upper bound. This comparison evaluates clustering quality within the benchmark regime rather than extrapolation to $K>10$.

\paragraph{Experimental setup.}
We reuse the 49 synthetic and 44 real-world datasets and sampled observations from Sec.~\ref{sec:experiment} (seed 42). Both baselines follow the same preprocessing protocol as the classical baselines (App.~\ref{app:data_preprocessing}). TabClustPFN uses the unchanged checkpoint under unknown $K$. We compare package defaults with variants ($^{\dagger}$) selected using silhouette scores without ground-truth labels:

\begin{itemize}[leftmargin=*]
    \item \textbf{DP-means} (\texttt{pdc-dp-means} v0.0.8). Defaults use $\delta=1$ and eight initial clusters. DP-means$^{\dagger}$ searches $\delta$ over distinct positive quantiles $\{0.05,0.10,\ldots,0.95\}$ of pairwise squared Euclidean distances, using one initial cluster, five initializations, and at most 100 iterations.
    \item \textbf{Dip-means} (\texttt{ClustPy} v0.0.2). Defaults start from one cluster with significance level $0.001$. Dip-means$^{\dagger}$ searches significance levels $\{0.001,0.005,0.01,0.05\}$.
\end{itemize}

\begin{table}[ht]
\centering
\caption{\textbf{Comparison with adaptive-cardinality baselines under unknown $K$.} ARI and NMI are reported as median [IQR], and $k$-MAE as the median absolute cardinality error. Unmarked baselines use package defaults; $^{\dagger}$ denotes silhouette-selected hyperparameters.}
\label{tab:adaptive_k_baselines}
\setlength{\tabcolsep}{4pt}
\footnotesize
\resizebox{\linewidth}{!}{%
\begin{tabular}{lcccccc}
\toprule
Method & \multicolumn{3}{c}{Synthetic ($49$ datasets)} & \multicolumn{3}{c}{Real-world ($44$ datasets)} \\
\cmidrule(lr){2-4}\cmidrule(lr){5-7}
 & ARI $\uparrow$ & NMI $\uparrow$ & $k$-MAE $\downarrow$ & ARI $\uparrow$ & NMI $\uparrow$ & $k$-MAE $\downarrow$ \\
\midrule
TabClustPFN & \textbf{0.8308 [0.39]} & \textbf{0.8000 [0.42]} & \textbf{0} & \textbf{0.3853 [0.42]} & \textbf{0.4237 [0.36]} & \textbf{1} \\
\midrule
DP-means & 0.0552 [0.08] & 0.3484 [0.29] & 302 & 0.0351 [0.10] & 0.2767 [0.25] & 138 \\
DP-means$^{\dagger}$ & 0.2892 [0.78] & 0.4340 [0.67] & 3 & 0.2774 [0.41] & 0.3424 [0.36] & 2 \\
Dip-means & 0.0000 [0.00] & 0.0000 [0.00] & 3 & 0.0000 [0.24] & 0.0000 [0.44] & \textbf{1} \\
Dip-means$^{\dagger}$ & 0.0000 [0.00] & 0.0000 [0.00] & 4 & 0.0558 [0.39] & 0.1245 [0.48] & \textbf{1} \\
\bottomrule
\end{tabular}%
}
\end{table}

\paragraph{Results and design trade-off.}
Table~\ref{tab:adaptive_k_baselines} shows that TabClustPFN achieves higher median ARI and NMI than both baselines under either configuration on both benchmarks. It also has the lowest median $k$-MAE, tied with Dip-means on real-world data. Thus, allowing cluster counts beyond CIN's support does not by itself yield better clustering within the evaluated regime. CIN additionally provides a posterior over $K$, while the PIN--CIN decomposition supports inference conditioned on a user-specified $K$. The trade-off is that the current checkpoint cannot infer $K>K_{\max}$; extending this support remains future work.

\subsection{Statistical significance tests}
\label{app:significance}

We further assess whether the observed performance gains are statistically reliable across datasets. 
For each baseline method $b$ and each evaluation dataset $s$, we compute the paired ARI difference
\[
    \Delta_b(s) 
    = \mathrm{ARI}_{\ours{}}(s) - \mathrm{ARI}_{b}(s).
\]
Thus, $\Delta_b(s)>0$ indicates that \ours{} achieves a higher ARI than baseline $b$ on dataset $s$. 
All significance tests are conducted on these per-dataset paired differences, separately for the synthetic benchmark and the real-world benchmark.

We use two complementary paired tests. 
First, we apply the one-sided paired $t$-test with alternative hypothesis
\[
    H_1: \mathbb{E}[\Delta_b] > 0,
\]
which tests whether \ours{} has a larger mean ARI than baseline $b$ across datasets. 
The paired $t$-test is commonly used for paired algorithm comparisons, but it relies on a normality assumption for the paired differences. 
Since clustering scores are bounded, e.g., ARI lies in $[-1,1]$ and NMI lies in $[0,1]$, their empirical distributions need not be normal. 
We therefore also report the one-sided Wilcoxon signed-rank test, a non-parametric paired test that does not require normality of the paired differences and tests whether the distribution of $\Delta_b$ is shifted toward positive values. 
This follows the recommendation of Dem\v{s}ar~\citep{demsar2006statistical} to use non-parametric tests when comparing learning algorithms over multiple datasets.

In addition to pairwise tests, we report critical-difference (CD) diagrams based on average ARI ranks. 
For each dataset, all methods are ranked according to ARI, with lower rank indicating better performance. 
The CD diagram visualizes the average rank of each method and the critical difference required for two average ranks to be considered statistically distinguishable under the corresponding rank-based post-hoc comparison. 
We use CD diagrams as a complementary multi-method, rank-based view of the results, while the paired tests above directly compare \ours{} with each baseline in terms of ARI differences.

\begin{figure}[ht]
    \centering
    \includegraphics[width=\textwidth]{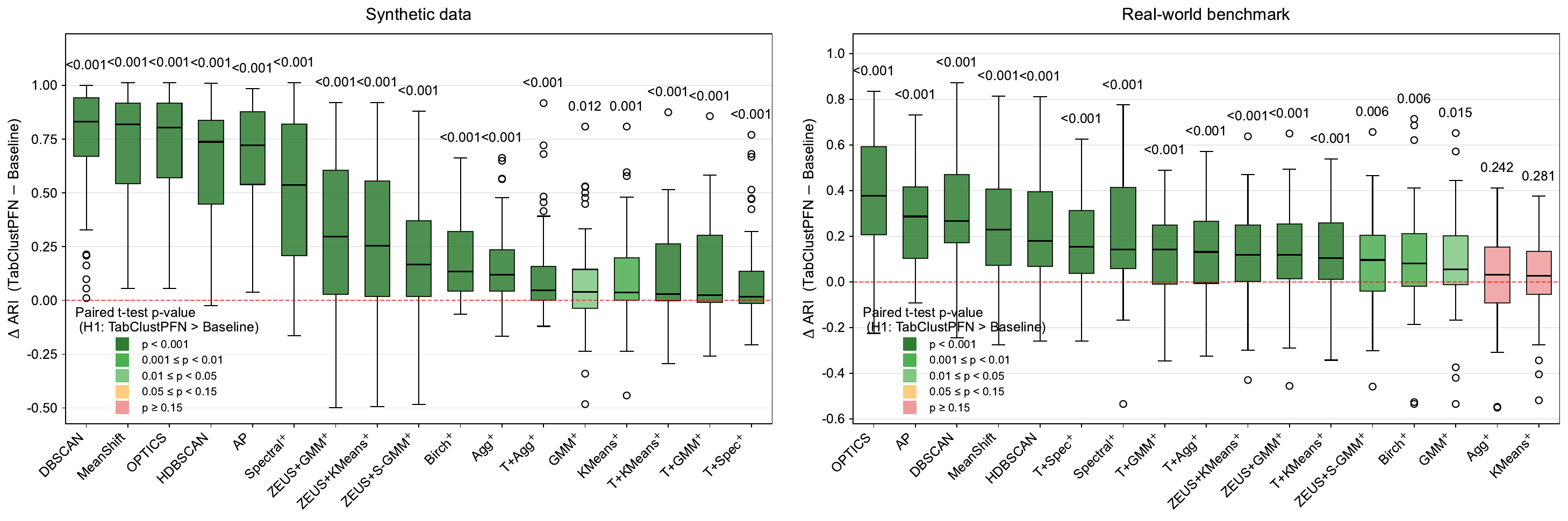}
    \caption{
    Paired $t$-test results based on per-dataset ARI differences $\Delta_b = \mathrm{ARI}_{\ours{}}-\mathrm{ARI}_{b}$.
    The dashed red line denotes zero difference.
    Positive values indicate that \ours{} outperforms the corresponding baseline.
    Colors indicate the one-sided paired $t$-test $p$-value under the alternative hypothesis $H_1:\mathrm{ARI}_{\ours{}}>\mathrm{ARI}_{b}$.
    }
    \label{fig:ttest-significance}
\end{figure}

\textbf{Paired $t$-tests.}
Figure~\ref{fig:ttest-significance} shows the distributions of $\Delta_b(s)$ and the one-sided paired $t$-test $p$-values. 
On the synthetic benchmark, \ours{} achieves positive paired differences against all baselines, and all paired $t$-tests reject the null hypothesis at the $0.05$ level. 
Most comparisons are highly significant with $p<0.001$. 
On the real-world benchmark, \ours{} also achieves positive median ARI differences against the vast majority of baselines. 
The paired $t$-tests show significant improvements over most baselines; the only non-significant comparisons are against Agg$^+$ and KMeans$^+$, which are the strongest classical competitors in this benchmark. 
Importantly, non-significance in these two comparisons does not indicate that the baselines outperform \ours{}; rather, the observed paired differences are not large enough to reject the null hypothesis under the paired $t$-test. 
\ours{} still achieves higher ARI on the majority of real-world datasets.

\begin{figure}[ht]
    \centering
    \includegraphics[width=\textwidth]{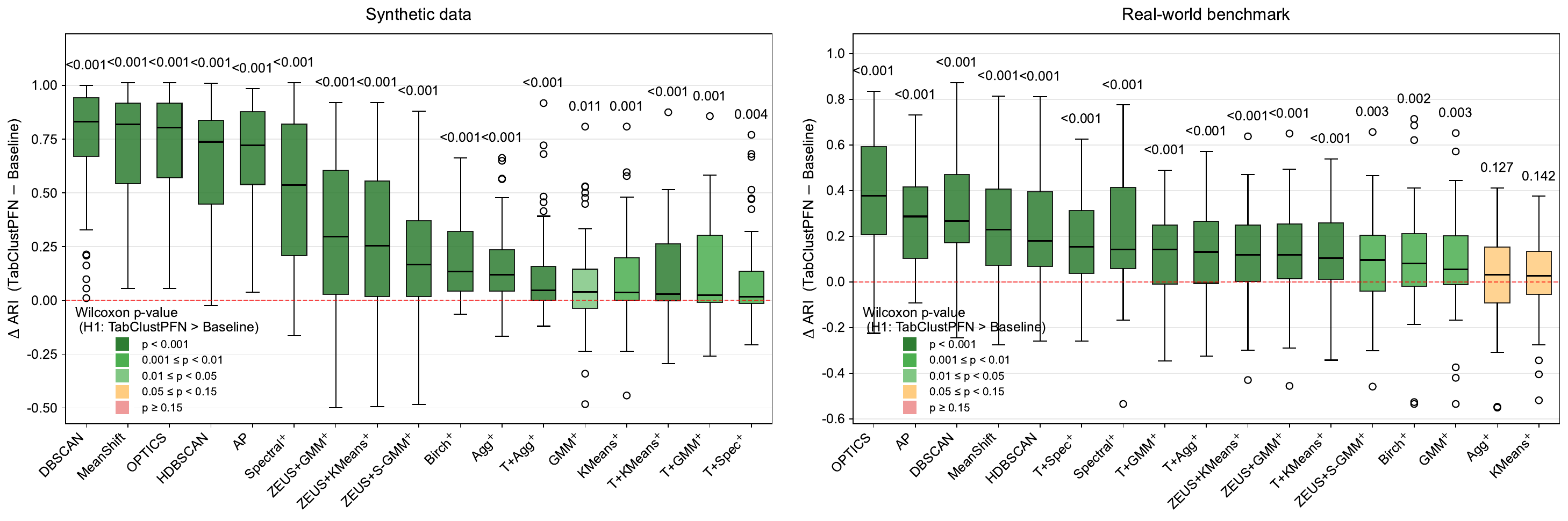}
    \caption{
    Wilcoxon signed-rank test results based on per-dataset ARI differences $\Delta_b = \mathrm{ARI}_{\ours{}}-\mathrm{ARI}_{b}$.
    Positive values indicate that \ours{} outperforms the corresponding baseline.
    Colors indicate the one-sided Wilcoxon signed-rank test $p$-value under the alternative hypothesis $H_1:\mathrm{ARI}_{\ours{}}>\mathrm{ARI}_{b}$.
    }
    \label{fig:wilcoxon-significance}
\end{figure}

\textbf{Wilcoxon signed-rank tests.}
Figure~\ref{fig:wilcoxon-significance} reports the corresponding one-sided Wilcoxon signed-rank tests. 
The results are consistent with the paired $t$-tests while avoiding the normality assumption on the paired differences. 
On the synthetic benchmark, \ours{} significantly outperforms all baselines. 
On the real-world benchmark, the Wilcoxon test yields the same qualitative conclusion and supports significant improvements over most baselines.

\begin{figure}[ht]
    \centering
    \begin{minipage}{0.49\textwidth}
        \centering
        \includegraphics[width=\textwidth]{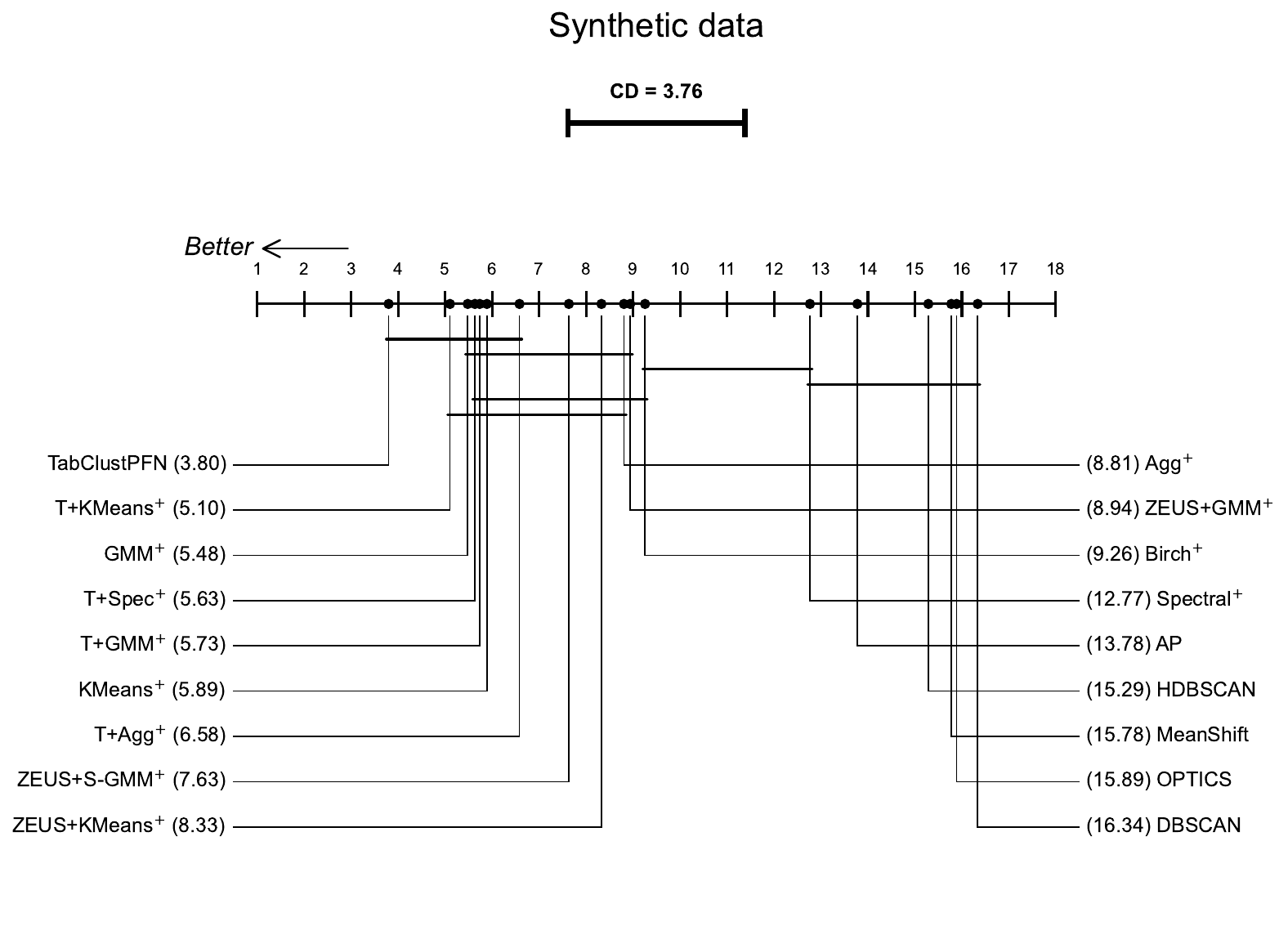}
    \end{minipage}
    \hfill
    \begin{minipage}{0.49\textwidth}
        \centering
        \includegraphics[width=\textwidth]{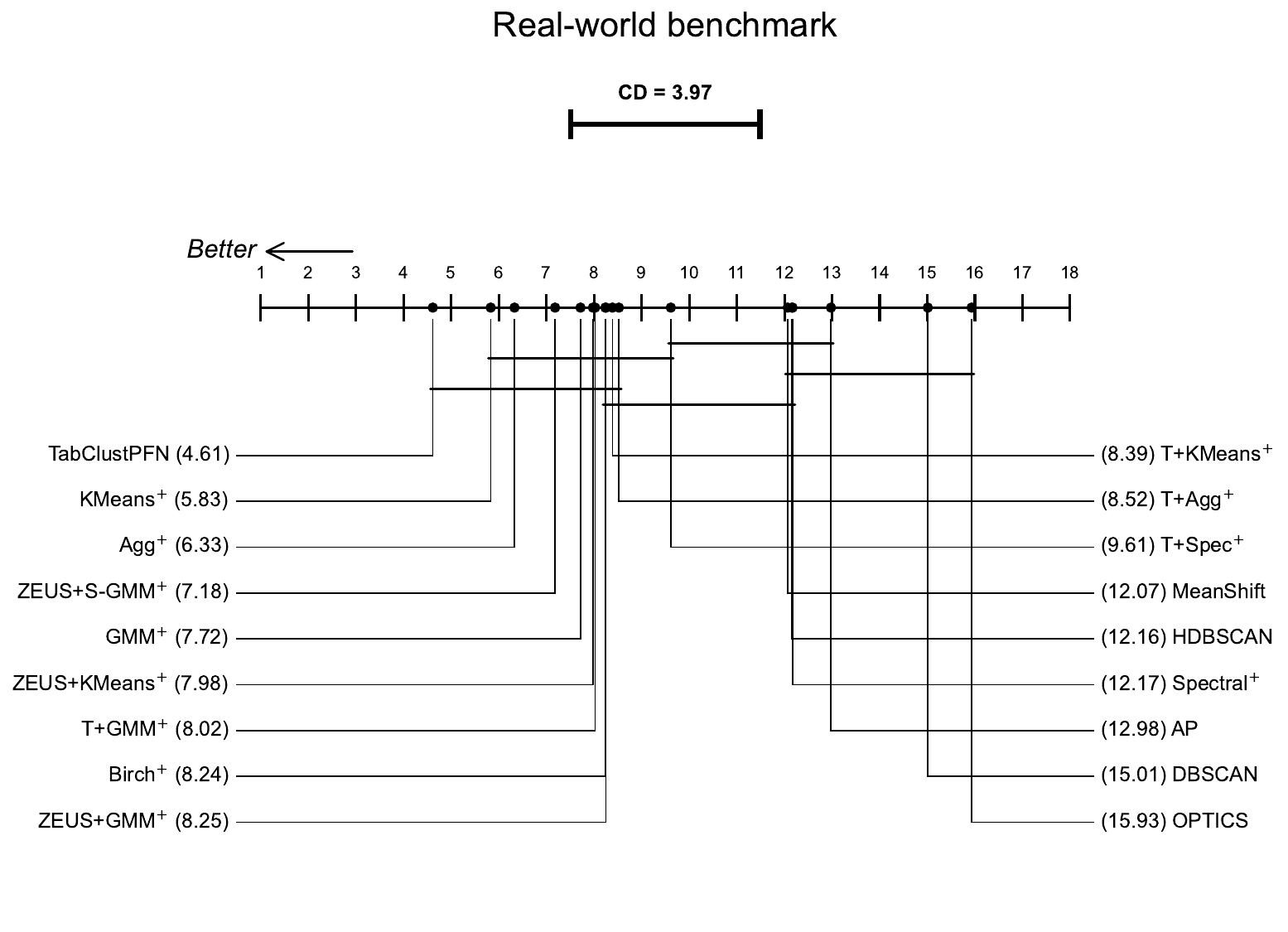}
    \end{minipage}
    \caption{
    Critical-difference (CD) diagrams based on ARI ranks.
    Lower average rank indicates better performance.
    }
    \label{fig:cd-significance}
\end{figure}

\textbf{Critical-difference diagrams.}
Figure~\ref{fig:cd-significance} provides a complementary rank-based comparison using critical-difference (CD) diagrams. 
On both the synthetic and real-world benchmarks, \ours{} achieves the best average ARI rank, with average ranks of $3.80$ and $4.61$, respectively. 
This ranking-based view is consistent with the paired tests above and further supports the overall advantage of \ours{} across benchmarks.

\subsection{Calibration of cardinality inference}
\label{app:calibration}

We evaluate the calibration of the cardinality posterior $\hat p(K\mid X)$ predicted by CIN. 
Since $\hat p(K\mid X)$ is a categorical distribution over candidate cluster cardinalities, we assess calibration through set-valued coverage. 
Given a nominal level $\alpha$, we sort candidate cardinalities in descending order of $\hat p(K\mid X)$ and construct the smallest prediction set whose cumulative posterior mass reaches $\alpha$:
\[
    \widehat{\mathcal{K}}_{\alpha}(X)
    =
    \{K_{(1)},\ldots,K_{(m_\alpha)}\},
    \qquad
    m_\alpha
    =
    \min\left\{
    m:\sum_{\ell=1}^{m}\hat p(K_{(\ell)}\mid X)\ge \alpha
    \right\},
\]
where $\hat p(K_{(1)}\mid X)\ge \hat p(K_{(2)}\mid X)\ge \cdots$. 
We then compute the empirical coverage
\[
    \frac{1}{S}\sum_{s=1}^{S}
    \mathbf{1}\!\left\{
    K_s^\ast \in \widehat{\mathcal{K}}_{\alpha}(X_s)
    \right\},
\]
where $K_s^\ast$ is the ground-truth cluster cardinality for dataset $s$. 
A well-calibrated posterior should have empirical coverage close to the nominal level, while systematic undercoverage would indicate overconfidence.

Cardinality calibration is considered instead of point-level assignment calibration, since the latter is not directly well-defined in clustering. 
Since cluster labels are identifiable only up to permutation, assessing assignment calibration would require a post-hoc matching between predicted clusters and ground-truth labels, making the notion of a ``correct'' label dependent on a dataset-level permutation. 
By contrast, the cluster cardinality $K$ is label-free and comparable across datasets.

We therefore evaluate set-valued calibration of $\hat p(K\mid X)$ on 3,200 held-out datasets generated from the GMM prior, covering feature dimensions from $2$ to $64$, sample sizes from $500$ to $1000$, and maximum cluster count $K_{\max}=10$.



\subsection{Pairwise win-rate comparison on real-world benchmark}
\label{app:pairwise_win_rate}
\begin{figure}[!htb]
\centering
\includegraphics[width=0.8\linewidth]{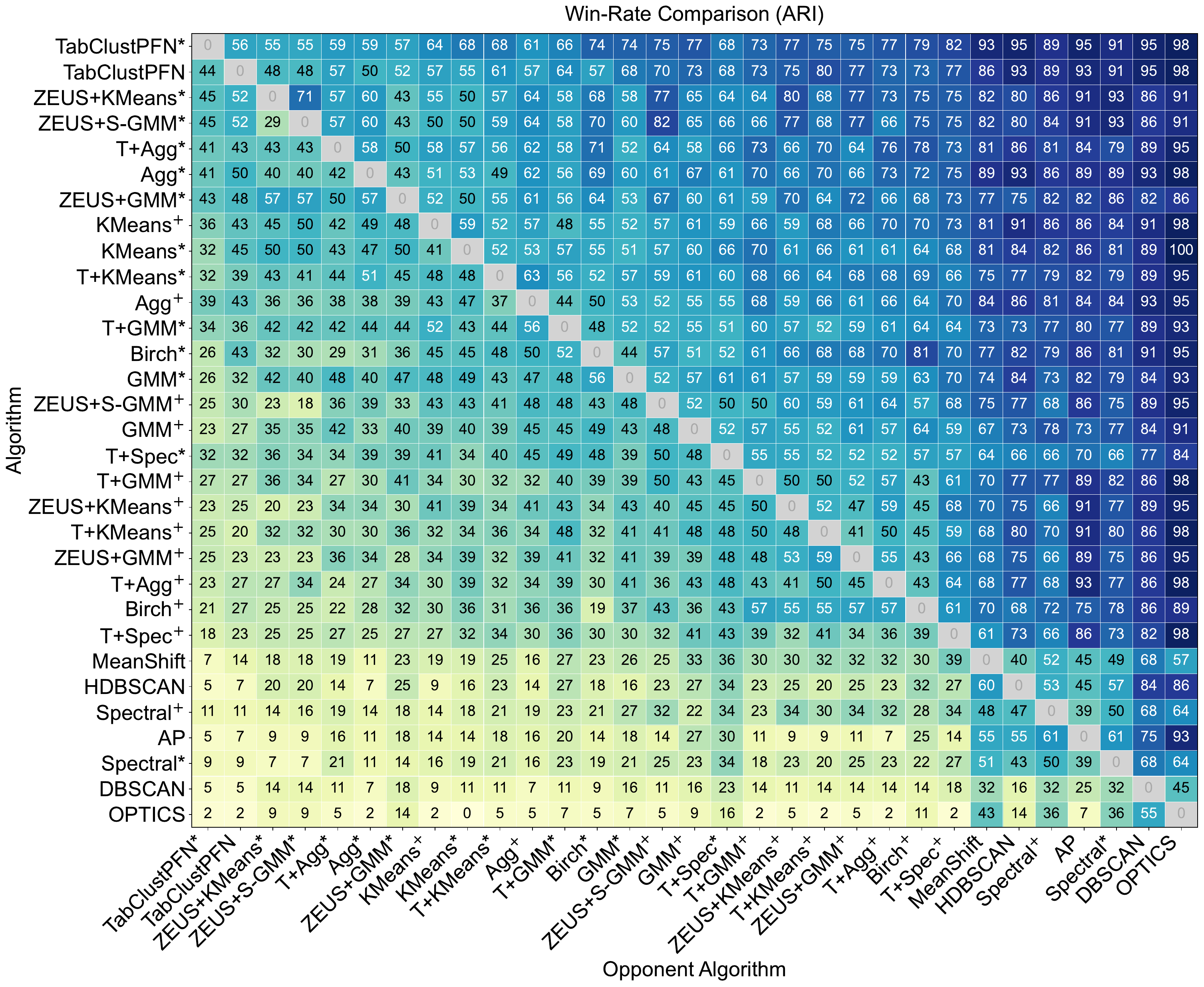}
\caption{\textbf{Pairwise win rate comparison on real-world benchmark}. Each entry shows the percentage of datasets on which the row method achieves a higher ARI than the column method.}
\label{fig:win_rate}
\end{figure}

To compare methods beyond average metrics, we visualize pairwise win rates across datasets in Fig.~\ref{fig:win_rate}.
The heatmap entry at row $i$ and column $j$ denotes the percentage of datasets on which method $i$ achieves a higher ARI than method $j$. 
Values greater than $50$ therefore indicate that method $i$ outperforms method $j$ on the majority of datasets, while values below $50$ indicate the opposite. 
Diagonal entries are undefined and omitted.

The figure compares all $31$ methods, including those that assume known $K$ and those that infer $K$ from the data. 
As such, it provides a global view of relative robustness across datasets rather than isolating specific evaluation settings.

TabClustPFN$^*$ (with known $K$) exhibits win counts above $50$ against all other methods, indicating consistent superiority across the majority of datasets. 
In contrast, TabClustPFN (with inferred $K$) does not outperform TabClustPFN$^*$, ZEUS+Kmeans$^*$, ZEUS+GMM$^*$, and Agg$^*$, but achieves higher win counts than the remaining baselines. 
This behavior is expected, as these four methods are provided with the true number of clusters, giving them additional information unavailable in the unknown-$K$ setting.

\subsection{Results grouped by information on the number of clusters}
\label{app:rank_group_by_knowledge_on_k}
\begin{figure}[!htbp]
\centering
\includegraphics[width=0.35\linewidth]{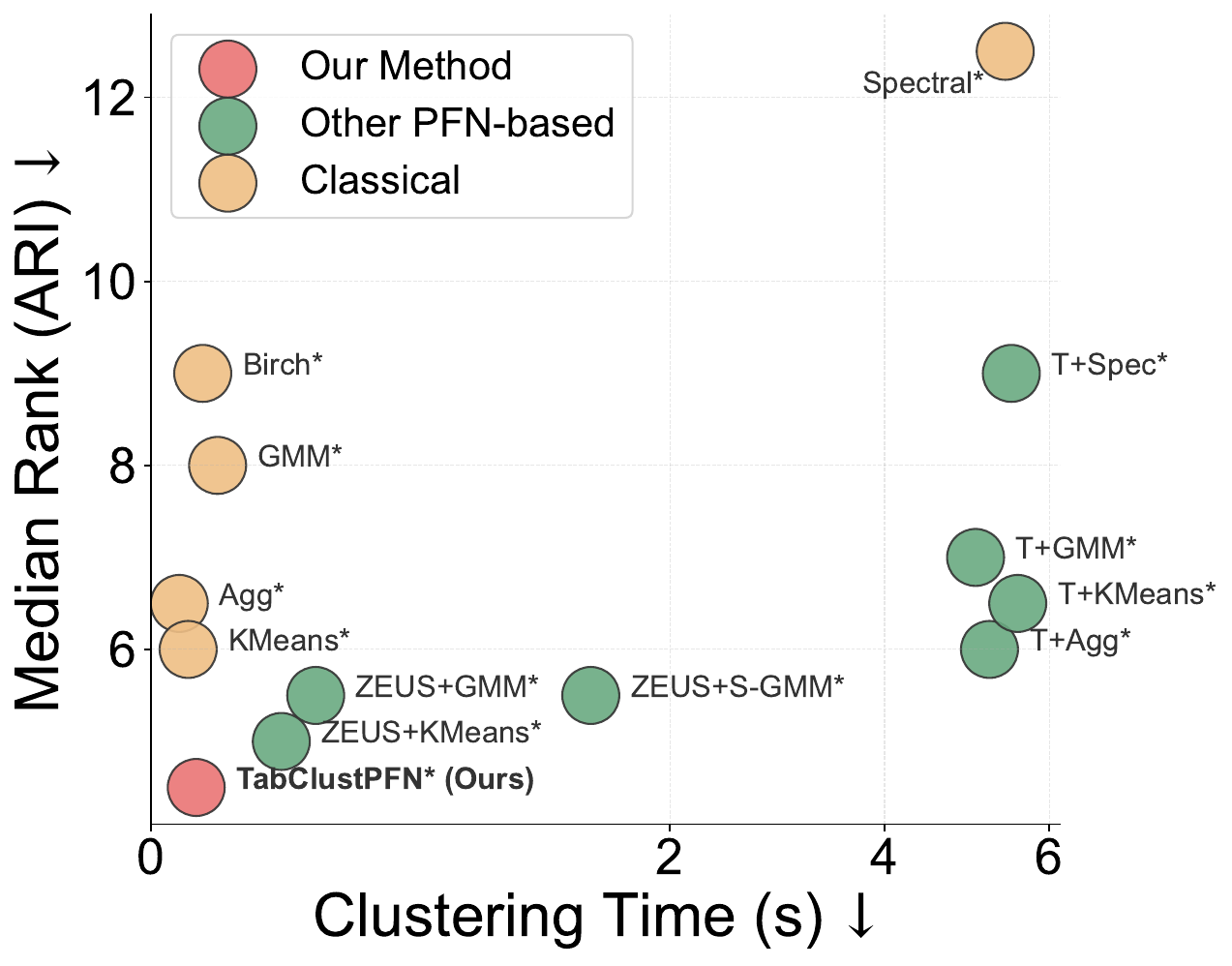}
\includegraphics[width=0.35\linewidth]{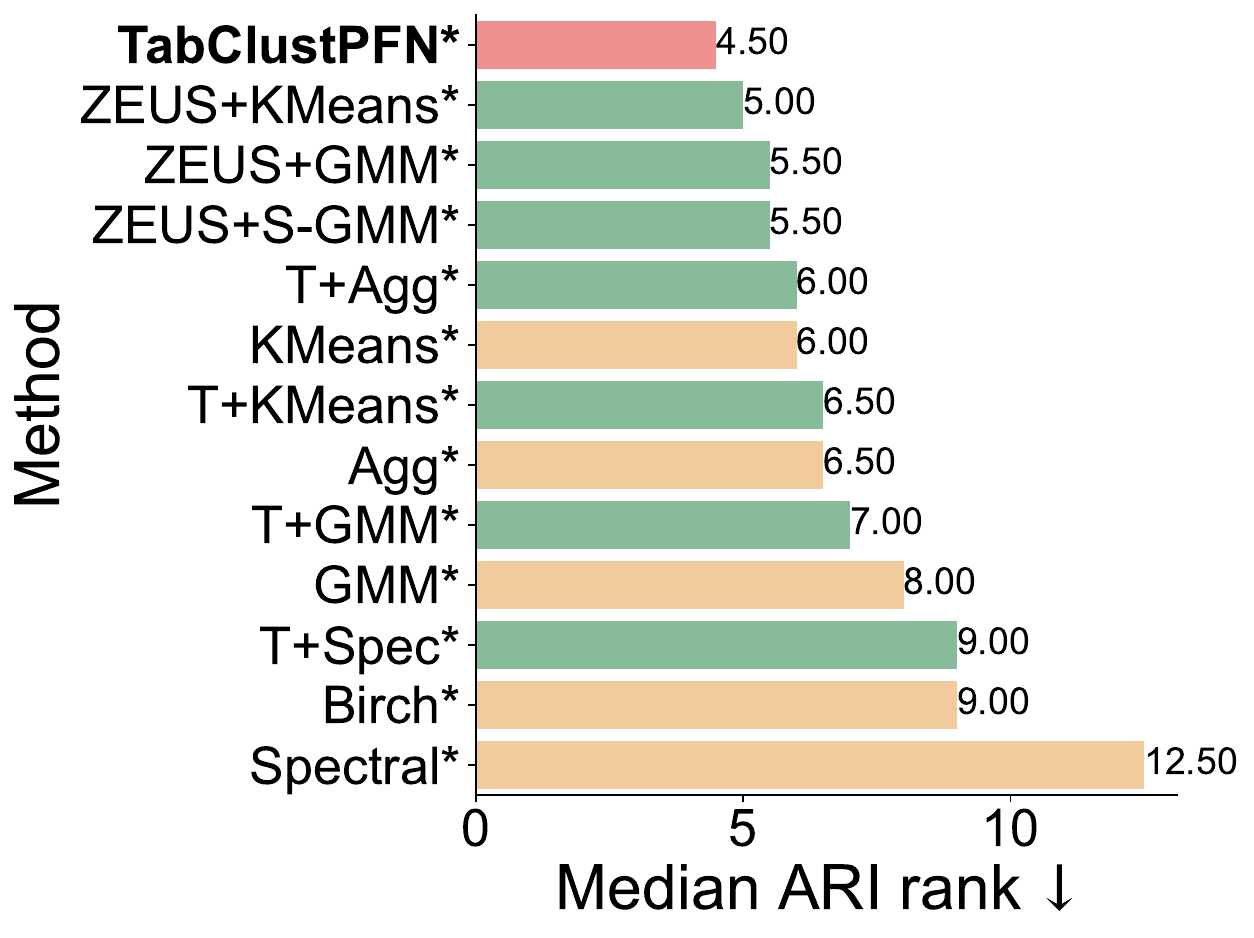}
\includegraphics[width=0.35\linewidth]{figures/real_auto_k_time_vs_rank.pdf}
\includegraphics[width=0.35\linewidth]{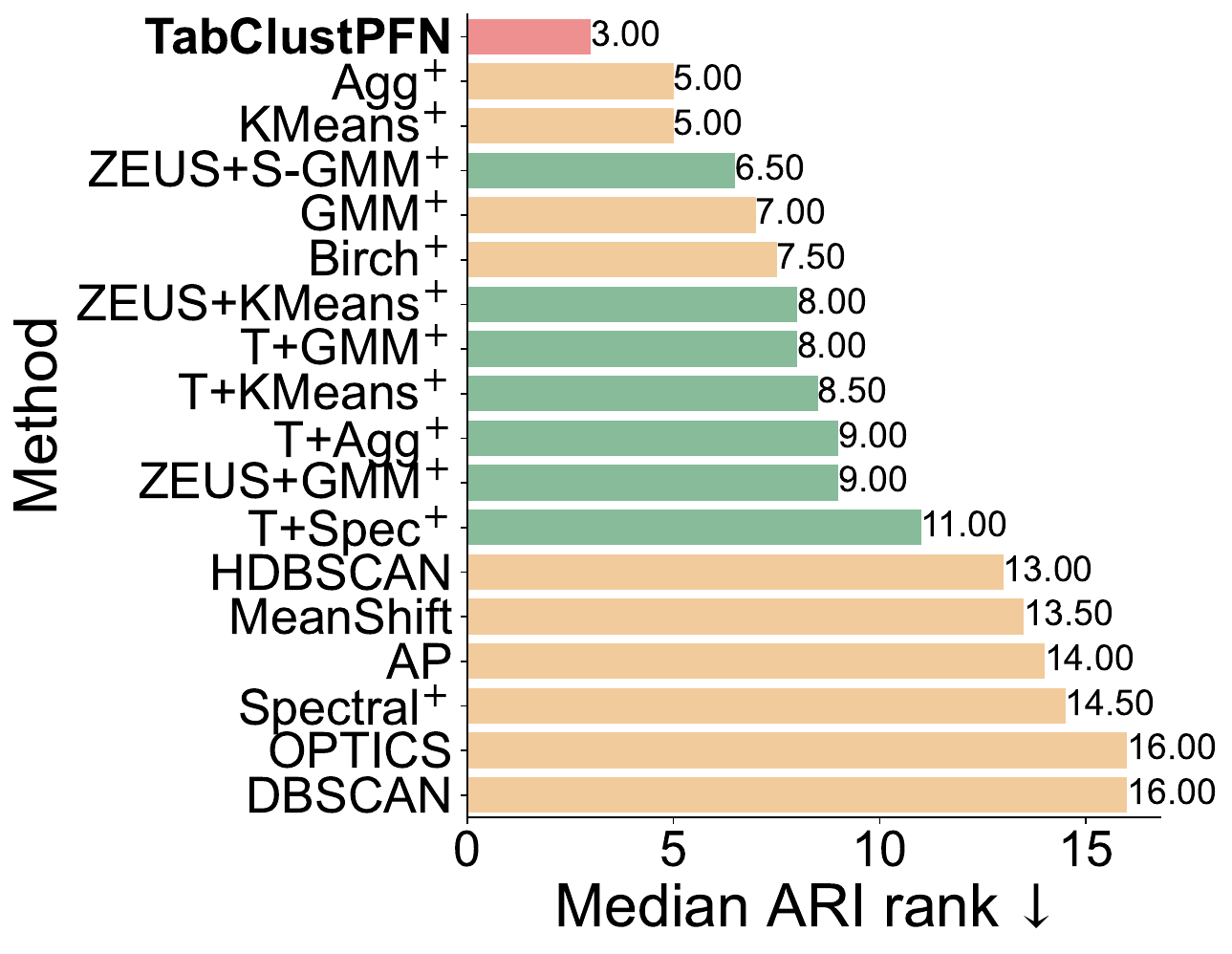}
\caption{\textbf{Runtime and median ARI rank grouped by knowledge of $K$ on real-world benchmark}. Top row: methods with known $K$. Bottom row: methods with unknown $K$.}
\label{fig:finer_grained_ari_time}
\end{figure}

In the main paper, we report the median ARI rank of all methods on the real-world benchmark. 
To provide a more controlled comparison, we further group methods according to whether the number of clusters $K$ is given or must be inferred. 
Fig.~\ref{fig:finer_grained_ari_time} reports median ARI ranks for methods with known $K$ (top row) and methods with unknown $K$ (bottom row).
This stratified evaluation isolates performance differences under equal information conditions. 
In both settings, \ours{} achieves the best (lowest) median ARI rank among the compared methods, indicating consistently strong clustering performance given the same level of prior knowledge about $K$.

We additionally plot inference time against median ARI rank. 
These scatter plots illustrate the joint relationship between computational cost and clustering quality. 
\ours{} lies in the favorable region of this space, achieving the best clustering performance (lowest median ARI rank) while requiring substantially less inference time than computationally intensive baselines such as Spectral clustering, and remaining more efficient than other PFN-based approaches.

\input{sections/appendix/ood}

%% file: sections/appendix/ood.tex
\subsection{Out-of-distribution generalization}
\label{app:ood_generalization}

\subsubsection{Synthetic distribution shifts}
\label{app:ood_synthetic_shifts}

\paragraph{Experimental setup.}
We use the main benchmark's unknown-$K$ evaluation protocol and preprocessing (Sec.~\ref{sec:experiment}; App.~\ref{app:data_preprocessing}). Both experiments use $N=1000$, $K=3$, and $D\in\{2,10,30,50\}$. Component centers are drawn uniformly from $[-5,5]^D$.

\begin{enumerate}[leftmargin=*,label=(\roman*),nosep]
\item \textbf{Cluster-size imbalance.} We sample isotropic Gaussian components with covariance $s_k I$, $s_k\sim\mathcal{U}(0.3,0.8)$, and target sizes proportional to $(r,1,1)$ for $r\in\{1,2,5,10,20,50,100\}$. Integer sizes are adjusted to sum to $1000$.
\item \textbf{Heavy-tailed subpopulations.} We sample approximately equal-sized Student-$t$ components with degrees of freedom $\nu\in\{1,2,3,5,10,30,1000\}$. Samples follow $x=\mu_k+\sigma_k z/\sqrt{v/\nu}$, where $z\sim\mathcal{N}(0,I_D)$, $v\sim\chi^2_\nu$, and $\sigma_k\sim\mathcal{U}(0.5,1.0)$. Smaller $\nu$ gives heavier tails; $\nu=1000$ approximates Gaussian components.
\end{enumerate}

\begin{figure}[htbp]
\centering
\includegraphics[width=\linewidth]{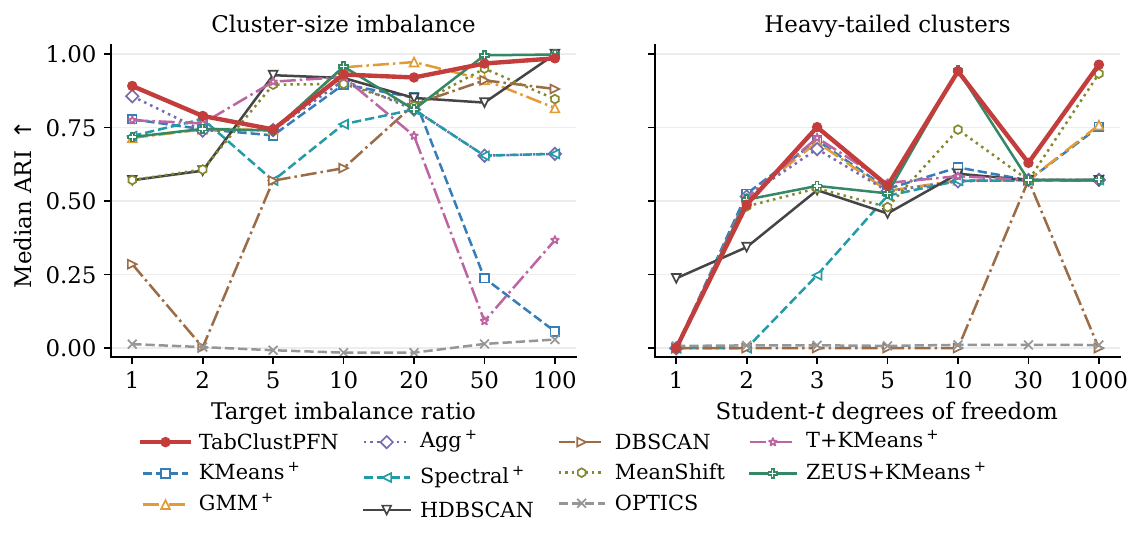}
\caption{\textbf{Synthetic distribution shifts at $D=2$.} Left: cluster-size imbalance. Right: heavy-tailed subpopulations. Each point reports median ARI ten data-generation seeds ($0$--$9$) for one condition under unknown $K$.}
\label{fig:ood_distribution_shifts}
\end{figure}

\begin{figure}[htbp]
\centering
\includegraphics[width=\linewidth]{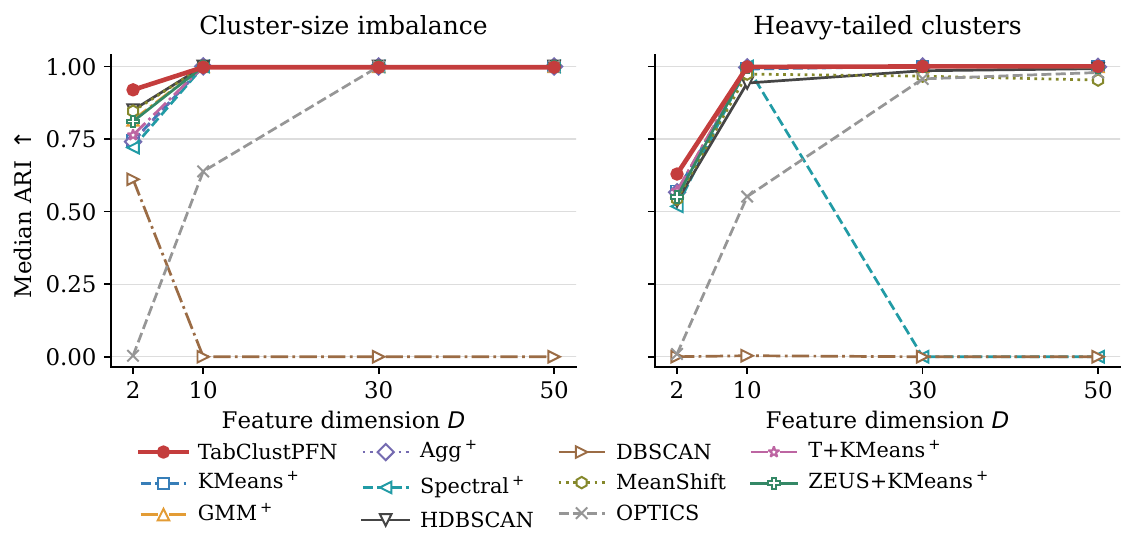}
\caption{\textbf{Synthetic distribution shifts across feature dimensions.} For each method and dimension, we first compute median ARI ten data-generation seeds ($0$--$9$) within each condition, then take the median across the seven imbalance ratios (left) or seven degrees of freedom (right).}
\label{fig:ood_distribution_shifts_by_dimension}
\end{figure}

\paragraph{Results.}
Figure~\ref{fig:ood_distribution_shifts} compares methods across shift conditions at $D=2$, where heavy-tailed clusters with small $\nu$ are particularly challenging. Figure~\ref{fig:ood_distribution_shifts_by_dimension} summarizes the trend across dimensions: TabClustPFN's median ARI approaches one for both shift families at higher dimensions.

\subsubsection{Real-world OOD evaluation}
\label{app:ood_real_world}
\paragraph{Datasets and evaluation.}
Table~\ref{tab:ood_real_datasets} lists the 6 additional datasets selected from OpenML-CC18, TabArena, and the AutoML benchmark. Dataset selection and preprocessing follow Apps.~\ref{sec:dataset_stats} and~\ref{app:data_preprocessing}, within the OOD range specified in the main text. We compare with KMeans$^+$, Agg$^+$, HDBSCAN, and ZEUS+KMeans$^+$ to cover centroid-based, hierarchical, density-based, and PFN-based clustering, respectively. The reported TabClustPFN results use direct inference without dimensionality reduction.

\begin{table}[htbp]
\centering
\caption{\textbf{The 6 real-world OOD datasets.} OpenML IDs identify the exact datasets. $N$ and $D$ are the original sample and feature counts.}
\label{tab:ood_real_datasets}
\setlength{\tabcolsep}{6pt}
\footnotesize
\begin{tabular}{rlrrr}
\toprule
OpenML ID & Dataset & $N$ & $D$ & $K$ \\
\midrule
458 & analcatdata\_authorship & 841 & 70 & 4 \\
12 & mfeat-factors & 2000 & 216 & 10 \\
46980 & MIC & 1699 & 111 & 8 \\
14 & mfeat-fourier & 2000 & 76 & 10 \\
41143 & jasmine & 2984 & 144 & 2 \\
40966 & MiceProtein & 1080 & 77 & 8 \\
\bottomrule
\end{tabular}%

\end{table}

\paragraph{Results.}
Table~\ref{tab:ood_real_world} reports absolute metrics for the selected collection. TabClustPFN has the highest median ARI/NMI and lowest median cardinality error, with mean runtime close to KMeans$^+$. The broad IQRs indicate substantial variation across datasets.

\begin{table}[htbp]
\centering
\caption{\textbf{Real-world OOD results on the 6 datasets in Table~\ref{tab:ood_real_datasets}.} Median [IQR] ARI/NMI, median $k$-MAE, and mean clustering time. Bold marks the best result in each column.}
\label{tab:ood_real_world}
\setlength{\tabcolsep}{5pt}
\footnotesize
\begin{tabular}{lcccc}
\toprule
Method & ARI $\uparrow$ & NMI $\uparrow$ & $k$-MAE $\downarrow$ & Time (s) $\downarrow$ \\
\midrule
\ours{} & \textbf{0.200 [0.371]} & \textbf{0.352 [0.462]} & \textbf{1} & 0.124 \\
ZEUS+KMeans$^+$ & 0.165 [0.441] & 0.327 [0.487] & 5 & 1.363 \\
KMeans$^+$ & 0.183 [0.469] & 0.229 [0.523] & 4 & \textbf{0.106} \\
Agg$^+$ & 0.155 [0.470] & 0.289 [0.545] & 4 & 1.792 \\
HDBSCAN & 0.083 [0.064] & 0.141 [0.319] & 7 & 1.016 \\
\bottomrule
\end{tabular}%

\end{table}

\FloatBarrier

%% file: sections/appendix/ablation.tex
\section{Ablation study details}
\label{app:ablation_details}
This section provides additional ablation results.
Details of the baselines used in the \pin{} loss ablation are given in Sec.~\ref{app:pin_ablation_loss}, followed by the \pin{} decoder ablation (Sec.~\ref{app:pin_decoder_ablation}) and further analysis of the \cin{} loss (Sec.~\ref{app:cin_loss_ablation}).

\subsection{Different losses in \pin{} loss ablation}
\label{app:pin_ablation_loss}
SoftARI is the default objective for \pin{} (Eq.~\eqref{eq:pin_loss}). To evaluate its effectiveness, we keep the PIN architecture, training protocol, and priors fixed, and replace only the training objective with several alternatives.
Throughout, for a dataset with $N$ observations and ground-truth cluster labels $\{z_i\}_{i=1}^N$ ($z_i \in \{1,\dots,K\}$), PIN produces logits $\ell_i \in \mathbb{R}^K$ and soft assignments $P_{ik}=\mathrm{softmax}(\ell_i)_k$. 
Let $Z \in \{0,1\}^{N\times K}$ be the one-hot encoding of $\{z_i\}$, i.e., $Z_{ij}=\mathbbm{1}[z_i=j]$. 
We define the soft confusion matrix
\begin{equation*}
    M = P^{\top} Z \in \mathbb{R}^{K \times K}, \quad M_{kj} = \sum_{i=1}^{N} P_{ik} Z_{ij}.
\end{equation*}
Intuitively, $M_{kj}$ is the total probability mass assigned to predicted cluster $k$ by samples whose true label is $j$. 
Using $P$, instead of hard assignments from $\arg \max$, keeps $M$ differentiable w.r.t. the logits, enabling gradient-based optimization.

A central challenge in clustering is label switching: relabeling predicted clusters corresponds to permuting the columns of $P$, which in turn permutes the rows of the soft confusion matrix $M$. An objective is permutation-invariant if it remains unchanged under such reindexing of $M$ (e.g., by depending only on row/column-order-agnostic statistics) or if it explicitly optimizes over the best reindexing (e.g., via matching or a differentiable relaxation). We highlight this property for each ablated objective below.

\subsubsection{SoftNMI loss}
We compute a differentiable variant of normalized mutual information (SoftNMI) between the predicted partition and the ground-truth labels using the soft confusion matrix $M$. We first form the empirical joint distribution and marginals
\[
\hat{p}_{kj}=\frac{M_{kj}}{N}, \qquad 
\hat{p}_k=\sum_{j=1}^K \hat{p}_{kj}, \qquad 
\hat{p}_j=\sum_{k=1}^K \hat{p}_{kj}.
\]
The corresponding entropies are
\[
H(\hat{p}_k)=-\sum_{k=1}^K \hat{p}_k \log \hat{p}_k, \qquad
H(\hat{p}_j)=-\sum_{j=1}^K \hat{p}_j \log \hat{p}_j,
\]
and the mutual information is
\[
I \;=\; \sum_{k=1}^K\sum_{j=1}^K \hat{p}_{kj}\log\frac{\hat{p}_{kj}}{\hat{p}_k\hat{p}_j}.
\]
We adopt the arithmetic-mean normalization
\[
\mathrm{SoftNMI} \;=\; \frac{2I}{H(\hat{p}_k)+H(\hat{p}_j)}.
\]
The resulting training objective is
\[
\mathcal{L}_{\mathrm{SoftNMI}} \;=\; 1 - \mathrm{SoftNMI}.
\]
This objective is permutation-invariant: permuting predicted cluster indices only reorders the index $k$ in the sums above, leaving the value unchanged.

\subsubsection{Matching+CE loss}
We consider a matching-based baseline that resolves label switching by explicitly selecting the best one-to-one reindexing between predicted clusters and ground-truth labels. Using the soft confusion matrix $M \in \mathbb{R}^{K \times K}$ as the matching score matrix, we compute a permutation matrix
\begin{equation*}
    \Pi^\star \in \arg \max_{\Pi \in \mathcal{P}_K} \langle \Pi, M \rangle_F,
\end{equation*}
where $\mathcal{P}_K = \{\Pi \in \{0, 1\}^{K \times K} : \Pi \mathbf{1} = \mathbf{1}, \Pi^{\top} \mathbf{1} = \mathbf{1} \}$ and $\langle A, B \rangle_F = \sum_{k,j} A_{kj} B_{kj}$.
Let $\pi^\star$ be the permutation induced by $\Pi^\star$ (i.e., $\Pi^\star_{k,j} = 1 \Leftrightarrow k = \pi^\star(j)$). 
We then remap each ground-truth label $Z_i$ to a target predicted-cluster index $\tilde{Z}_i = \pi^\star(Z_i)$, and apply standard cross-entropy on the logits:
\begin{equation*}
    \mathcal{L}_{\text{Match+CE}} = \frac{1}{N} \sum_{i=1}^{N} \text{CE}(\ell_i, \tilde{Z}_i).
\end{equation*}
Permutation invariance comes from the maximization over $\pi$: any reindexing of predicted clusters corresponds to permuting rows of $M$, but the optimal matching (and the resulting best-achievable agreement) is unchanged after re-solving the assignment. In practice, the matching step is discrete and is treated as a stop-gradient operation; gradients flow only through the cross-entropy term given the matched targets.

\subsubsection{Matching+SoftAcc loss}


To avoid the non-differentiability of discrete matching in $\mathcal{P}_K$, we replace $\Pi^\star$ with a differentiable relaxation $\Pi \in [0, 1]^{K \times K}$ obtained via Sinkhorn normalization, and define a soft matched agreement by $\langle \Pi, M \rangle_F$, which reduces to $\langle \Pi^\star, M \rangle_F$ when $\Pi$ is a hard permutation matrix.

We relax $\mathcal{P}_K$ to its convex hull, i.e., the Birkhoff polytope of doubly-stochastic matrices $\mathcal{B}_K = \{ \Pi \geq 0 : \Pi \mathbf{1} = \mathbf{1}, \Pi^{\top} \mathbf{1} = \mathbf{1} \}$. Intuitively, $\mathcal{B}_K$ contains ``soft permutations'': each row and column sums to one, so $\Pi_{kj}$ can be interpreted as a fractional one-to-one alignment weight between predicted cluster $k$ and true cluster $j$. We obtain such a $\Pi$ by applying Sinkhorn-Knopp normalization~\citep{Sinkhorn1967} to the log-scaled affinities:

\begin{equation*}
    \Pi = \text{Sinkhorn} \left( \frac{\log(M + \epsilon)}{\epsilon} \right) \in [0, 1]^{K \times K},
\end{equation*}

which iteratively normalizes rows and columns to produce an approximately doubly-stochastic matrix. The temperature $\epsilon$ controls how ``hard'' the alignment is: as $\epsilon$ decreases, $\Pi$ becomes increasingly peaked and approaches a permutation-like matrix, recovering a continuous analogue of hard matching while remaining differentiable.

Given $\Pi$, we define a soft matched accuracy as the expected aligned agreement under this soft one-to-one correspondence:

\begin{equation*}
    \text{SoftAcc} = \frac{1}{N} \langle \Pi, M \rangle_F = \frac{1}{N} \sum_{k=1}^K \sum_{j=1}^K \Pi_{kj} M_{kj}, \quad \mathcal{L}_{\text{Match+SoftAcc}} = 1 - \text{SoftAcc}.
\end{equation*}

This quantity is a natural relaxation of matched clustering accuracy: when $\Pi$ is a hard permutation matrix, $\langle \Pi, M \rangle_F$ reduces to the sum of the entries of $M$ on the permuted diagonal, i.e., the total correctly aligned mass (and dividing by $N$ yields an accuracy-like normalization). When $\Pi$ is soft, the same expression computes the agreement averaged over fractional matchings, hence a ``soft'' accuracy. Because both Sinkhorn normalization and the Frobenius inner product are differentiable w.r.t. $M$, gradients flow through $M = P^{\top} Z$ to the soft assignments $P$, enabling end-to-end training.

\subsection{\pin{} decoder ablation baselines}
\label{app:pin_decoder_ablation}


To isolate the contribution of the decoder, we keep the PIN encoder unchanged and replace only the \pin{} decoder with two non-iterative alternatives. All variants take the same per-point representations $R^{(0)} \in \mathbb{R}^{N \times D}$ from the \pin{} encoder as input and output per-point cluster-assignment logits. For a fair comparison, we align model capacity by using the same embedding dimension, number of heads, and feedforward width, and by matching the total number of attention layers in the decoder (counting both self-attention and cross-attention).


\subsubsection{Na\"ve (SA+MLP) decoder}
The Na\"ive decoder predicts cluster assignments directly from point embeddings, without maintaining explicit cluster representations.
We reuse the self-attention operator $\mathrm{SA}(\cdot)$ defined in Eq.~\eqref{eq:sa_def} and stack $2L$ SA blocks (without positional encodings) to refine the point features:
\begin{equation*}
    R^{(l+1)} = \mathrm{SA}(R^{(l)}), \quad l=0,\dots,2L-1,
    \qquad
    \tilde{R} = R^{(2L)} \in \mathbb{R}^{N \times d}.
\end{equation*}
We use $2L$ blocks to match the parameter budget of other ablations.
We then apply a point-wise two-layer MLP head (hidden size $2d$) to produce logits over $K_{\max}$ clusters for every point:
\begin{equation*}
    \ell = \mathrm{MLP}(\tilde{R}) \in \mathbb{R}^{N \times K_{\max}}.
\end{equation*}
For a dataset with $K$ clusters, we use the first $K$ coordinates $\ell^{(K)} = (\ell_{:,1}, ..., \ell_{:,K}) \in \mathbb{R}^{N \times K}$ and mask the remaining coordinates as invalid.
This baseline follows the SA+MLP intuition: stacked self-attention captures point--point interactions, while the MLP produces $K$-way assignment logits per point.
Since clusters are only implicitly represented in the output layer, this baseline does not maintain cluster prototypes.




\subsubsection{Non-iterative Transformer decoder}
This baseline introduces explicit cluster representations but performs decoding without iterative refinement.
We first compute set-aware point features by stacking $L$ self-attention blocks (Eq.~\eqref{eq:sa_def}) on the input set, without positional encodings:
\begin{equation*}
    R^{(l+1)} = \mathrm{SA}\!\left(R^{(l)}\right),\quad l=0,\dots,L-1,
    \qquad
    R^{(L)} \in \mathbb{R}^{N\times d}.
\end{equation*}

We then introduce $K_{\max}$ learnable cluster prototypes $C^{(0)}\in\mathbb{R}^{K_{\max}\times d}$ and update them for $L$ layers using (i) prototype self-attention (Eq.~\eqref{eq:sa_def}) and (ii) prototype-to-data cross-attention $\mathrm{CA}_{C\leftarrow R}(\cdot,\cdot)$ (Eq.~\eqref{eq:ca_def}), both without positional encodings:
\begin{equation*}
    \tilde{C}^{(\ell)} = \mathrm{SA}\!\left(C^{(\ell)}\right), \qquad
    C^{(\ell+1)} = \mathrm{CA}_{C\leftarrow R}\!\left(\tilde{C}^{(\ell)}, R^{(L)}\right),
    \quad \ell=0,\dots,L-1.
\end{equation*}

Given the final prototypes $C^{(L)}$ and encoded point features $R^{(L)}$, we compute data-to-prototype assignment logits using the same projection, normalization, and similarity computation as the main model, producing $\ell\in\mathbb{R}^{N\times K_{\max}}$.
For a dataset with $K$ clusters, we select the first $K$ coordinates to form $\ell^{(K)}\in\mathbb{R}^{N\times K}$ and mask the remaining coordinates as invalid.
The rest of the pipeline follows the main experiments.

\subsection{\cin{} ordinal loss ablation}
\label{app:cin_loss_ablation}


\cin{} predicts the number of clusters $K$, which naturally has an ordinal structure. Our default CIN loss, however, treats $K \in \{2, \dots, K_{\max}\}$ as a categorical label and optimizes a flat multi-class cross-entropy objective. To assess whether explicitly leveraging the ordering of $K$ is beneficial, we replace this objective with an \textit{order-aware ordinal logistic regression loss}, while keeping the PIN backbone, CIN fingerprint construction, and the overall training protocol unchanged.
Concretely, for each candidate $K \in \{2, \dots, K_{\max}\}$, we compute the same normalized Gram matrix $G^{(K)} = \frac{1}{N} (P^{(K)})^\top P^{(K)}$ and the same permutation-invariant fingerprint $g^{(K)} \in \mathbb{R}^{K(K+1)/2}$ as in App.~\ref{app:cin}, and concatenate $\{g^{(K)}\}_{K=2}^{K_{\max}}$ into a fixed-length feature vector. Following our decoupled optimization design, we detach $\ell^{(K)}$ and hence $P^{(K)}$, $G^{(K)}$, and $g^{(K)}$, when forming the input features, so the ordinal logistic regression loss only updates the head parameters and does not backpropagate into \pin{}.

\begin{table}[!htbp]
\centering
\caption{\textbf{\cin{} loss ablation}. Median $k$MAE on synthetic and real datasets using different \cin{} losses.}
\label{tab:cin_loss_ablation}
\begin{tabular}{lcc}
\toprule
CIN loss & Synthetic & Real \\
\midrule
Order-aware & 1.0 & 2.5        \\
CE (Ours)     & \textbf{0.0} & \textbf{1.0} \\
\bottomrule
\end{tabular}
\end{table}

The order-aware CIN head first maps the concatenated fingerprints through an MLP to produce a scalar score $\text{CIN}_{\text{Order-aware}}(X) \in \mathbb{R}$. We then introduce monotonic thresholds $\{b_K\}_{K=2}^{K_{\max}-1}$ parameterized by unconstrained variables $\{\delta_j\}_{j=1}^{K_{\max}-2}$ via
\[
b_K = \sum_{j=1}^{K-1} \text{softplus}(\delta_j), \quad K = 2, \dots, K_{\max}-1,
\]
which enforces $b_2 \le b_3 \le \dots \le b_{K_{\max}-1}$ automatically. With a positive scale $s$, the ordinal logits are defined as
\[
\eta_K(X) = s(\text{CIN}_{\text{Order-aware}}(X) - b_K), \quad K = 2, \dots, K_{\max}-1,
\]
where $\eta_K(X)$ is interpreted as the logit for the binary event $\Pr(K^\star > K \mid X)$. Given the ground-truth cardinality $K^\star$, we construct binary targets $\mathbb{I}[K^\star > K]$ and optimize the average binary cross-entropy across thresholds:
\[
\mathcal{L}_{\text{ord}} = \frac{1}{K_{\max} - 2} \sum_{K=2}^{K_{\max}-1} \text{BCEWithLogits}(\eta_K(X), \mathbb{I}[K^\star > K]).
\]

At inference time, we predict $\hat{K}$ by counting how many thresholds are exceeded:
\[
\hat{K} = 2 + \sum_{K=2}^{K_{\max}-1} \mathbb{I}[\eta_K(X) \ge 0],
\]
which yields $\hat{K} \in \{2, \dots, K_{\max}\}$ and respects the natural ordering of $K$.

Tab.~\ref{tab:cin_loss_ablation} reports the median $k$MAE of \ours{} pretrained with order-aware ordinal logistic regression loss.
Overall, incorporating ordinal supervision does not yield improvements over the default cross-entropy training of \cin{}.


%% file: sections/checklist.tex
\section*{NeurIPS Paper Checklist}

\begin{enumerate}

\item {\bf Claims}
    \item[] Question: Do the main claims made in the abstract and introduction accurately reflect the paper's contributions and scope?
    \item[] Answer: \answerYes{} 
    \item[] Justification: The abstract and Sec.~\ref{sec:intro} explicitly state the contributions—a hybrid GMM+ZEUS pretraining prior, a decoupled PIN–CIN architecture for joint inference of $\mZ$ and $K$ in a single forward pass, the permutation-invariant SoftARI objective, and SOTA results on a 44-dataset real-world benchmark—all supported by the experiments in Sec.~\ref{sec:experiment} and ablations in Sec.~\ref{sec:ablation}.
    \item[] Guidelines:
    \begin{itemize}
        \item The answer \answerNA{} means that the abstract and introduction do not include the claims made in the paper.
        \item The abstract and/or introduction should clearly state the claims made, including the contributions made in the paper and important assumptions and limitations. A \answerNo{} or \answerNA{} answer to this question will not be perceived well by the reviewers. 
        \item The claims made should match theoretical and experimental results, and reflect how much the results can be expected to generalize to other settings. 
        \item It is fine to include aspirational goals as motivation as long as it is clear that these goals are not attained by the paper. 
    \end{itemize}

\item {\bf Limitations}
    \item[] Question: Does the paper discuss the limitations of the work performed by the authors?
    \item[] Answer: \answerYes{}
    \item[] Justification: Limitations are discussed in Sec.~\ref{sec:experiment} and Sec.~\ref{sec:limits}, including the operating regime ($N\!\le\!1000$, $D\!\le\!64$, $K\!\le\!10$) used in pretraining and evaluation, and the open directions of extending to noisy and missing-value settings.
    \item[] Guidelines:
    \begin{itemize}
        \item The answer \answerNA{} means that the paper has no limitation while the answer \answerNo{} means that the paper has limitations, but those are not discussed in the paper. 
        \item The authors are encouraged to create a separate ``Limitations'' section in their paper.
        \item The paper should point out any strong assumptions and how robust the results are to violations of these assumptions (e.g., independence assumptions, noiseless settings, model well-specification, asymptotic approximations only holding locally). The authors should reflect on how these assumptions might be violated in practice and what the implications would be.
        \item The authors should reflect on the scope of the claims made, e.g., if the approach was only tested on a few datasets or with a few runs. In general, empirical results often depend on implicit assumptions, which should be articulated.
        \item The authors should reflect on the factors that influence the performance of the approach. For example, a facial recognition algorithm may perform poorly when image resolution is low or images are taken in low lighting. Or a speech-to-text system might not be used reliably to provide closed captions for online lectures because it fails to handle technical jargon.
        \item The authors should discuss the computational efficiency of the proposed algorithms and how they scale with dataset size.
        \item If applicable, the authors should discuss possible limitations of their approach to address problems of privacy and fairness.
        \item While the authors might fear that complete honesty about limitations might be used by reviewers as grounds for rejection, a worse outcome might be that reviewers discover limitations that aren't acknowledged in the paper. The authors should use their best judgment and recognize that individual actions in favor of transparency play an important role in developing norms that preserve the integrity of the community. Reviewers will be specifically instructed to not penalize honesty concerning limitations.
    \end{itemize}

\item {\bf Theory assumptions and proofs}
    \item[] Question: For each theoretical result, does the paper provide the full set of assumptions and a complete (and correct) proof?
    \item[] Answer: \answerNA{} 
    \item[] Justification: The paper does not present formal theoretical results or theorems; the mathematical formulation is used to define the model and training objectives.
    \item[] Guidelines:
    \begin{itemize}
        \item The answer \answerNA{} means that the paper does not include theoretical results. 
        \item All the theorems, formulas, and proofs in the paper should be numbered and cross-referenced.
        \item All assumptions should be clearly stated or referenced in the statement of any theorems.
        \item The proofs can either appear in the main paper or the supplemental material, but if they appear in the supplemental material, the authors are encouraged to provide a short proof sketch to provide intuition. 
        \item Inversely, any informal proof provided in the core of the paper should be complemented by formal proofs provided in appendix or supplemental material.
        \item Theorems and Lemmas that the proof relies upon should be properly referenced. 
    \end{itemize}

    \item {\bf Experimental result reproducibility}
    \item[] Question: Does the paper fully disclose all the information needed to reproduce the main experimental results of the paper to the extent that it affects the main claims and/or conclusions of the paper (regardless of whether the code and data are provided or not)?
    \item[] Answer: \answerYes{}
    \item[] Justification: The prior-data generation, network architecture, loss functions, optimizer and learning-rate schedule, and evaluation protocol are fully described in Secs.~\ref{sec:methods}--\ref{sec:experiment} and Apps.~\ref{app:training_details}--\ref{app:experimental_details}, and a public code repository is linked in the abstract.
    \item[] Guidelines:
    \begin{itemize}
        \item The answer \answerNA{} means that the paper does not include experiments.
        \item If the paper includes experiments, a \answerNo{} answer to this question will not be perceived well by the reviewers: Making the paper reproducible is important, regardless of whether the code and data are provided or not.
        \item If the contribution is a dataset and\slash or model, the authors should describe the steps taken to make their results reproducible or verifiable. 
        \item Depending on the contribution, reproducibility can be accomplished in various ways. For example, if the contribution is a novel architecture, describing the architecture fully might suffice, or if the contribution is a specific model and empirical evaluation, it may be necessary to either make it possible for others to replicate the model with the same dataset, or provide access to the model. In general. releasing code and data is often one good way to accomplish this, but reproducibility can also be provided via detailed instructions for how to replicate the results, access to a hosted model (e.g., in the case of a large language model), releasing of a model checkpoint, or other means that are appropriate to the research performed.
        \item While NeurIPS does not require releasing code, the conference does require all submissions to provide some reasonable avenue for reproducibility, which may depend on the nature of the contribution. For example
        \begin{enumerate}
            \item If the contribution is primarily a new algorithm, the paper should make it clear how to reproduce that algorithm.
            \item If the contribution is primarily a new model architecture, the paper should describe the architecture clearly and fully.
            \item If the contribution is a new model (e.g., a large language model), then there should either be a way to access this model for reproducing the results or a way to reproduce the model (e.g., with an open-source dataset or instructions for how to construct the dataset).
            \item We recognize that reproducibility may be tricky in some cases, in which case authors are welcome to describe the particular way they provide for reproducibility. In the case of closed-source models, it may be that access to the model is limited in some way (e.g., to registered users), but it should be possible for other researchers to have some path to reproducing or verifying the results.
        \end{enumerate}
    \end{itemize}

\item {\bf Open access to data and code}
    \item[] Question: Does the paper provide open access to the data and code, with sufficient instructions to faithfully reproduce the main experimental results, as described in supplemental material?
    \item[] Answer: \answerYes{}
    \item[] Justification: A public code repository is linked in the abstract (\url{https://github.com/Tianqi-Zhao/TabClustPFN}); all real-world benchmark datasets are drawn from public sources (OpenML-CC18, TabArena, AutoML, ZEUS, UCI) with their IDs/names listed in App.~\ref{sec:dataset_stats}, and the synthetic prior generators are described in App.~\ref{app:prior_data}.
    \item[] Guidelines:
    \begin{itemize}
        \item The answer \answerNA{} means that paper does not include experiments requiring code.
        \item Please see the NeurIPS code and data submission guidelines (\url{https://neurips.cc/public/guides/CodeSubmissionPolicy}) for more details.
        \item While we encourage the release of code and data, we understand that this might not be possible, so \answerNo{} is an acceptable answer. Papers cannot be rejected simply for not including code, unless this is central to the contribution (e.g., for a new open-source benchmark).
        \item The instructions should contain the exact command and environment needed to run to reproduce the results. See the NeurIPS code and data submission guidelines (\url{https://neurips.cc/public/guides/CodeSubmissionPolicy}) for more details.
        \item The authors should provide instructions on data access and preparation, including how to access the raw data, preprocessed data, intermediate data, and generated data, etc.
        \item The authors should provide scripts to reproduce all experimental results for the new proposed method and baselines. If only a subset of experiments are reproducible, they should state which ones are omitted from the script and why.
        \item At submission time, to preserve anonymity, the authors should release anonymized versions (if applicable).
        \item Providing as much information as possible in supplemental material (appended to the paper) is recommended, but including URLs to data and code is permitted.
    \end{itemize}

\item {\bf Experimental setting/details}
    \item[] Question: Does the paper specify all the training and test details (e.g., data splits, hyperparameters, how they were chosen, type of optimizer) necessary to understand the results?
    \item[] Answer: \answerYes{}
    \item[] Justification: Training details, optimizer, learning-rate schedule, pretraining steps, model architecture, benchmark construction, preprocessing, and baseline settings are described in Secs.~\ref{sec:methods}--\ref{sec:experiment} and Apps.~\ref{app:training_details}--\ref{app:experimental_details}.
    \item[] Guidelines:
    \begin{itemize}
        \item The answer \answerNA{} means that the paper does not include experiments.
        \item The experimental setting should be presented in the core of the paper to a level of detail that is necessary to appreciate the results and make sense of them.
        \item The full details can be provided either with the code, in appendix, or as supplemental material.
    \end{itemize}

\item {\bf Experiment statistical significance}
    \item[] Question: Does the paper report error bars suitably and correctly defined or other appropriate information about the statistical significance of the experiments?
    \item[] Answer: \answerYes{}
    \item[] Justification: We report median ARI/NMI ranks with IQRs (Tab.~\ref{tab:autok_gen_rank_comparison}, Fig.~\ref{fig:main_exp}). To assess significance across datasets, we conduct one-sided paired $t$-tests (Fig.~\ref{fig:ttest-significance}) and Wilcoxon signed-rank tests (Fig.~\ref{fig:wilcoxon-significance}) on per-dataset paired ARI differences, and report critical-difference (CD) diagrams (Fig.~\ref{fig:cd-significance}) based on average ARI ranks; full details are in App.~\ref{app:significance}.
    \item[] Guidelines:
    \begin{itemize}
        \item The answer \answerNA{} means that the paper does not include experiments.
        \item The authors should answer \answerYes{} if the results are accompanied by error bars, confidence intervals, or statistical significance tests, at least for the experiments that support the main claims of the paper.
        \item The factors of variability that the error bars are capturing should be clearly stated (for example, train/test split, initialization, random drawing of some parameter, or overall run with given experimental conditions).
        \item The method for calculating the error bars should be explained (closed form formula, call to a library function, bootstrap, etc.)
        \item The assumptions made should be given (e.g., Normally distributed errors).
        \item It should be clear whether the error bar is the standard deviation or the standard error of the mean.
        \item It is OK to report 1-sigma error bars, but one should state it. The authors should preferably report a 2-sigma error bar than state that they have a 96\% CI, if the hypothesis of Normality of errors is not verified.
        \item For asymmetric distributions, the authors should be careful not to show in tables or figures symmetric error bars that would yield results that are out of range (e.g., negative error rates).
        \item If error bars are reported in tables or plots, the authors should explain in the text how they were calculated and reference the corresponding figures or tables in the text.
    \end{itemize}

\item {\bf Experiments compute resources}
    \item[] Question: For each experiment, does the paper provide sufficient information on the computer resources (type of compute workers, memory, time of execution) needed to reproduce the experiments?
    \item[] Answer: \answerYes{}
    \item[] Justification: Sec.~\ref{sec:methods} (Model pretraining) reports that pretraining is run on four RTX~5090 GPUs for $10{,}000$ steps at a cost of approximately $92$ GPU hours; per-iteration training time and per-dataset inference runtime are additionally reported in Tab.~\ref{tab:pin_loss_ablation}, Fig.~\ref{fig:performance}, and Tab.~\ref{tab:performance_comparison}.
    \item[] Guidelines:
    \begin{itemize}
        \item The answer \answerNA{} means that the paper does not include experiments.
        \item The paper should indicate the type of compute workers CPU or GPU, internal cluster, or cloud provider, including relevant memory and storage.
        \item The paper should provide the amount of compute required for each of the individual experimental runs as well as estimate the total compute. 
        \item The paper should disclose whether the full research project required more compute than the experiments reported in the paper (e.g., preliminary or failed experiments that didn't make it into the paper). 
    \end{itemize}
    
\item {\bf Code of ethics}
    \item[] Question: Does the research conducted in the paper conform, in every respect, with the NeurIPS Code of Ethics \url{https://neurips.cc/public/EthicsGuidelines}?
    \item[] Answer: \answerYes{}
    \item[] Justification: The research conforms to the NeurIPS Code of Ethics: it uses only synthetic data and publicly available, license-compliant tabular benchmarks, involves no human subjects or personally identifiable information, and preserves anonymity in submission.
    \item[] Guidelines:
    \begin{itemize}
        \item The answer \answerNA{} means that the authors have not reviewed the NeurIPS Code of Ethics.
        \item If the authors answer \answerNo, they should explain the special circumstances that require a deviation from the Code of Ethics.
        \item The authors should make sure to preserve anonymity (e.g., if there is a special consideration due to laws or regulations in their jurisdiction).
    \end{itemize}

\item {\bf Broader impacts}
    \item[] Question: Does the paper discuss both potential positive societal impacts and negative societal impacts of the work performed?
    \item[] Answer: \answerNA{}
    \item[] Justification: This is foundational research on unsupervised tabular clustering with no direct path to high-stakes deployment; the potential positive impact is faster automated exploratory analysis, and we are not aware of specific negative societal impacts beyond those generic to clustering methods.
    \item[] Guidelines:
    \begin{itemize}
        \item The answer \answerNA{} means that there is no societal impact of the work performed.
        \item If the authors answer \answerNA{} or \answerNo, they should explain why their work has no societal impact or why the paper does not address societal impact.
        \item Examples of negative societal impacts include potential malicious or unintended uses (e.g., disinformation, generating fake profiles, surveillance), fairness considerations (e.g., deployment of technologies that could make decisions that unfairly impact specific groups), privacy considerations, and security considerations.
        \item The conference expects that many papers will be foundational research and not tied to particular applications, let alone deployments. However, if there is a direct path to any negative applications, the authors should point it out. For example, it is legitimate to point out that an improvement in the quality of generative models could be used to generate Deepfakes for disinformation. On the other hand, it is not needed to point out that a generic algorithm for optimizing neural networks could enable people to train models that generate Deepfakes faster.
        \item The authors should consider possible harms that could arise when the technology is being used as intended and functioning correctly, harms that could arise when the technology is being used as intended but gives incorrect results, and harms following from (intentional or unintentional) misuse of the technology.
        \item If there are negative societal impacts, the authors could also discuss possible mitigation strategies (e.g., gated release of models, providing defenses in addition to attacks, mechanisms for monitoring misuse, mechanisms to monitor how a system learns from feedback over time, improving the efficiency and accessibility of ML).
    \end{itemize}
    
\item {\bf Safeguards}
    \item[] Question: Does the paper describe safeguards that have been put in place for responsible release of data or models that have a high risk for misuse (e.g., pre-trained language models, image generators, or scraped datasets)?
    \item[] Answer: \answerNA{}
    \item[] Justification: The released artifact is a tabular clustering PFN trained on synthetic priors and standard public benchmarks; it is not a generative model, language model, or scraped-data release, and poses no high risk of misuse that would require additional safeguards.
    \item[] Guidelines:
    \begin{itemize}
        \item The answer \answerNA{} means that the paper poses no such risks.
        \item Released models that have a high risk for misuse or dual-use should be released with necessary safeguards to allow for controlled use of the model, for example by requiring that users adhere to usage guidelines or restrictions to access the model or implementing safety filters. 
        \item Datasets that have been scraped from the Internet could pose safety risks. The authors should describe how they avoided releasing unsafe images.
        \item We recognize that providing effective safeguards is challenging, and many papers do not require this, but we encourage authors to take this into account and make a best faith effort.
    \end{itemize}

\item {\bf Licenses for existing assets}
    \item[] Question: Are the creators or original owners of assets (e.g., code, data, models), used in the paper, properly credited and are the license and terms of use explicitly mentioned and properly respected?
    \item[] Answer: \answerYes{}
    \item[] Justification: 
    All existing assets are properly credited and used in accordance with their licenses: TabICL~\citep{qu2025tabicl} (BSD 3-Clause), ZEUS~\citep{marszaek2025zeus} (Apache-2.0), iResNet~\citep{pmlr-v97-behrmann19a} (MIT), MixSim~\citep{mixsim} (GPL $\geq 2$), \texttt{scikit-learn} baselines (BSD 3-Clause), and the public benchmarks OpenML-CC18~\citep{openmlsuite} (CC0/Public Domain per OpenML policy), TabArena~\citep{erickson2025tabarena} (Apache-2.0), the AutoML benchmark~\citep{automl} (MIT), and the UCI clustering collection~\citep{milaan9_clustering_datasets} (MIT).
    \item[] Guidelines:
    \begin{itemize}
        \item The answer \answerNA{} means that the paper does not use existing assets.
        \item The authors should cite the original paper that produced the code package or dataset.
        \item The authors should state which version of the asset is used and, if possible, include a URL.
        \item The name of the license (e.g., CC-BY 4.0) should be included for each asset.
        \item For scraped data from a particular source (e.g., website), the copyright and terms of service of that source should be provided.
        \item If assets are released, the license, copyright information, and terms of use in the package should be provided. For popular datasets, \url{paperswithcode.com/datasets} has curated licenses for some datasets. Their licensing guide can help determine the license of a dataset.
        \item For existing datasets that are re-packaged, both the original license and the license of the derived asset (if it has changed) should be provided.
        \item If this information is not available online, the authors are encouraged to reach out to the asset's creators.
    \end{itemize}

\item {\bf New assets}
    \item[] Question: Are new assets introduced in the paper well documented and is the documentation provided alongside the assets?
    \item[] Answer: \answerYes{}
    \item[] Justification: We release the pretrained \ours{} model, the GMM and ZEUS prior-data generators through the public repository linked in the abstract, with accompanying documentation and reproducibility scripts.
    \item[] Guidelines:
    \begin{itemize}
        \item The answer \answerNA{} means that the paper does not release new assets.
        \item Researchers should communicate the details of the dataset\slash code\slash model as part of their submissions via structured templates. This includes details about training, license, limitations, etc. 
        \item The paper should discuss whether and how consent was obtained from people whose asset is used.
        \item At submission time, remember to anonymize your assets (if applicable). You can either create an anonymized URL or include an anonymized zip file.
    \end{itemize}

\item {\bf Crowdsourcing and research with human subjects}
    \item[] Question: For crowdsourcing experiments and research with human subjects, does the paper include the full text of instructions given to participants and screenshots, if applicable, as well as details about compensation (if any)? 
    \item[] Answer: \answerNA{}
    \item[] Justification: The paper does not involve any crowdsourcing or research with human subjects.
    \item[] Guidelines:
    \begin{itemize}
        \item The answer \answerNA{} means that the paper does not involve crowdsourcing nor research with human subjects.
        \item Including this information in the supplemental material is fine, but if the main contribution of the paper involves human subjects, then as much detail as possible should be included in the main paper. 
        \item According to the NeurIPS Code of Ethics, workers involved in data collection, curation, or other labor should be paid at least the minimum wage in the country of the data collector. 
    \end{itemize}

\item {\bf Institutional review board (IRB) approvals or equivalent for research with human subjects}
    \item[] Question: Does the paper describe potential risks incurred by study participants, whether such risks were disclosed to the subjects, and whether Institutional Review Board (IRB) approvals (or an equivalent approval/review based on the requirements of your country or institution) were obtained?
    \item[] Answer: \answerNA{}
    \item[] Justification: The paper does not involve human subjects, so no IRB approval (or equivalent) is required.
    \item[] Guidelines:
    \begin{itemize}
        \item The answer \answerNA{} means that the paper does not involve crowdsourcing nor research with human subjects.
        \item Depending on the country in which research is conducted, IRB approval (or equivalent) may be required for any human subjects research. If you obtained IRB approval, you should clearly state this in the paper. 
        \item We recognize that the procedures for this may vary significantly between institutions and locations, and we expect authors to adhere to the NeurIPS Code of Ethics and the guidelines for their institution. 
        \item For initial submissions, do not include any information that would break anonymity (if applicable), such as the institution conducting the review.
    \end{itemize}

\item {\bf Declaration of LLM usage}
    \item[] Question: Does the paper describe the usage of LLMs if it is an important, original, or non-standard component of the core methods in this research? Note that if the LLM is used only for writing, editing, or formatting purposes and does \emph{not} impact the core methodology, scientific rigor, or originality of the research, declaration is not required.
    \item[] Answer: \answerNA{}
    \item[] Justification: LLMs are not used as an important, original, or non-standard component of the core methodology; any LLM use was limited to writing/editing assistance and did not affect the scientific content or originality of the research.
    \item[] Guidelines:
    \begin{itemize}
        \item The answer \answerNA{} means that the core method development in this research does not involve LLMs as any important, original, or non-standard components.
        \item Please refer to our LLM policy in the NeurIPS handbook for what should or should not be described.
    \end{itemize}

\end{enumerate}

%% file: main.bbl
\begin{thebibliography}{10}

\bibitem{aggarwal2001surprising}
C.~C. Aggarwal, A.~Hinneburg, and D.~A. Keim.
\newblock On the surprising behavior of distance metrics in high dimensional
  space.
\newblock In {\em International conference on database theory}, pages 420--434.
  Springer, 2001.

\bibitem{pmlr-v97-behrmann19a}
J.~Behrmann, W.~Grathwohl, R.~T.~Q. Chen, D.~Duvenaud, and J.-H. Jacobsen.
\newblock Invertible residual networks.
\newblock In {\em International Conference on Machine Learning}, Proceedings of
  Machine Learning Research, 2019.

\bibitem{bhaskaran2025transformer}
P.~Bhaskaran and T.~Viering.
\newblock Transformers can do {Bayesian} clustering.
\newblock {\em arXiv preprint arXiv:2510.24318}, 2025.

\bibitem{openmlsuite}
B.~Bischl, G.~Casalicchio, M.~Feurer, P.~Gijsbers, F.~Hutter, M.~Lang, R.~G.
  Mantovani, J.~N. van Rijn, and J.~Vanschoren.
\newblock Open{ML} benchmarking suites.
\newblock In {\em Proceedings of the Neural Information Processing Systems
  Track on Datasets and Benchmarks}, 2021.

\bibitem{crouse2016implementing}
D.~F. Crouse.
\newblock On implementing 2d rectangular assignment algorithms.
\newblock {\em IEEE Transactions on Aerospace and Electronic Systems},
  52(4):1679--1696, 2016.

\bibitem{demsar2006statistical}
J.~Dem{\v{s}}ar.
\newblock Statistical comparisons of classifiers over multiple data sets.
\newblock {\em Journal of Machine Learning Research}, 7(1):1--30, 2006.

\bibitem{erickson2025tabarena}
N.~Erickson, L.~Purucker, A.~Tschalzev, D.~Holzm{\"u}ller, P.~M. Desai,
  D.~Salinas, and F.~Hutter.
\newblock Tabarena: A living benchmark for machine learning on tabular data.
\newblock In {\em Proceedings of the 39th Conference on Neural Information
  Processing Systems (NeurIPS)}, 2025.

\bibitem{ester1996density}
M.~Ester, H.-P. Kriegel, J.~Sander, and X.~Xu.
\newblock A density-based algorithm for discovering clusters in large spatial
  databases with noise.
\newblock In {\em Proceedings of the Second International Conference on
  Knowledge Discovery and Data Mining}, page 226–231, 1996.

\bibitem{flynt2019sari}
A.~Flynt, N.~Dean, and R.~Nugent.
\newblock {sARI}: a soft agreement measure for class partitions incorporating
  assignment probabilities.
\newblock {\em Advances in Data Analysis and Classification}, 13(1):303--323,
  2019.

\bibitem{automl}
P.~Gijsbers, M.~L.~P. Bueno, S.~Coors, E.~LeDell, S.~Poirier, J.~Thomas,
  B.~Bischl, and J.~Vanschoren.
\newblock Amlb: an automl benchmark.
\newblock {\em Journal of Machine Learning Research}, 25(101):1--65, 2024.

\bibitem{grinsztajn2025tabpfn}
L.~Grinsztajn, K.~Fl{\"o}ge, O.~Key, F.~Birkel, P.~Jund, B.~Roof, B.~J{\"a}ger,
  D.~Safaric, S.~Alessi, A.~Hayler, et~al.
\newblock Tabpfn-2.5: Advancing the state of the art in tabular foundation
  models.
\newblock {\em arXiv preprint arXiv:2511.08667}, 2025.

\bibitem{hollmann2025accurate}
N.~Hollmann, S.~M{\"u}ller, L.~Purucker, A.~Krishnakumar, M.~K{\"o}rfer, S.~B.
  Hoo, R.~T. Schirrmeister, and F.~Hutter.
\newblock Accurate predictions on small data with a tabular foundation model.
\newblock {\em Nature}, 637(8045):319--326, 2025.

\bibitem{hubert1985comparing}
L.~Hubert and P.~Arabie.
\newblock Comparing partitions.
\newblock {\em Journal of classification}, 2(1):193--218, 1985.

\bibitem{johnson1967hierarchical}
S.~C. Johnson.
\newblock Hierarchical clustering schemes.
\newblock {\em Psychometrika}, 32(3):241--254, 1967.

\bibitem{kalogeratos2012dipmeans}
A.~Kalogeratos and A.~Likas.
\newblock Dip-means: an incremental clustering method for estimating the number
  of clusters.
\newblock In {\em Advances in Neural Information Processing Systems},
  volume~25, 2012.

\bibitem{kulis2012revisiting}
B.~Kulis and M.~I. Jordan.
\newblock Revisiting k-means: New algorithms via {Bayesian} nonparametrics.
\newblock In {\em International Conference on Machine Learning}, 2012.

\bibitem{lee2019set}
J.~Lee, Y.~Lee, J.~Kim, A.~Kosiorek, S.~Choi, and Y.~W. Teh.
\newblock Set transformer: A framework for attention-based
  permutation-invariant neural networks.
\newblock In {\em International Conference on Machine Learning}, pages
  3744--3753. PMLR, 2019.

\bibitem{lee2019deep}
J.~Lee, Y.~Lee, and Y.~W. Teh.
\newblock Deep amortized clustering.
\newblock {\em arXiv preprint arXiv:1909.13433}, 2019.

\bibitem{li2025clustering}
P.~Li, F.~Li, J.~Wang, and Y.~Qian.
\newblock Clustering method for tabular data based on pretrained foundation
  models with synthetic data.
\newblock {\em Journal of Computer Research and Development}, 62(9), 2025.

\bibitem{lloyd1982least}
S.~Lloyd.
\newblock Least squares quantization in {PCM}.
\newblock {\em IEEE Transactions on Information Theory}, 28(2):129--137, 1982.

\bibitem{ma2025tabdpt}
J.~Ma, V.~Thomas, R.~Hosseinzadeh, A.~Labach, H.~Kamkari, J.~C. Cresswell,
  K.~Golestan, G.~Yu, A.~L. Caterini, and M.~Volkovs.
\newblock Tab{DPT}: Scaling tabular foundation models on real data.
\newblock In {\em The Thirty-ninth Annual Conference on Neural Information
  Processing Systems}, 2025.

\bibitem{marszaek2025zeus}
P.~Marsza{\l}ek, T.~Ku{\'s}mierczyk, W.~Wydma{\'n}ski, J.~Tabor, and
  M.~{\'S}mieja.
\newblock {ZEUS}: Zero-shot embeddings for unsupervised separation of tabular
  data.
\newblock In {\em The Thirty-ninth Annual Conference on Neural Information
  Processing Systems}, 2025.

\bibitem{mixsim}
V.~Melnykov, W.-C. Chen, and R.~Maitra.
\newblock {MixSim}: An {R} package for simulating data to study performance of
  clustering algorithms.
\newblock {\em Journal of Statistical Software}, 51(12):1--25, 2012.

\bibitem{milaan9_clustering_datasets}
milaan9.
\newblock Clustering-datasets.
\newblock \url{https://github.com/milaan9/Clustering-Datasets}, 2019.
\newblock GitHub repository, accessed January 2026.

\bibitem{ng2001spectral}
A.~Y. Ng, M.~I. Jordan, and Y.~Weiss.
\newblock On spectral clustering: analysis and an algorithm.
\newblock In {\em Advances in Neural Information Processing Systems},
  volume~14, page 849–856, 2001.

\bibitem{pakman2020neural}
A.~Pakman, Y.~Wang, C.~Mitelut, J.~Lee, and L.~Paninski.
\newblock Neural clustering processes.
\newblock In {\em International Conference on Machine Learning}, pages
  7455--7465. PMLR, 2020.

\bibitem{pearl2009causality}
J.~Pearl.
\newblock {\em Causality}.
\newblock Cambridge university press, 2009.

\bibitem{qu2025tabicl}
J.~Qu, D.~Holzm{\"u}ller, G.~Varoquaux, and M.~L. Morvan.
\newblock Tab{ICL}: A tabular foundation model for in-context learning on large
  data.
\newblock In {\em Forty-second International Conference on Machine Learning},
  2025.

\bibitem{qu2026tabiclv2}
J.~QU, D.~Holzm{\"u}ller, G.~Varoquaux, and M.~L. Morvan.
\newblock Tab{ICL}v2: A better, faster, scalable, and open tabular foundation
  model.
\newblock In {\em Forty-third International Conference on Machine Learning},
  2026.

\bibitem{rauf2025tabledc}
H.~T. Rauf, A.~Freitas, and N.~W. Paton.
\newblock {TableDC}: Deep clustering for tabular data.
\newblock {\em Proceedings of the ACM on Management of Data}, 3(3):1--28, 2025.

\bibitem{ren2024deep}
Y.~Ren, J.~Pu, Z.~Yang, J.~Xu, G.~Li, X.~Pu, P.~S. Yu, and L.~He.
\newblock Deep clustering: A comprehensive survey.
\newblock {\em IEEE transactions on neural networks and learning systems},
  36(4):5858--5878, 2025.

\bibitem{Sinkhorn1967}
R.~Sinkhorn and P.~Knopp.
\newblock Concerning nonnegative matrices and doubly stochastic matrices.
\newblock {\em Pacific Journal of Mathematics}, 21(2):343–348, May 1967.

\bibitem{zhang2025mitra}
X.~Zhang, D.~C. Maddix, J.~Yin, N.~Erickson, A.~F. Ansari, B.~Han, S.~Zhang,
  L.~Akoglu, C.~Faloutsos, M.~W. Mahoney, C.~Hu, H.~Rangwala, G.~Karypis, and
  B.~Wang.
\newblock Mitra: Mixed synthetic priors for enhancing tabular foundation
  models.
\newblock In {\em The Thirty-ninth Annual Conference on Neural Information
  Processing Systems}, 2025.

\end{thebibliography}
